\documentclass[10pt]{article}
\pdfoutput=1
\usepackage[centertags]{amsmath}
\usepackage{amssymb}
\usepackage[margin=1.5in]{geometry}
\usepackage{natbib}
\newcommand{\ra}{\rightarrow}
\newcommand{\cdt}{\!\cdot\!}
\newcommand{\eg}{e.g.\ }
\newcommand{\ie}{i.e.\ }
\newcommand{\etc}{etc.\ }
\newcommand{\wrt}{wrt.\ }
\newcommand{\prn}[1]{\left ( #1 \right )}
\newcommand{\brc}[1]{\left\{ #1 \right\}}
\newcommand{\brk}[1]{\left [ #1 \right ]}
\newcommand{\abs}[1]{\left\lvert #1 \right\rvert}
\newcommand{\nrm}[1]{\left\lVert #1 \right\rVert}
\newcommand{\av}[1]{\left\langle #1 \right\rangle}
\newcommand{\set}[2]{\left\{ #1 \,\middle |\, #2 \right\}}

\newcommand{\intd}[2][\rule{0mm}{0mm}]{\int #1\!\!\dr #2\,}
\newcommand{\dr}{\mathrm{d}}
\newcommand{\e}{\mathrm{e}}
\newcommand{\lmax}{_\text{max}}
\newcommand{\lmin}{_\text{min}}
\newcommand{\inv}{^{-1}}
\newcommand{\trans}{^\mathrm{T}}
\newcommand{\transp}{{}^\mathrm{T}}
\newcommand{\means}{\Longleftrightarrow}
\newcommand{\psq}{^{\prime\,2}}
\newcommand{\ppsq}{^{\prime\prime\,2}}
\newcommand{\CM}{\mathcal{M}}
\newcommand{\CO}{\mathcal{O}}
\newcommand{\R}{\mathbb{R}}
\renewcommand{\Pr}{\mathbb{P}}
\newcommand{\dt}{\dr t}
\newcommand{\I}{\mathbf{I}}
\newcommand{\U}{\mathbf{U}}
\newcommand{\Usp}{\mathcal{U}}
\newcommand{\proj}{\mathbf{A}}
\newcommand{\projsp}{\mathcal{A}}
\newcommand{\pang}{\boldsymbol{\theta}}
\newcommand{\prang}{\boldsymbol{\varphi}}

\newcommand{\B}{\mathbb{B}}
\newcommand{\Sp}{\mathbb{S}}
\newcommand{\grass}{\mathfrak{G}}
\newcommand{\gauss}{\mathfrak{g}}
\newcommand{\dist}{\mathcal{D}}
\newcommand{\distpr}{\dist_\proj}

\newcommand{\uv}{\mathbf{u}}
\newcommand{\vv}{\mathbf{v}}
\newcommand{\sv}{\mathbf{s}}
\newcommand{\xv}{\mathbf{x}}
\newcommand{\yv}{\mathbf{y}}
\newcommand{\ec}{x}
\newcommand{\ic}{\sigma}
\newcommand{\svd}{s}
\newcommand{\lng}{_\text{long}}
\newcommand{\shrt}{_\text{short}}
\newcommand{\thx}{\theta\lmax}
\newcommand{\thxc}{\theta_{\conec}}
\newcommand{\thxs}{\theta_{\cones}}
\newcommand{\clsz}{\gamma}
\newcommand{\cldm}{d}
\newcommand{\cell}{\mathcal{S}}
\newcommand{\conec}{\mathcal{C}}
\newcommand{\cones}{\mathcal{T}}
\newcommand{\gnt}{\mathcal{E}}
\newcommand{\gntc}{\gnt_{\conec}}
\newcommand{\gnts}{\gnt_{\cones}}
\DeclareMathOperator{\vol}{Vol}
\newcommand{\V}{\mathcal{V}}
\newcommand{\dbnd}{\overline{\delta}}
\newcommand{\dmx}{\delta_0}
\newcommand{\mbnd}{\overline{M}}
\newcommand{\mmn}{M^*}
\usepackage{graphicx}
\usepackage{adjustbox}
\newcommand{\aligntop}[1]{\adjustbox{valign=t}{#1}}
\usepackage[inline]{enumitem}
\newlist{myenuma}{enumerate*}{10}
\setlist[myenuma]{label=(\alph*),ref=\alph*}
\usepackage[hyper]{sl-titling}
\usepackage{hyperref}
\usepackage[capitalise]{cleveref}
\usepackage[all]{hypcap}
\newcommand{\suppname}{Appendices}
\newcommand{\supp}{}
\newcommand{\main}{}

\title{Random projections of random manifolds}%
\author{%
Subhaneil Lahiri,
\afref{af:stanford}
Peiran Gao,\afref{af:spacex}
and
Surya Ganguli\afref{af:stanford}
}
\affiliation{Department of Applied Physics, Stanford University, Stanford, CA 94305, USA \label{af:stanford}}
\affiliation{Space Exploration Technologies  Co., Hawthorne, CA 90250, USA \label{af:spacex}}
\date{}%
\begin{document}
\maketitle
\begin{abstract}
Interesting data often concentrate on low dimensional smooth manifolds inside a high dimensional ambient space.
Random projections are a simple, powerful tool for dimensionality reduction of such data.
Previous works have studied bounds on how many projections are needed to accurately preserve the geometry of these manifolds, given their intrinsic dimensionality, volume and curvature.
However, such works employ definitions of volume and curvature that are inherently difficult to compute.
Therefore such theory cannot be easily tested against numerical simulations to understand the tightness of the proven bounds.
We instead study typical distortions arising in random projections of an ensemble of smooth Gaussian random manifolds.
We find explicitly computable, approximate theoretical bounds on the number of projections required to accurately preserve the geometry of these manifolds.
Our bounds, while approximate, can only be violated with a probability that is exponentially small in the ambient dimension, and therefore they hold with high probability in cases of practical interest.
Moreover, unlike previous work, we test our theoretical bounds against numerical experiments on the actual geometric distortions that typically occur for random projections of random smooth manifolds.
We find our bounds are tighter than previous results by several orders of magnitude.
\end{abstract}
\tableofcontents



\section{Introduction}\label{m:sec:intro}

The very high dimensionality of modern datasets poses severe statistical and computational challenges for machine learning.
Thus dimensionality reduction methods that lead to a compressed or lower dimensional description of data is of great interest to a variety of fields.
A fundamental desideratum of dimensionality reduction is the preservation of distances between all pairs of data points of interest.
Since many machine learning algorithms depend only on pairwise distances between data points, often computation in the compressed space is almost as good as computation in the much higher dimensional ambient space.
In essence, many algorithms achieve similar performance levels in the compressed space at much higher computational efficiency, without a large sacrifice in statistical efficiency.
This logic applies, for example, to regression \citep{zhou2009compressed}, signal detection \citep{duarte2006sparse}, classification \citep{blum2006random,haupt2006compressive,davenport2007smashed,duarte2007multiscale}, manifold learning \citep{hegde2007random}, and nearest neighbor finding \citep{indyk1998approximate}.

Geometry preservation through dimensionality reduction is possible because, while data often lives in a very high dimensional ambient space, the data usually concentrates on much lower dimensional manifolds within the space.
Recent work has shown that even \emph{random} dimensionality reduction, whereby the data is projected onto a random subspace, can preserve the geometry of data or signal manifolds to surprisingly high levels of accuracy.
Moreover this accuracy scales favorably with the complexity of the geometric structure of the data.
For example the celebrated Johnson-Lindenstrauss (JL) lemma \citep{Johnson1984extensions,indyk1998approximate,Dasgupta2003JLlemma} states that the number of random projections needed to reliably achieve a fractional distance distortion on $P$ points that is less than $\epsilon$, scales as $\frac{\log P}{\epsilon^2}$, \ie only \emph{logarithmically} with $P$.
In compressed sensing, the space of $K$-sparse $N$ dimensional signals corresponds to a union of $K$ dimensional coordinate subspaces.
The fact that these signals can be reconstructed with only $\CO(K\log K/N)$ random measurements can be understood in terms of geometry preservation of this set through random projections \citep[see][]{candes2005decoding,baraniuk2008simple}.
Finally, while random projections are widely applied across many fields, a particularly interesting application domain lies in neuroscience.
Indeed, as reviewed by \citet{ganguli2012annrevs} and \citet{advani2013statistical}, random projections may be employed both by neuroscientists to gather information from neural circuits using many fewer measurements, as well as potentially by neural circuits to communicate information using many fewer neurons.
Moreover, the act of recording from a subset of neurons itself could potentially be modeled as a random projection \citep[see][]{simplicitycomp15}.

Perhaps one of the most universal hypotheses for low dimensional structure in data is a smooth $K$ dimensional manifold $\CM$ embedded as a submanifold in Euclidean space $\R^N$.
In particular, recent seminal works have shown random projections preserve the geometry of smooth manifolds to an accuracy that depends on the curvature and volume of the manifold \citep[see][]{Baraniuk2009JLmfld,clarkson2008tighter,verma2011note}.
However, the theoretical techniques used lead to highly complex measures of geometric complexity that are difficult to explicitly compute in general.
For example, the results of \citet{Baraniuk2009JLmfld} required knowing the condition number of $\CM$, which is the inverse of the smallest distance normal to $\CM$ in $\R^N$ at which the normal neighborhood of $\CM$ in $\R^N$ intersects itself.
In addition, the results of \citet{Baraniuk2009JLmfld} required knowledge of the geodesic covering regularity, which is related to the smallest number of points needed to form a cover of $\CM$, such that \emph{all} points in $\CM$ are within a given geodesic distance of the cover.
The number of points in such a covering is also required for the results of \citet{verma2011note}.
Alternatively, the results of \citet{clarkson2008tighter} required not only knowledge of the volume of $\CM$ under the standard Riemmannian volume measure, but also, a curvature measure related to the volume of the image of $\CM$ under the Gauss map, which maps each point $\xv \in\CM$ to its $K$ dimensional tangent plane $\mathcal{T}_\xv \CM$.
The image of the Gauss map is a submanifold of the Grassmannian $\grass_{K,N}$ of all $K$ dimensional subspaces of $\R^N$, and its volume must be computed with respect to the standard Riemannian volume measure on the Grassmannian.
Also, the results of \citet{clarkson2008tighter} required knowledge of the number of points needed to form covers with respect to both measures of volume of $\CM$ and curvature of $\CM$, with the latter described by the volume of the image of the Gauss map.

Moreover, it is unclear how tight either of the results in \citet{Baraniuk2009JLmfld} and \citet{clarkson2008tighter} actually are.
For example, the bounds on the number of required projections to preserve geometry in \cite{Baraniuk2009JLmfld} had constants that were $\CO(3000)$.
Potentially tighter bounds were found in \cite{clarkson2008tighter} by using the average curvature rather than maximum curvature of the manifold $\CM$, but the constants in these bounds were not explicitly computed in \cite{clarkson2008tighter}.
In practical applications of these dimensionality reduction techniques, the values of these constants are necessary to determine how many projections are required.
The bound derived by \cite{verma2011note} does contain explicit constants (although some were as large as $2^{18}$), but that bound applies to the distortion of the lengths of curves on the manifold, which is weaker than bounds on pairwise Euclidean distance (see \cref{s:eq:distcurv} in \cref{s:sec:distort}\supp).
Ideally, one would like to conduct simulations to understand the tightness of the bounds proven in these works.
However, a major impediment to testing theory against experiment by conducting such simulations lies in the difficulty of numerically evaluating geometric quantities like the manifold condition number, geodesic covering regularity,
Riemannian and Grassmannian volumes, 
and the sizes of coverings with respect to both of these volumes.

Here we take a different perspective by considering random projections of an \emph{ensemble} of random manifolds.
By studying the geometric distortion induced by random projections of typical random realizations of such smooth random manifolds,
we find much tighter, but approximate, bounds on the number or random projections required
to preserve geometry to a given accuracy.
Essentially, a shift in perspective from a fixed given manifold $\CM$, as studied in \citet{Baraniuk2009JLmfld} and \citet{clarkson2008tighter}, to an ensemble of random manifolds, as studied here, enables us to combine a sequence of approximations and inequalities to derive approximate bounds.
Our main approximations involve neglecting large fluctuations in the geometry of manifolds in our ensemble.
In particular, the probability of such fluctuations are exponentially suppressed by the ambient dimensionality $N$, so our bounds, while approximate, are exceedingly unlikely to be violated in practical situations of interest, as $N$ can be quite large.
Interestingly, our methods also enable us to numerically compute lower bounds on the manifold condition number and geodesic covering regularity for realizations of manifolds in our ensemble, thereby enabling us to numerically evaluate the tightness of the bounds proven by \cite{Baraniuk2009JLmfld} (but not by \cite{clarkson2008tighter} as precise constants were not provided there).

We find that our approximate bounds derived here on the number of projections required to preserve the geometry of manifolds are more than two orders of magnitude better than this previous bound.
Moreover, we conduct simulations to evaluate the exact scaling relations relating the probability and accuracy of geometry preservation under random projections to the number of random projections chosen and the manifold volume and curvature.
Our numerical experiments yield strikingly simple, and, to our knowledge, new scaling relations relating the accuracy of random projections to the dimensionality and geometry of smooth manifolds.


\section{Overall approach and background}\label{m:sec:strategy}

Here we introduce the model of random manifolds that we work with (\cref{m:sec:randmanmod}) and the notion of random projections and geometry preservation (\cref{m:sec:geodistrp}).
We then discuss our overall strategy for analyzing how accurately a random projection preserves the geometry of a random manifold (\cref{m:sec:strat}).


\subsection[A statistical model of smooth random submanifolds of \texorpdfstring{$\R^N$}{R**N}]{A statistical model of smooth random submanifolds of \texorpdfstring{$\R^N$}{R**N}}\label{m:sec:randmanmod}

We consider $K$ dimensional random Gaussian submanifolds, $\CM$, of $\R^N$, described by an embedding, $\ec^i = \phi^i(\ic^\alpha)$, where $\ec^i$ ($i,j = 1,\ldots,N$) are Cartesian coordinates for the ambient space $\R^N$, $\ic^\alpha$ ($\alpha,\beta = 1,\ldots,K$) are intrinsic coordinates on the manifold, and $\phi^i(\ic)$ are (multidimensional) Gaussian processes (as in the Gaussian process latent variable models of \cite{Lawrence2005gplvm}), with
\begin{equation}\label{m:eq:randmangauss}
  \av{\phi^i(\ic)} = 0,
  \qquad
  \av{\phi^i(\ic_1) \phi^j(\ic_2)} = Q^{ij}(\ic_1 - \ic_2).
\end{equation}
We assume each intrinsic coordinate $\sigma^\alpha$ has an extent $L_\alpha$, and the random embedding functions have a correlation length scale  $\lambda_\alpha$ along each intrinsic coordinate, so that the kernel is given by
\begin{equation}\label{m:eq:gaussgausskernel}
  Q^{ij}(\Delta\ic) = \frac{\ell^2}{N} \, \delta^{ij} \e^{-\frac{\rho}{2}},
  \qquad
  \rho = \sum_\alpha \prn{\frac{\Delta\ic^\alpha}{\lambda_\alpha}}^2.
\end{equation}
Here the kernel is translation invariant, and so is only a function of the separation in intrinsic coordinates, $\Delta\ic = \ic_1 - \ic_2$, while the embedding functions $\phi^i$ are independent across the ambient Cartesian coordinates.
While our results apply to more general functional forms of correlation decay in the kernel, in this work we focus our calculations on the choice of a Gaussian profile of decay, namely the factor $\e^{-\frac{\rho}{2}}$ in the kernel $Q$ in \cref{m:eq:gaussgausskernel}.


\subsection{Geometric distortion induced by random projections}\label{m:sec:geodistrp}

We are interested in how the geometry of a submanifold $\CM \subset \R^N$ is distorted by a projection onto a random $M$ dimensional subspace.
Let $\proj$ be an $M$ by $N$ random projection matrix whose rows from an orthonormal basis for this random subspace, drawn from a uniform distribution over the Grassmannian $\grass_{M,N}$ of all $M$ dimensional subspaces of $\R^N$.
The geometric distortion of a single point $\uv \in \R^N$ under any projection $\proj$ is defined as
\begin{equation*}
  \distpr(\uv) = \abs{\sqrt{\frac{N}{M}} \frac{\nrm{\proj\uv}_2}{\nrm{\uv}_2} - 1}.
\end{equation*}%
$\distpr(\uv)$ reflects the fractional change in the length of $\uv$ incurred by the projection, and the scaling with $N$ and $M$ is chosen so that its expected value over the random choice of $\proj$ is $0$.
More generally, the distortion of any subset $\cell \subset \R^N$ is defined as the worst case distortion over all of its elements:
\begin{equation*}
  \distpr(\cell) = \max_{\uv \in \cell} \, \distpr(\uv).
\end{equation*}%
Ideally we would like to guarantee a small worst case distortion $\distpr(\cell) \leq \epsilon$ with a small failure probability $\delta$ over the choice of random projection $\proj$, where $\delta$ is defined as
\begin{equation}\label{m:eq:failprob}
  \delta = \Pr\brk{\distpr(\cell) > \epsilon}.
\end{equation}
In general, the failure probability $\delta$ will grow as the geometric complexity or size of the set $\cell$ grows, the desired distortion level $\epsilon$ decreases, or the projection dimensionality $M$ decreases.

For practical applications, one is often faced with the task of choosing the number of random projections, $M$, to use.
To minimize computational costs, one would like to minimize $M$ while maintaining a desired small distortion level $\epsilon$ and failure probability $\delta$ in \cref{m:eq:failprob} to ensure the accuracy of subsequent computations.
Thus an important quantity is the minimum projection dimensionality $\mmn(\epsilon, \delta)$ that is \emph{necessary} to guarantee distortion at \emph{most} $\epsilon$ with a success probability of \emph{at least}  $1 - \delta$:
\begin{equation*}
  \mmn(\epsilon, \delta) = \min\, M \in \mathbb{N} \quad \text{s.t.} \quad \Pr\brk{\distpr(\cell) \leq \epsilon} \geq 1 - \delta.
\end{equation*}%
Any upper bound $\dbnd(\epsilon, M)$ on the failure probability $\delta$, at fixed projection dimensionality $M$ and distortion $\epsilon$, naturally yields an upper bound $\mbnd(\epsilon, \delta)$ on the minimum projection dimensionality $\mmn(\epsilon, \delta)$ required to guarantee a prescribed distortion $\epsilon$ and failure probability $\delta$ such that \cref{m:eq:failprob} holds:
\begin{equation}\label{m:eq:dbnd_mbnd}
  \delta \leq \dbnd(\epsilon, M)
  \qquad \implies \qquad
  \mmn(\epsilon, \delta) \leq \mbnd(\epsilon, \delta),
\end{equation}
where the function $\mbnd(\epsilon, \delta)$ is found by inverting the function $\dbnd(\epsilon, M)$.
The function $\mbnd(\epsilon, \delta)$ allows us to give an answer to the question:  what projection dimensionality $M$ is \emph{sufficient} to guarantee that the failure probability $\delta$ in \cref{m:eq:failprob} is less than a prescribed failure probability $\dmx$ at a prescribed distortion level $\epsilon$?
Given an upper bound $\delta \leq \dbnd(\epsilon, M)$, a sufficient condition for the guarantee is choosing $M$ such that $\dbnd(\epsilon, M) \leq \dmx$.
Thus, \cref{m:eq:dbnd_mbnd} yields a sufficient condition on the projection dimensionality $M$ to guarantee distortion at most $\epsilon$ with success probability at least $1 - \dmx$:
\begin{equation*}
  M \geq \mbnd(\epsilon, \dmx)
  \qquad \implies \qquad
  \Pr\brk{\distpr(\cell) \leq \epsilon} \geq 1 - \dmx.
\end{equation*}%

For example, in the simplest case where $\cell$ is a single point in $\R^N$, a simple concentration of measure argument \citep[see][]{Dasgupta2003JLlemma} yields an upper bound on the failure probability. At small $\epsilon \ll 1$ this argument yields,
\begin{equation}\label{m:eq:JLsingle}
  \delta \leq \dbnd(\epsilon, M) = 2 \, \e^{-\frac{1}{4} M \epsilon^2}
  \qquad \implies \qquad
  \mmn(\epsilon, \delta) \leq \mbnd(\epsilon, \delta) = \frac{4 \ln(2/\delta)}{\epsilon^2}.
\end{equation}
More generally, when the set $\cell$ consists of all $\binom{ P }{ 2}$ displacement vectors between a cloud of $P$ points, a simple union bound applied to this result yields the celebrated JL lemma \citep[see][]{Johnson1984extensions,indyk1998approximate,Dasgupta2003JLlemma}.  When $\epsilon \ll 1$, the JL lemma yields
\begin{multline}\label{m:eq:JLfail}
  \delta \leq \dbnd(\epsilon, M) = 2 \tbinom{ P }{ 2} \e^{-\frac{1}{4}M \epsilon^2} \, \approx \, \e^{-\frac{1}{4}(M \epsilon^2 - 8 \ln P)}
  \\ \implies \quad
  \mmn(\epsilon, \delta) \leq \mbnd(\epsilon, \delta) =  \frac{8 \ln P + 4 \ln \frac{2}{\delta}}{\epsilon^2}.
\end{multline}
Thus, remarkably, because the failure probability of preserving the length of a single point is \emph{exponentially} suppressed by $M$, the minimal number of random projections $M^*$ required to preserve the pairwise geometry of a cloud of $P$ points, within distortion $\epsilon$ with success probability $1-\delta$, scales at most {\it logarithmically} with the number of points.
Indeed, any choice of $M \geq \mbnd(\epsilon,\delta)$ guarantees the geometry preservation condition in \cref{m:eq:failprob}.

Another instructive example, which we will use below, is a JL type lemma when the set $\cell \subset \R^N$ is a $K$ dimensional linear subspace.
This result was proven by \citet[Lemma 5.1]{baraniuk2008simple}, and the proof strategy is as follows.
First, a random projection $\proj$ preserves distances between all pairs of points in a $K$ dimensional subspace with distortion $\epsilon$ if and only if it preserves all points on the unit sphere $\Sp^{K-1}$ with distortion $\epsilon$, due to the linearity of both the projection and the subspace.
Then a covering argument reveals that if any projection $\proj$ preserves a particular set of $(\tfrac{12}{\epsilon})^K$ points (that cover $\Sp^{K-1}$) with distortion less than $\tfrac{\epsilon}{2}$, then $\proj$ will preserve \emph{all} points in  $\mathbb S^{K-1}$ with distortion less than $\epsilon$.
Finally, applying the JL lemma with distortion $\tfrac{\epsilon}{2}$ to the covering set of $(\tfrac{12}{\epsilon})^K$ points yields an upper bound on the failure probability in \cref{m:eq:failprob} when $\cell$ is a $K$ dimensional subspace:
\begin{equation}\label{m:eq:JLsubspace}
  \delta \leq 2 (\tfrac{12}{\epsilon})^K \e^{-\frac{1}{4}M \prn{\frac{\epsilon}{2}}^2}
    =  \e^{-\prn{\frac{1}{16}M \epsilon^2 - K \ln \tfrac{12}{\epsilon} -\ln 2}}
    \quad \implies \quad
  \mmn(\epsilon, \delta) < \tfrac{16 \brk{ K \ln \frac{12}{\epsilon} + \ln \frac{2}{\delta}}}{\epsilon^2}.
\end{equation}
Thus again, \emph{exponential} suppression of the failure probability on a single point by the number of projections, implies the minimal number of projections $M^*$ required to preserve geometry needs to grow at most \emph{logarithmically} in the volume of a cube with sides of length $L$, or equivalently, linearly in the subspace dimensionality $K$.
Again, any choice of $M$ greater than the upper bound on  $M^*(\epsilon,\delta)$ in \cref{m:eq:JLsubspace} is sufficient to guarantee the geometry preservation condition in \cref{m:eq:failprob} when $\cell$ is a $K$ dimensional linear subspace.
As we will see below, this condition is only a sufficient condition; it is not a necessary condition as the geometry of a subspace can be preserved at the same distortion level and success probability with fewer projections.


\subsection{Strategy for analyzing random projections of smooth random manifolds}\label{m:sec:strat}

In this work, we are interested in the set of all displacement vectors, or chords, between all pairs of points in a random manifold $\CM$:
\begin{equation}\label{m:eq:chordsp}
  \cell = \CM - \CM \equiv \set{ \, \uv = \xv_1 - \xv_2 \in \R^N \, }{ \, \xv_1, \xv_2 \in \CM \, }.
\end{equation}
We equate the notion of preserving the geometry of the manifold $\CM$, to ensuring that all chords are preserved with a small distortion $\epsilon$ and small failure probability $\delta$ in \cref{m:eq:failprob}.
As discussed above, this condition is sufficient to guarantee that many machine learning algorithms, that depend only on pairwise distances, can operate in the compressed $M$ dimensional random subspace almost as well as in the original ambient space $\R^N$.

\begin{figure}[tbp]
  \centering
  \begin{myenuma}
    \item\aligntop{\includegraphics[width=0.7\linewidth]{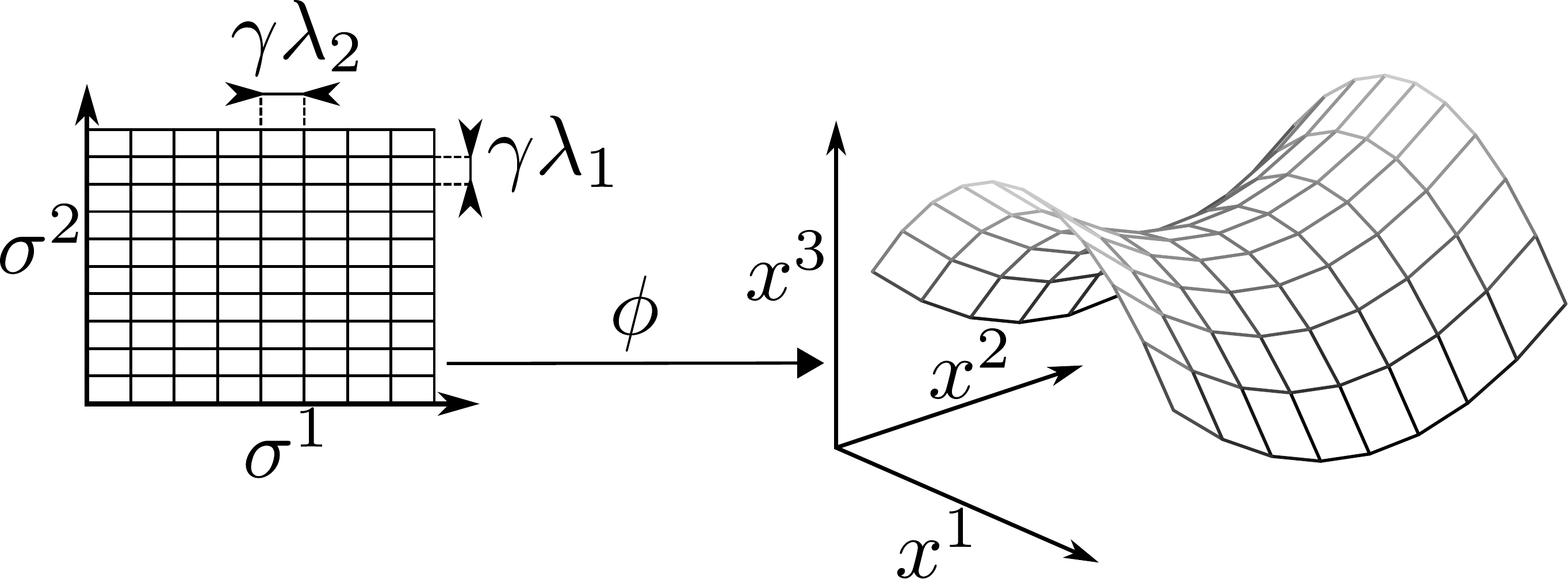}}\label{m:fig:embed}
    \hspace{0.02\linewidth}
    \item\hspace{0.01\linewidth}\aligntop{\includegraphics[width=0.15\linewidth]{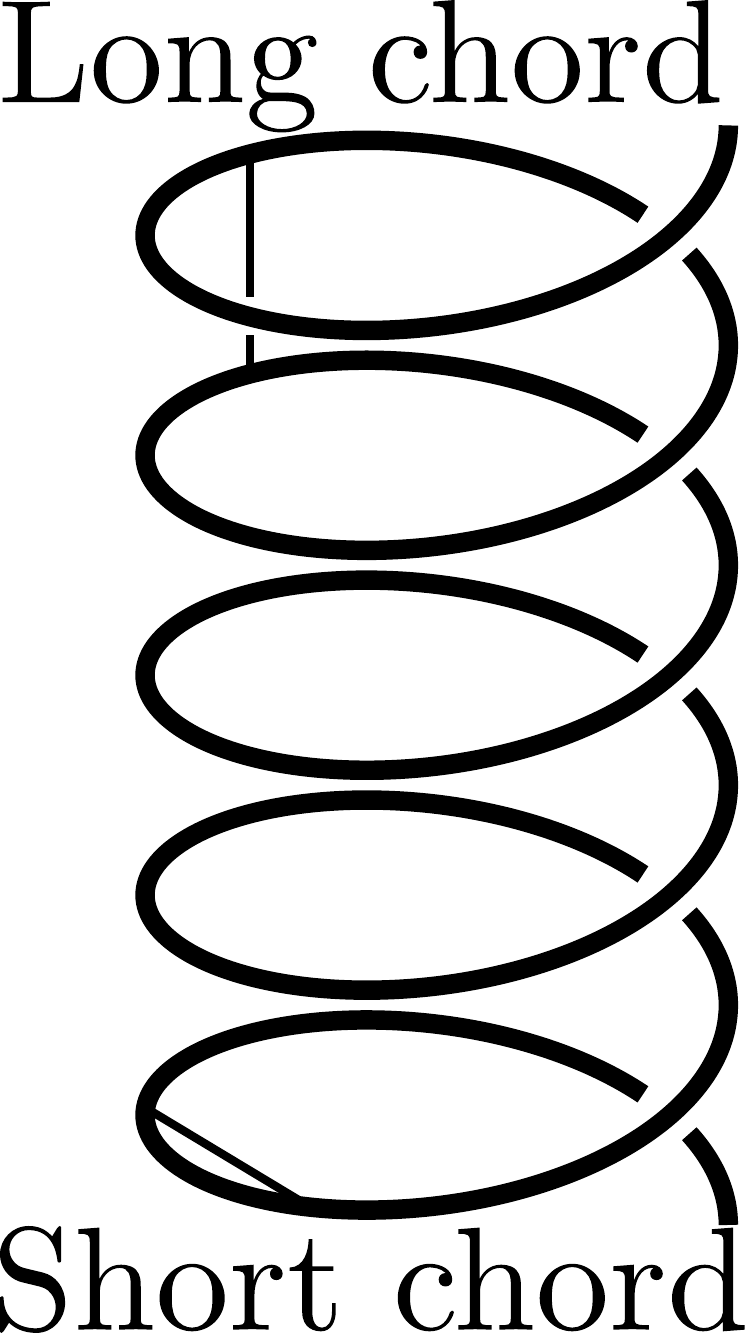}}\label{m:fig:helix}
  \end{myenuma}
  \caption[Schematic description of manifolds, cells and chords]{Schematic description of manifolds, cells and chords.
  (\ref{m:fig:embed}) Embedding map $\phi$ from intrinsic coordinates, $\ic^\alpha$, to extrinsic coordinates, $\ec^i$, with a partition into cells in intrinsic coordinates.
  (\ref{m:fig:helix}) A helix, with examples of long chords, which are not parallel to any tangent vector, and short chords, which are all approximately tangential.
  }\label{m:fig:mfld}
\end{figure}

Because the set of all chords in \cref{m:eq:chordsp} is infinite, one cannot simply use the union bound to bound the failure probability $\delta$ of achieving a given distortion $\epsilon$ as was done in the JL lemma for a point cloud in \cref{m:eq:JLfail}.
However, the probability of failure for different chords are correlated due to the smoothness of $\CM$.
In essence, nearby chords will have similar distortions.
To exploit these correlations to bound the failure probability, we partition the manifold into cells, $\cell_{\vec{m}} \subset \R^N$ (see \cref{m:fig:mfld}\ref{m:fig:embed}):
\begin{equation}\label{m:eq:celldef}
 \cell_{\vec{m}} =  \set{  \ec^i = \phi^i(\ic^\alpha)}{  \abs{ \ic^\alpha - \ic^\alpha_{\vec{m}} } \leq \frac{\clsz}{2}\lambda_\alpha  \,\, \forall \,\, \alpha = 1,\dots,K  },
  \qquad
  \ic^\alpha_{\vec{m}} = \prn{m^\alpha + \frac{1}{2}}\clsz\lambda_\alpha.
\end{equation}
Here  $\vec{m}$ denotes a $K$-tuple of integers indexing the cells $\cell_{\vec{m}}$, $\ic_{\vec{m}}$ is the intrinsic coordinate of the center of a cell, and every cell has a linear extent, in intrinsic coordinates, that is a fraction $\gamma$ of the autocorrelation length $\lambda_\alpha$ in each dimension $\alpha$.
When $\clsz \ll 1$, these cells will be smaller than the typical length scale of curvature of $\CM$.
This means that all chords starting in one cell and ending in another will be approximately parallel in $\R^N$ and so have similar distortion.
Our overall strategy to bound the failure probability in \cref{m:eq:failprob} for all possible chords in \cref{m:eq:chordsp} is to quantify how similar the distortion is for all chords starting in one cell and ending in another, and then apply the union bound to the finite set of all pairs of cells.
We must consider two types of chords: long chords between two different cells, and short chords between two points in the same cell (\cref{m:fig:mfld}\ref{m:fig:helix}).
By the union bound, the total failure probability  is bounded by the sum of the failure probabilities for long and short chords.

For long chords, we must first ensure that the distortion of \emph{all} chords between a given pair of cells is less than $\epsilon$.
In \cref{m:sec:distinter}, we show this condition is \emph{guaranteed} if the distortion of the chord joining their centers is less than a function $\gntc(\epsilon, \thxc)$, defined in \cref{m:eq:distcondlong}, where $\sin\thxc = \cldm/x$, $\cldm$ is the diameter of a ball that contains a cell and $x$ is the length of the chord joining the centers, both measured in the ambient space.
To apply this result, we need to know the typical diameter of cells and the distance between them in random manifolds.
In \cref{m:sec:cellsep}, we show that these quantities take specific values with high probability.
Then, we can bound the probability that \emph{all} long chords beginning in one cell and ending in another have distortion less than $\epsilon$ by the probability that the central chord connecting the two cell centers has distortion less than $\gntc(\epsilon,\thxc)$.
By the union bound, the failure probability for preserving the geometry of all central chords is bounded by the sum of the failure probabilities for each central chord.
In turn the failure probability for each central chord can be computed via the JL lemma (see \cref{m:eq:JLfail} and \cite{Johnson1984extensions,indyk1998approximate,Dasgupta2003JLlemma}).
In \cref{m:sec:long}, we combine these results to bound the failure probability for preserving the length of all long chords
of $\CM$.

For the small values of $\clsz \ll 1$ that we consider (see \cref{s:eq:cellopt}), the short chords beginning and ending in the same cell will all be parallel to some tangent vector of $\CM$ (see \cref{s:eq:shorttangent} in \cref{s:sec:cellcurv}\supp\ for justification).
Thus bounding the distortion of all short chords is equivalent to bounding the distortion of all tangent planes.
Corresponding to each cell, $\cell_{\vec{m}}$, there is a set of subspaces, $\gauss\,\cell_{\vec{m}}$, comprising the tangent planes of $\CM$ at all points in $\cell_{\vec{m}}$.%
\footnote{The set $\gauss\,\cell_{\vec{m}}$ is a subset of the Grassmannian, $\grass_{K,N}$, the set of all $K$ dimensional subspaces of $\R^N$.
The Gauss map, $\gauss$, takes a point on the manifold, $\CM$, to the point in the Grassmannian corresponding to its tangent plane.}
We need to ensure that the distortion of all tangent planes in $\gauss\,\cell_{\vec{m}}$ is less than $\epsilon$.
In \cref{m:sec:distintra}, we show that this is \emph{guaranteed} if the distortion of the central tangent plane $\Usp_{\vec{m}}$ is less than a function $\gnts(\epsilon,\thxs)$, defined in \cref{m:eq:distcondshort}, where $\thxs$ is the largest principal angle between $\Usp_{\vec{m}}$ and any other tangent plane in $\gauss\,\cell_{\vec{m}}$.
In \cref{m:sec:cellcurv} we show that $\thxs$ is bounded by a specific value with high probability in our model of random manifolds $\CM$.
Then, we can bound the probability that \emph{all} short chords beginning and ending in the same cell have a distortion less than $\epsilon$ by the probability that the central tangent plane of the cell has a distortion less than $\gnts(\epsilon,\thxs)$.
By the union bound, the failure probability over all central tangent planes is bounded by the sum of the failure probabilities for each central tangent plane.
In turn, the failure probability for each central tangent plane can be computed via the subspace form of the JL lemma in \cref{m:eq:JLsubspace} and \cite{baraniuk2008simple}.
In \cref{m:sec:short}, we combine these results to bound the failure probability for preserving the length of all short chords of $\CM$.

Finally, the results  for long and short chords will be combined in \cref{m:sec:all}, culminating in \cref{m:eq:numprojprob} where we determine how many dimensions $M$ a random projection requires to ensure, with low failure probability, that the geometric distortion of a random manifold $\CM$ is less than $\epsilon$.


\section{Bounding the distortion of long and short chords}\label{m:sec:boundcell}

Here we begin the strategy outlined in \cref{m:sec:strat}.
In particular, in \cref{m:sec:distinter} we bound the distortion of all long intercellular chords beginning and ending in two different cells in terms of the distortion of the central chord connecting the two cell centers.
Also, in \cref{m:sec:distintra} we bound the distortion of all short intracellular chords within a cell in terms of the distortion of the central tangent plane of the cell.
These bounds also depend on the typical size, separation, and curvature of these cells, but we postpone the calculation of such typical cell geometry in random manifolds to \cref{m:sec:randman}.


\subsection{Distortion of long chords in terms of cell diameter, separation and central chords}\label{m:sec:distinter}

\begin{figure}[tbp]
  \begin{myenuma}
    \item\aligntop{\includegraphics[height=0.32\linewidth]{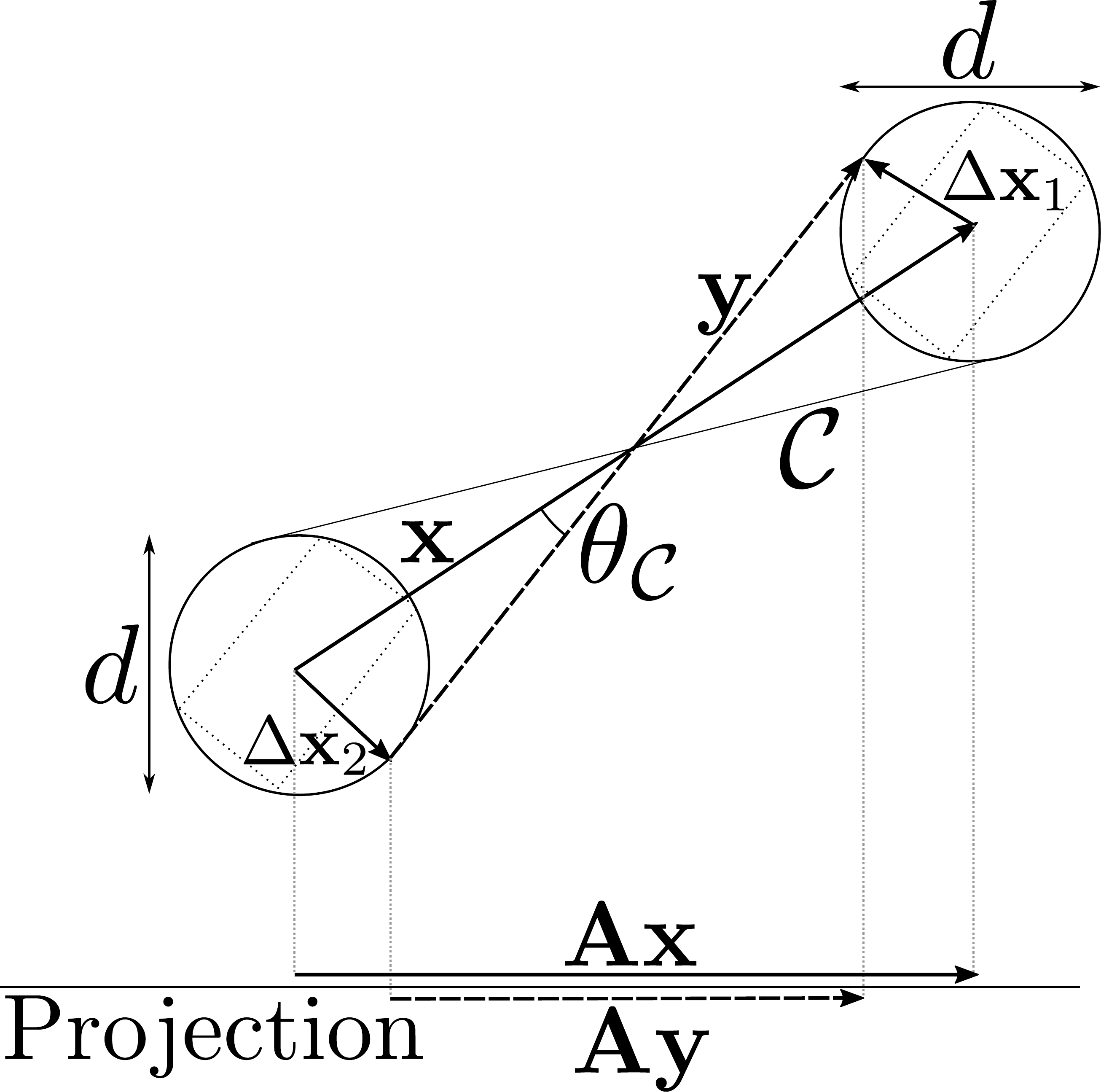}}\label{m:fig:conceptchord}
    \hspace{0.02\linewidth}
    \item\aligntop{\includegraphics[height=0.32\linewidth]{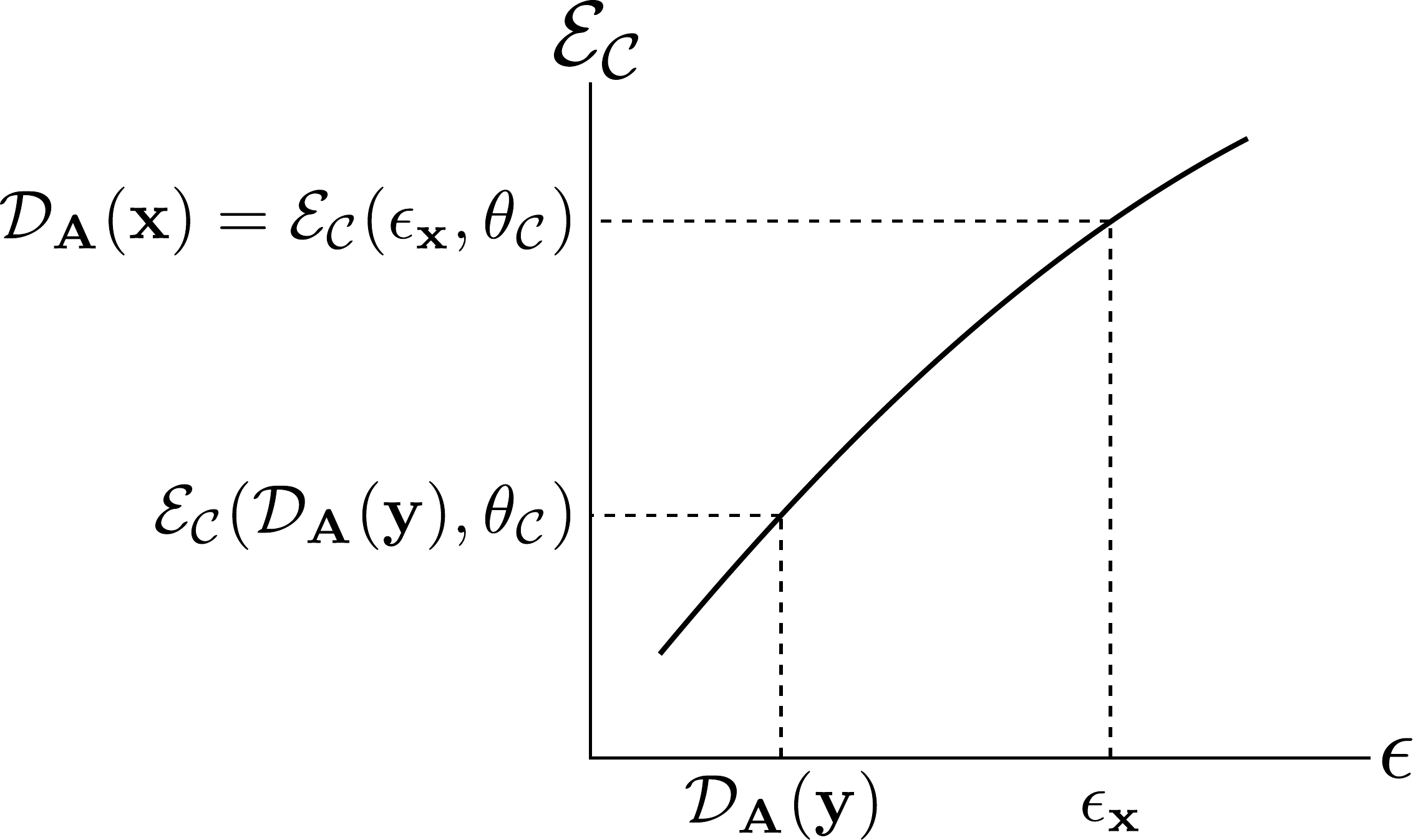}}\label{m:fig:conceptineq}
    \\[2\baselineskip]
    \item\aligntop{\includegraphics[height=0.32\linewidth]{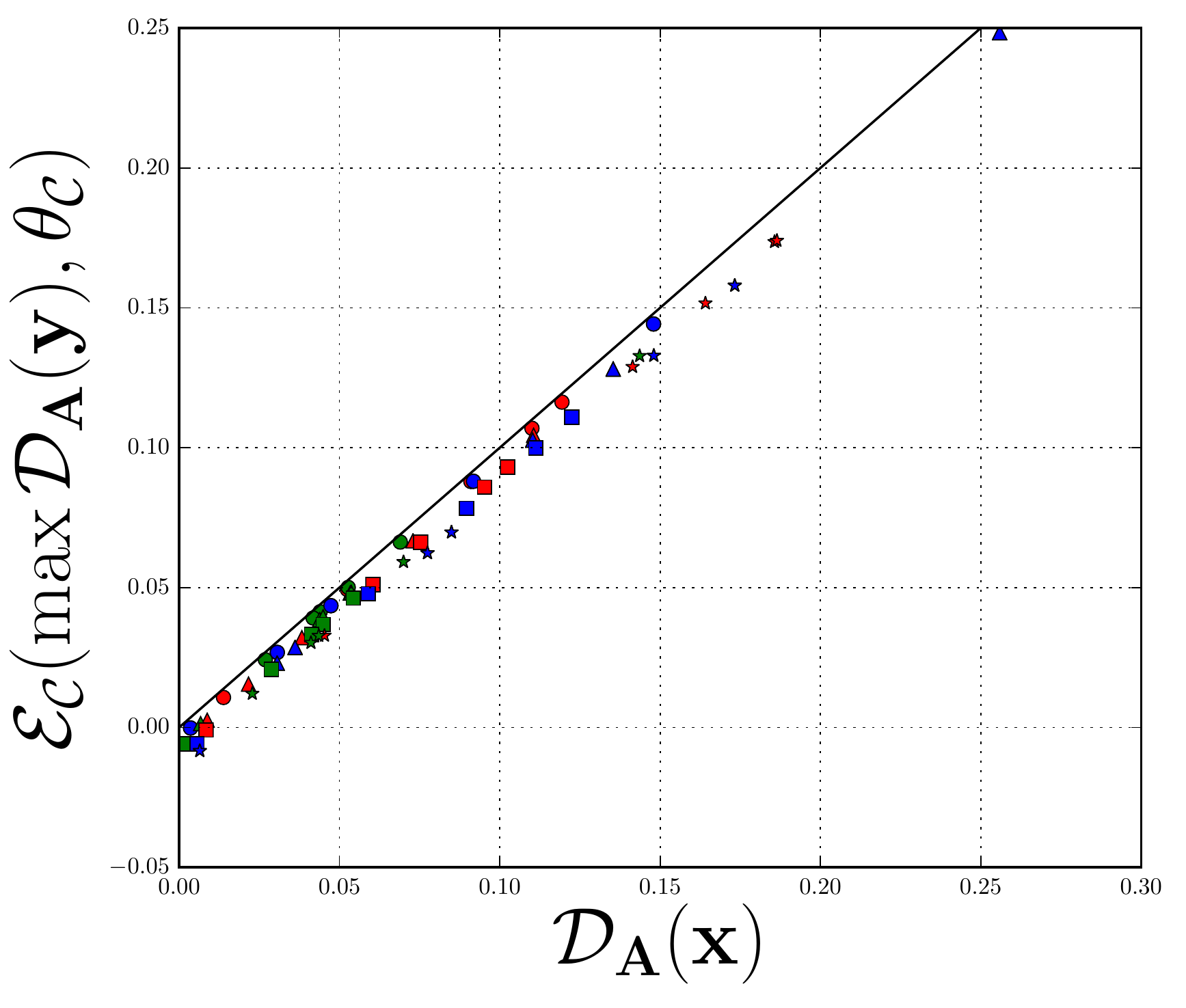}}\label{m:fig:chorddist}
    \item\aligntop{\includegraphics[height=0.32\linewidth]{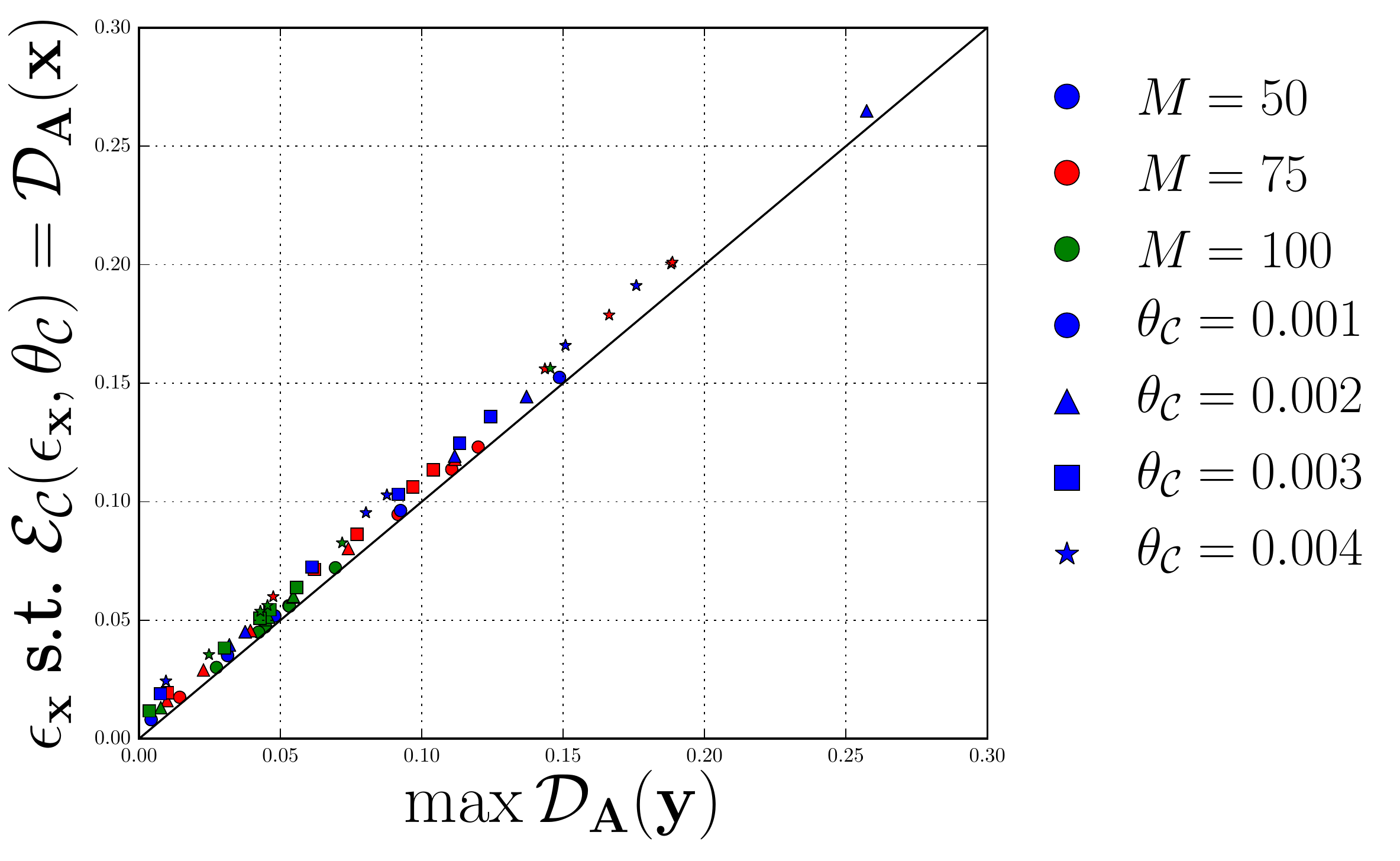}}\label{m:fig:chorddistinv}
  \end{myenuma}
  \caption[Bounding distortion of long chords between two cells]{
    (\ref{m:fig:conceptchord}) Schematic description of the relevant vectors and projections:
    the diameter of the balls around the endpoints of the chord, $\cldm$, is chosen so that they enclose the two cells,
     $\xv$ is the chord between the centers of two cells,
     $\yv$ is another vector between the two cells,
     $\proj\xv$ and $\proj\yv$ are their projections to an $M$ dimensional space.
    (\ref{m:fig:conceptineq}) Schematic illustration of the bounds on $\distpr(\xv)$ and $\distpr(\yv)$.
     If $\gntc(\epsilon_{\xv}, \thxc) = \distpr(\xv)$, then we must have $\distpr(\yv) \leq \epsilon_{\xv}$ for all $\yv \in \conec$.
     Then the monotonicity of $\gntc$ implies that $\gntc(\distpr(\yv), \thxc) \leq \gntc(\epsilon_{\xv}, \thxc) = \distpr(\xv)$.
    (\ref{m:fig:chorddist}) Tests of \cref{m:eq:distcondlong} with randomly generated vectors $\xv$ and $\yv$, with $N=1,000$.
     For each $\xv$, we sample $\yv$ 200,000 times on the boundary of the cone $\conec$ and record only the largest distortion $\distpr(\yv)$.
     The distortion of $\xv$, plotted against the function $\gntc$ in \cref{m:eq:distcondlong} evaluated at the worst-case distortion of $\yv$, its claimed lower bound.
    (\ref{m:fig:chorddistinv}) Same as (\ref{m:fig:chorddist}), except the largest distortion $\distpr(\yv)$ is plotted against $\epsilon_{\xv}$, the solution of $\gntc(\epsilon_{\xv}, \thxc) = \distpr(\xv)$, its claimed upper bound.
    }\label{m:fig:cellchord}
\end{figure}

Consider the set of all long chords between two different cells whose centers are at $\xv_1$ and $\xv_2$ in $\R^N$.
Each cell can be completely contained by a ball of diameter $d$, centered on the cell, where $d$ is the typical cell diameter which we compute in  \cref{m:sec:cellsep}.
The set of chords between these two balls forms a cone that we refer to as the chordal cone, $\conec$.
By construction, $\conec$ contains all long chords between the two cells.
An important measure of the size of this cone $\conec$ is the maximal angle between any chord $\yv \in \conec$ and the central chord $\xv = \xv_1 - \xv_2$.
This maximal angle is achieved by the outermost vectors tangent to the boundaries of each ball (see \cref{m:fig:cellchord}\ref{m:fig:conceptchord}), yielding $\thxc = \sin\inv \frac{\cldm}{x}$, where $x = \nrm{\xv}$ is the typical separation between cells, which we compute in \cref{m:sec:cellsep}.

Now how small must the central distortion $\distpr(\xv)$ be to \emph{guarantee} that $\distpr(\yv) \leq \epsilon$ for \emph{all} chords $\yv$ in $\conec$, and thus all long chords between the two cells?
Call this quantity $\gntc(\epsilon,\thxc)$:
\begin{equation}\label{m:eq:gntcdef}
  \distpr(\xv) \leq \gntc(\epsilon,\thxc)
  \qquad \implies \qquad
  \distpr(\yv) \leq \epsilon \quad \forall \, \yv \, \in \conec.
\end{equation}
As shown in \cref{s:sec:distinter}\supp, in the limit of large $N$ and small $\epsilon$ and $\clsz$, this guarantee holds when the central distortion of $\xv$ under $\proj$ satisfies the bound:
\begin{equation}\label{m:eq:distcondlong}
  \distpr(\xv) \lesssim \epsilon - \sqrt{\frac{N}{M}}\sin\thxc \equiv \gntc(\epsilon,\thxc).
\end{equation}
Intuitively, the larger the cone size $\thxc$, the smaller the central distortion of $\xv$ under $\proj$ must be to guarantee that all chords $\yv \in \conec$ have distortion less than $\epsilon$ under the \emph{same} projection $\proj$.
Conversely, for small cone sizes $\thxc$, all chords $\yv \in \conec$ will be almost parallel to the central chord $\xv$, and therefore have similar distortion to it under any projection $\proj$. Then the distortion of $\xv$ need not be much smaller than $\epsilon$ to guarantee that \cref{m:eq:gntcdef} holds.
We note that to ensure the distortion $\epsilon$ of all chords in the cone is not much larger than the central distortion $\gntc$, the cone size $\thxc$ must be small.
Indeed $\thxc$ must be $\CO(\sqrt{M/N}\epsilon)$, and this is the regime of chordal cone size we will employ below.

We can argue that $\gntc(\distpr(\yv), \thxc)$ is a lower bound on $\distpr(\xv)$ for all $\yv \in \conec$ as follows.
Consider an $\epsilon_{\xv}$ that obeys $\gntc(\epsilon_{\xv}, \thxc) = \distpr(\xv)$.  Then by \cref{m:eq:gntcdef}, we must have $\distpr(\yv) \leq \epsilon_{\xv}$ for all $\yv \in \conec$.
Then the monotonicity of $\gntc$ implies that $\gntc(\distpr(\yv), \thxc) \leq \gntc(\epsilon_{\xv}, \thxc) = \distpr(\xv)$.
This situation is depicted in \cref{m:fig:cellchord}\ref{m:fig:conceptineq}.
This suggests that we can test \cref{m:eq:distcondlong} by randomly generating a vector $\xv$ and a projection $\proj$, and then randomly generating many vectors $\yv$ in the cone $\conec$.
We can then find the $\yv$ with the largest distortion and verify that $\distpr(\xv) \geq \gntc(\distpr(\yv), \thxc)$.
Alternatively, we can solve $\gntc(\epsilon_{\xv}, \thxc) = \distpr(\xv)$ for $\epsilon_{\xv}$ and then verify that $\distpr(\yv) \leq \epsilon_{\xv}$.
These tests can be seen in \cref{m:fig:cellchord}\ref{m:fig:chorddist}\ref{m:fig:chorddistinv}, where we compare $\distpr(\xv)$ to $\gntc(\distpr(\yv), \thxc)$ and $\epsilon_{\xv}$ to $\distpr(\yv)$, having chosen the $\yv$ with the largest distortion.
We see that the bounds were satisfied using the expression in \cref{m:eq:distcondlong}, and the bound is tighter for smaller $\thxc$, which is exactly the regime in which we will employ this bound below.


\subsection{Distortion of short chords in terms of cell curvature and central tangent planes}\label{m:sec:distintra}

\begin{figure}[tbp]
  \centering
  \begin{myenuma}
    \item\aligntop{\includegraphics[height=0.3\linewidth]{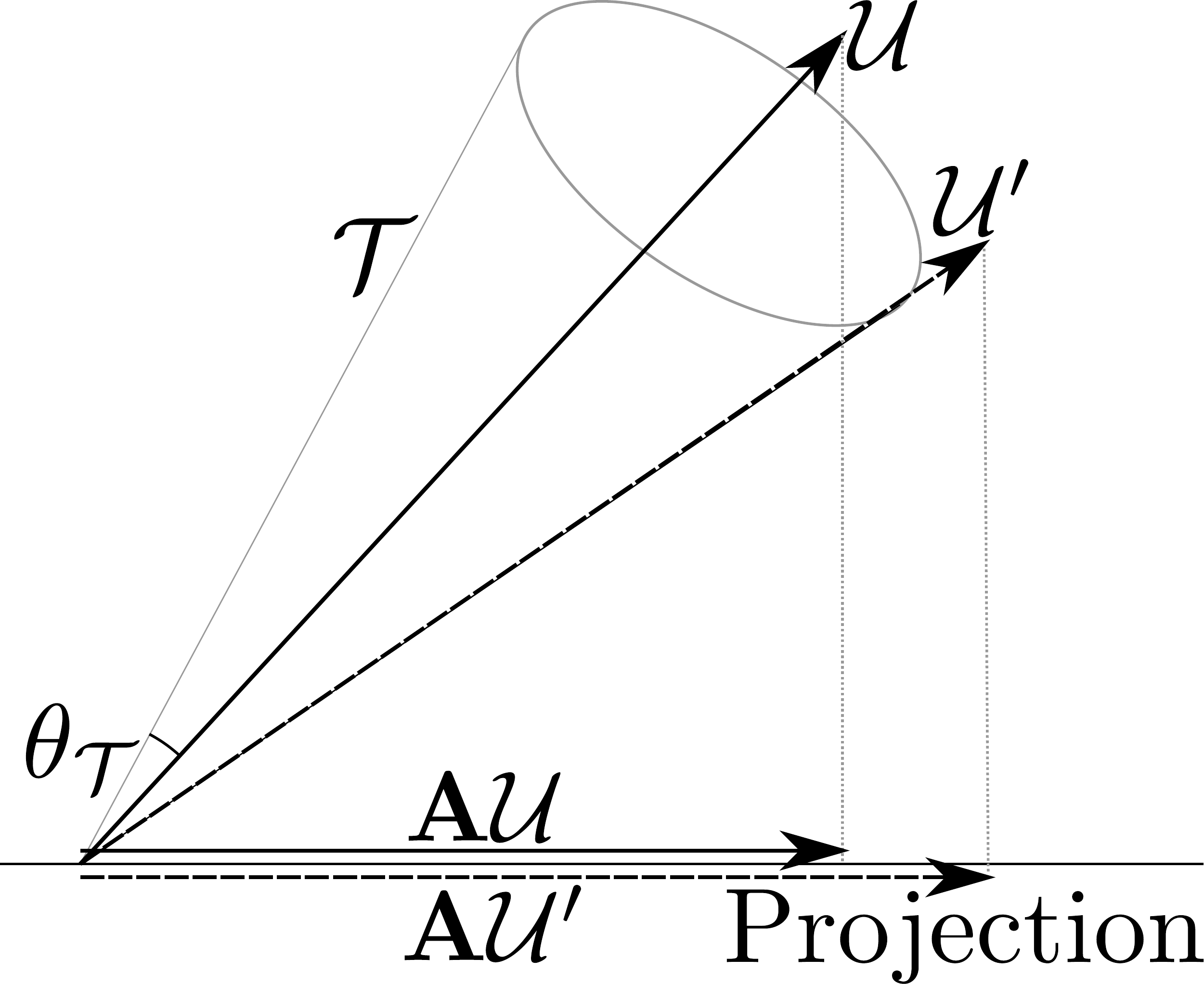}}\label{m:fig:concepttang}
    \hspace{0.02\linewidth}
    \item\aligntop{\includegraphics[height=0.3\linewidth]{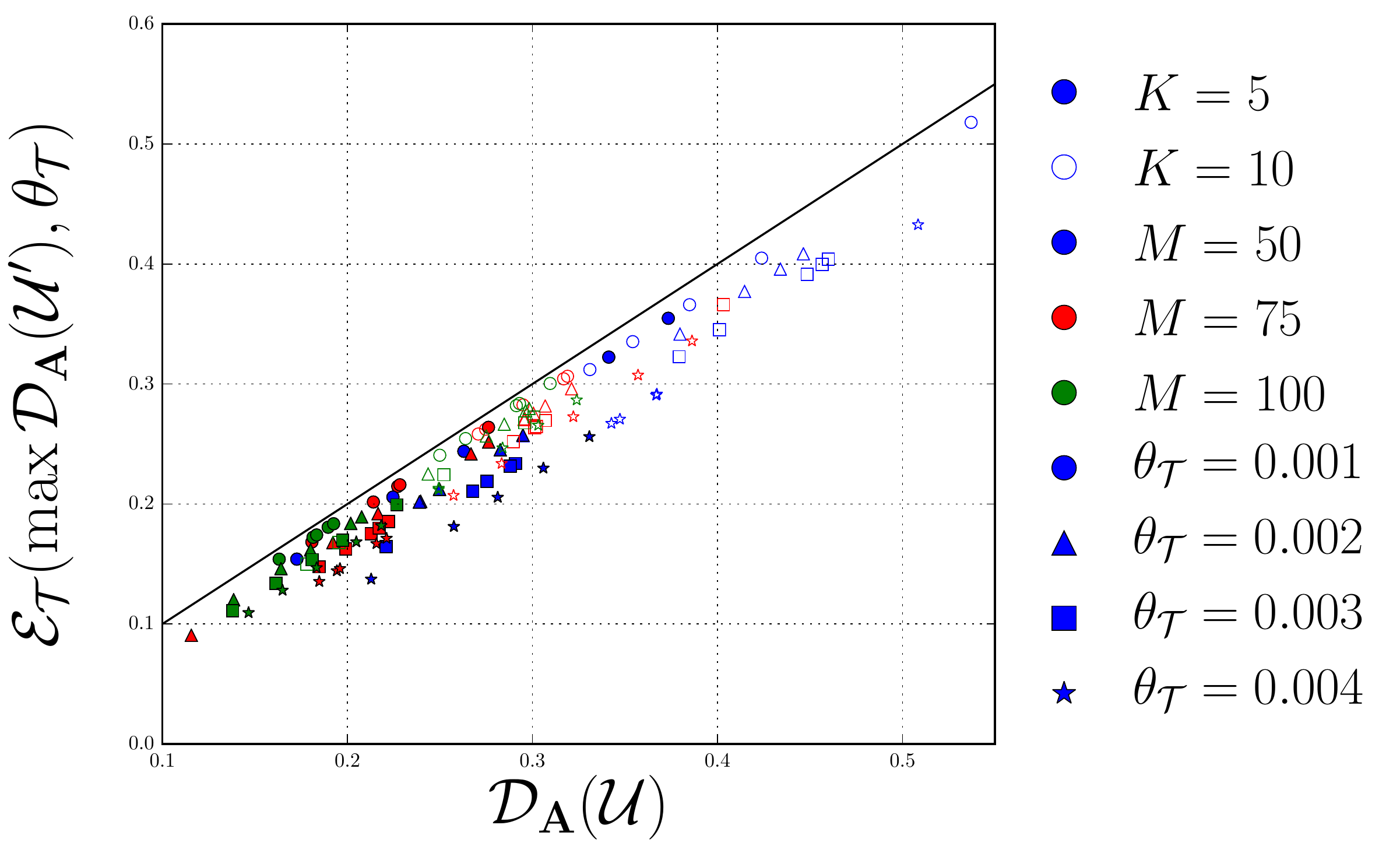}}\label{m:fig:tangdist}
  \end{myenuma}
  \caption[Bounding distortion of short chords within cells]{
      (\ref{m:fig:concepttang}) Schematic description of the relevant vector spaces and projections:
    $\thxs$ is the largest principal angle between the tangent plane at the cell center, $\Usp$, and the subspaces at the edges of the cone, which encloses the tangent planes of all points in the cell.
    $\Usp'$ is the tangent plane at another point in the cell,
    $\proj\Usp$ and $\proj\Usp'$ are the projections of the subspaces to an $M$ dimensional space.
    (\ref{m:fig:tangdist}) Tests of \cref{m:eq:distcondshort} with randomly generated $K$ dimensional subspaces $\Usp,\Usp'$, with $N=1,000$.
    We sample $\Usp'$ 200,000 times on the boundary of the cone, $\cones$, and use the one with the largest distortion.
    The distortion of $\Usp$, plotted against the function $\gnts$ in \cref{m:eq:distcondshort} evaluated at the worst-case distortion of $\Usp'$, its claimed lower bound.
    }\label{m:fig:celltang}
\end{figure}

As discussed in \cref{m:sec:strat}, for small cells with $\clsz \ll 1$, all short chords within a single cell will be parallel to a tangent vector at some point in the cell (see \cref{s:eq:shorttangent} in \cref{s:sec:cellcurv}\supp\ for justification).
Thus to bound the distortion of all short chords, we focus on bounding the distortion of all tangent planes in a cell.
Let $\Usp \in \grass_{K,N}$ be the tangent plane at a cell center.
Also, assume all tangent planes $\Usp'$ at all other points in the cell have principal angles with $\Usp$ that satisfy $\theta_a \leq \thxs$ (see \cref{s:sec:angles}\supp\ for a review of principal angles between subspaces).
This maximal principal angle $\thxs$ depends on the size and curvature of cells, which we compute in \cref{m:sec:cellcurv}.
The set of all subspaces in $\grass_{K,N}$ with all principal angles $\theta_a \leq \thxs$ forms a ``cone'' of subspaces that we will refer to as the tangential cone, $\cones$ (see \cref{m:fig:celltang}\ref{m:fig:concepttang} for a schematic).

Now how small must the distortion of the central subspace $\distpr(\Usp)$ be to \emph{guarantee} that $\distpr(\Usp') \leq \epsilon$ for \emph{all} subspaces $\Usp'$ in $\cones$ (and thus all chords within the cell)?
Call this quantity $\gnts(\epsilon,\thxs)$.
\begin{equation}\label{m:eq:gntsdef}
  \distpr(\Usp) \leq \gnts(\epsilon,\thxs)
  \qquad \implies \qquad
  \distpr(\Usp') \leq \epsilon \quad \forall \,\,  \Usp' \in \cones.
\end{equation}
As shown in \cref{s:sec:distintra}\supp, in the limit of large $N$ and small $\epsilon$ and $\clsz$, this guarantee is valid when the distortion of $\Usp$ satisfies the bound:
\begin{equation}\label{m:eq:distcondshort}
   \distpr(\Usp) \lesssim \epsilon - \frac{N}{M} \, \sin\thxs \equiv \gnts(\epsilon,\thxs).
\end{equation}
Intuitively, when $\thxs$ is large, $\Usp'$ can lie in very different directions to $\Usp$, so the distortion of $\Usp$ needs to be made very small to ensure that the distortion of $\Usp'$ lies within the limit in \cref{m:eq:gntsdef}.
When $\thxs$ is small, $\Usp'$ must lie in similar directions to $\Usp$, so the distortion of $\Usp$ does not need to be much smaller than $\epsilon$, as $\Usp'$ will be almost parallel to it and will therefore have a similar distortion.
We note that to ensure the distortion $\epsilon$ of all tangent planes in the tangential cone is not much larger than the central distortion $\gnts$, the cone size $\thxs$ must be small.
Indeed $\thxs$ must be $\CO(M\epsilon/N)$, and this is the regime of tangential cone size we will employ below.

For the same reason as for the long chords in \cref{m:sec:distinter}, $\gnts(\Usp', \thxs)$ should be a lower bound for $\distpr(\Usp)$ for all $\Usp' \in \cones$ (see \cref{m:fig:cellchord}\ref{m:fig:conceptineq}).
This means that we can test \cref{m:eq:distcondshort} by randomly generating a subspace $\Usp$ and a projection $\proj$, and then randomly generating many subspaces $\Usp'$ in the cone $\cones$.
We can then find the $\Usp'$ with the largest distortion and verify that $\distpr(\Usp) \geq \gnts(\distpr(\Usp'), \thxs)$.
These tests can be seen in \cref{m:fig:celltang}\ref{m:fig:tangdist}, where we compare $\distpr(\Usp)$ to $\gnts(\Usp', \thxc)$.
We see that this bound was satisfied with the expression in \cref{m:eq:distcondshort}, and the bound is tighter for smaller $\thxs$, which is exactly the regime in which we will employ this bound below.


\section{The typical geometry of smooth random Gaussian manifolds}\label{m:sec:randman}

Here we compute several geometric properties of cells that were needed in \cref{m:sec:boundcell}, namely
their diameter, the distance between their centers, and the maximum principal angle between their tangent planes at cell centers and all other tangent planes in the same cell.
In particular, we compute their typical values in the ensemble of random manifolds defined in \cref{m:sec:randmanmod}.


\subsection{Separation and size of cells}\label{m:sec:cellsep}

\begin{figure}[tbp]
  \centering
  \begin{myenuma}
    \item\aligntop{\includegraphics[width=0.44\linewidth]{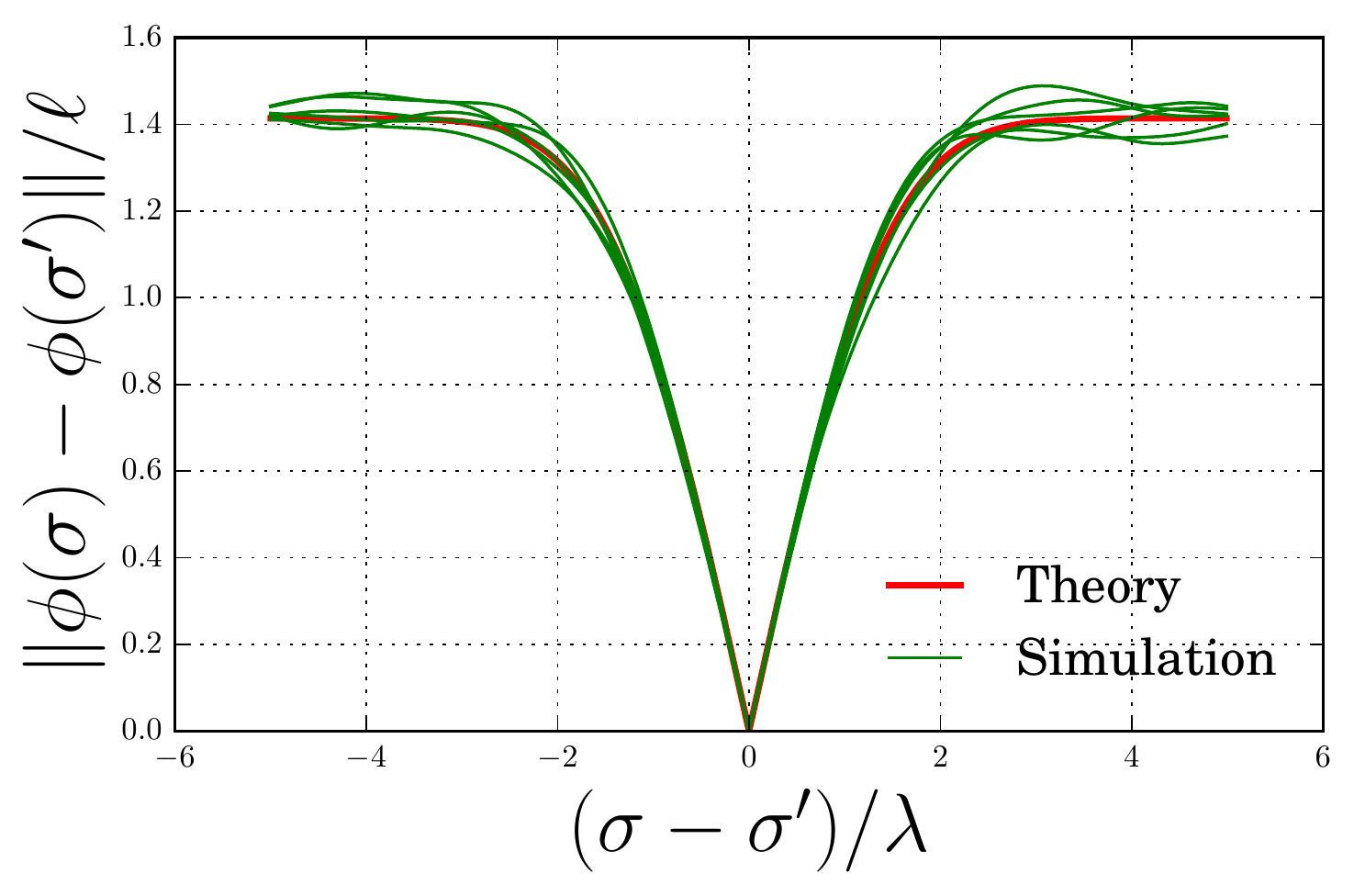}}\label{m:fig:curvedist}
    \hspace{0.02\linewidth}
    \item\aligntop{\includegraphics[width=0.44\linewidth]{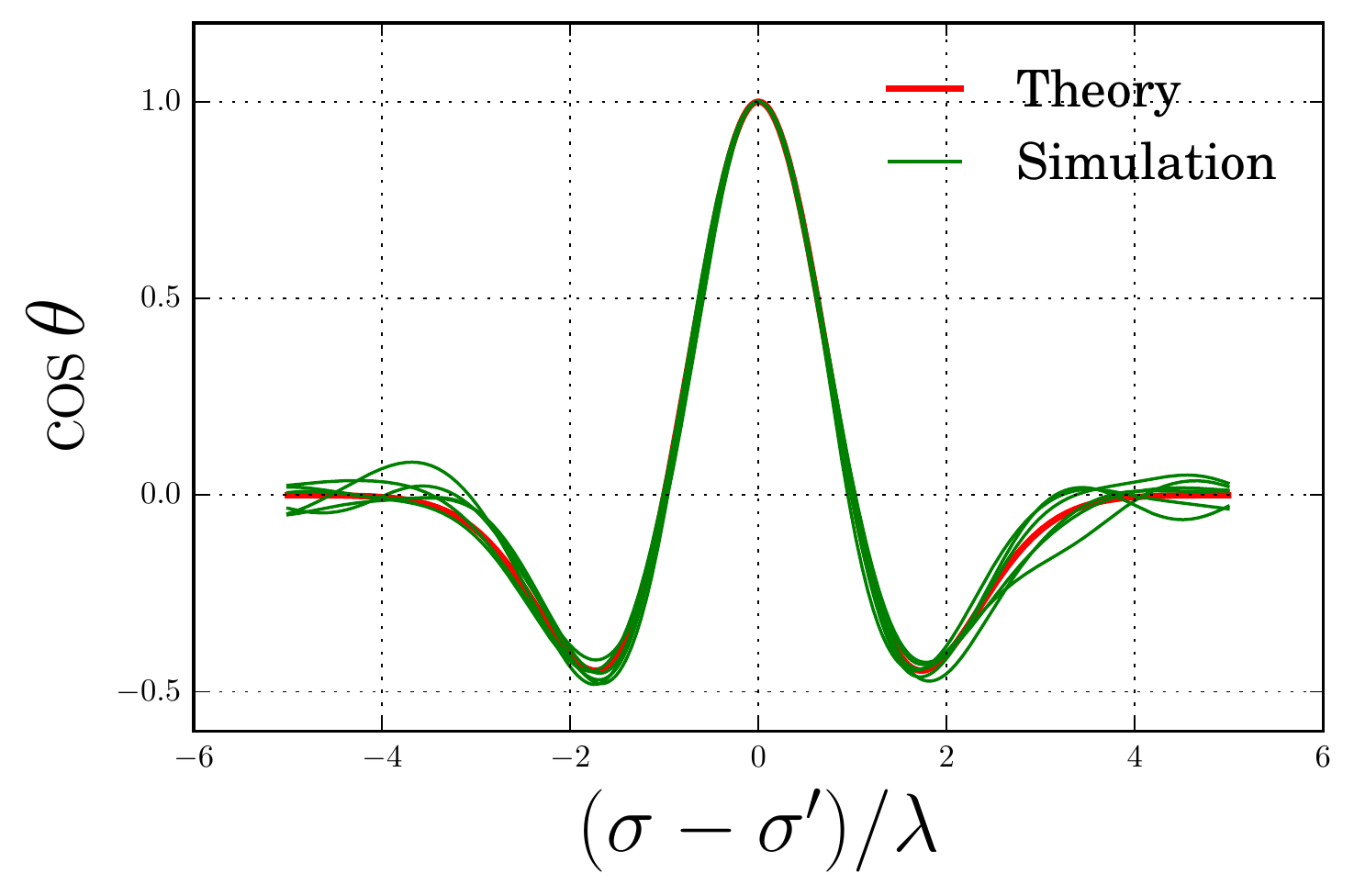}}\label{m:fig:curveang}
  \end{myenuma}
  \caption[Testing formulae for random curves]{
   Testing \cref{m:eq:gaussgausschord} and \cref{m:eq:curvecos} with random 1-dimensional curves generated by sampling from the Gaussian processes in \cref{m:eq:randmangauss,m:eq:gaussgausskernel} for 1024 evenly spaced values of $\ic$, with $N=1000$, $L = 10$ and $\lambda = 1$.
   We compute (\ref{m:fig:curvedist}) the Euclidean distance and (\ref{m:fig:curveang}) the inner product of unit tangent vectors, between all sampled points on the curve and the central point.
  }\label{m:fig:randcurve}
\end{figure}

To compute the cell diameter $d$ and cell separation $x$, which determine the size $\thxc$ of a chordal cone $\conec$
in \cref{m:sec:distinter}, we first compute the Euclidean distance between two points on the manifold.
Detailed calculations can be found in \cref{s:sec:cellsep}\supp.
There, we work in the limit of large $N$, so that sums over ambient coordinates $i=1,\dots,N$ are self-averaging, and can be replaced by their expectations, as the size of typical fluctuations is $\CO(1/N)$ and the probability of large $\CO(1)$ deviations is exponentially suppressed in $N$.
Thus by working in the large $N$ limit, we can neglect fluctuations in the geometry of the random manifold.
In this limit, for random manifolds described by \cref{m:eq:randmangauss} and \cref{m:eq:gaussgausskernel}, the squared distance between two points on the manifold has expected value
\begin{equation}\label{m:eq:gaussgausschord}
  \nrm{\xv_1 - \xv_2}^2
     = \sum_i\brk{\phi^i(\ic_1) - \phi^i(\ic_2)}^2
     = 2\ell^2 \prn{1 - \e^{-\frac{\rho}{2}}}.
\end{equation}
\cref{m:fig:randcurve}\ref{m:fig:curvedist} and \cref{m:fig:randsurf}\ref{m:fig:surfdist} reveal that this formula matches well with numerical simulations of randomly generated manifolds for $K=1$ and $K=2$ respectively, especially at small separations $\rho$.

Given the definition of cells in  \cref{m:eq:celldef}, \cref{m:eq:gaussgausschord} predicts the mean distance between two cell centers to be,
\begin{equation*}
  x_{\vec{m}\vec{n}}^2 = \nrm{\xv_{\vec{m}} - \xv_{\vec{n}}}^2 = 2 \ell ^2 \prn{1 - \e^{-\frac{\clsz^2 \nrm{\vec{m} - \vec{n}}^2}{2}}}.
\end{equation*}%
This distance increases with $\vec{m} - \vec{n}$, but saturates, since at large $N$, since $\nrm{\boldsymbol{\phi(\ic)}} = \ell$ with high probability, effectively confining $\CM$ to a sphere, thereby bounding the distance between cells.

\Cref{m:eq:gaussgausschord} also predicts the diameter, or twice the distance from the cell center to a corner, to be
\begin{equation*}
  {\cldm} = 2\sqrt{2} \ell \prn{1 - \e^{-\frac{\clsz^2 K}{8}}}^{\frac{1}{2}} \approx {\clsz\,\ell\sqrt{K}} \quad \text{for} \quad  \clsz \ll 1.
\end{equation*}%
Thus the maximum angle between the central chord and any other chord between the two cells obeys
\begin{equation}\label{m:eq:cellchang}
  \sin\thxc = \frac{\cldm}{x} \approx \clsz \sqrt{\frac{K/2}{1 - \e^{-\frac{\clsz^2 \nrm{\vec{m} - \vec{n}}^2}{2}}}}.
\end{equation}

\subsection{Curvature of cells}\label{m:sec:cellcurv}
The curvature of a cell is related to how quickly tangent planes rotate in $\R^N$ as one moves across the cell.
Here we compute the principal angles between tangent planes belonging to the same cell.
This calculation allows us to find an upper bound on $\thxs$, the largest principal angle between the tangent plane at the center of a cell, $\Usp_{\vec{m}}$, and any other tangent plane in the same cell.
 $\thxs$ determines the size of the ``tangential cone''  $\cones$ that appears in  \cref{m:eq:distcondshort} in \cref{m:sec:distintra}. Detailed calculations can be found in \cref{s:sec:cellcurv}\supp.
There, as we described in \cref{m:sec:cellsep}, we work in the large $N$ limit so we can neglect fluctuations in the curvature, as typical curvature values concentrate about their mean.

\begin{figure}[tbp]
  \raggedright
  \begin{myenuma}
    \item\aligntop{\includegraphics[width=0.96\linewidth]{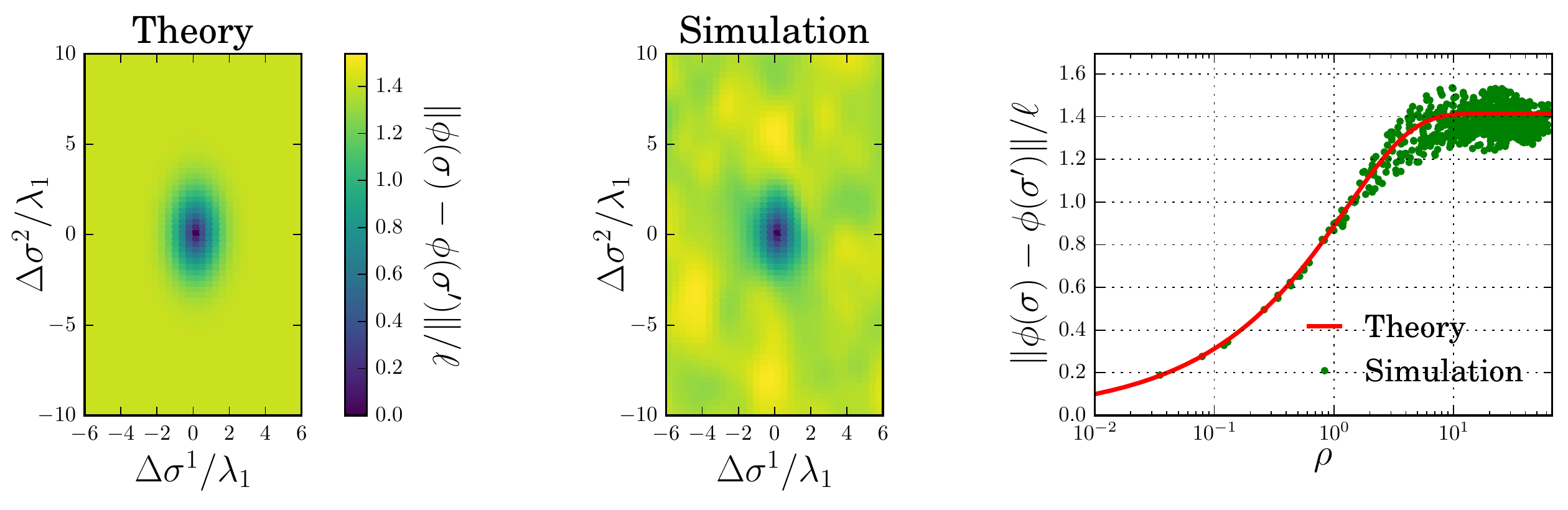}}\label{m:fig:surfdist}

    \item\aligntop{\includegraphics[width=0.96\linewidth]{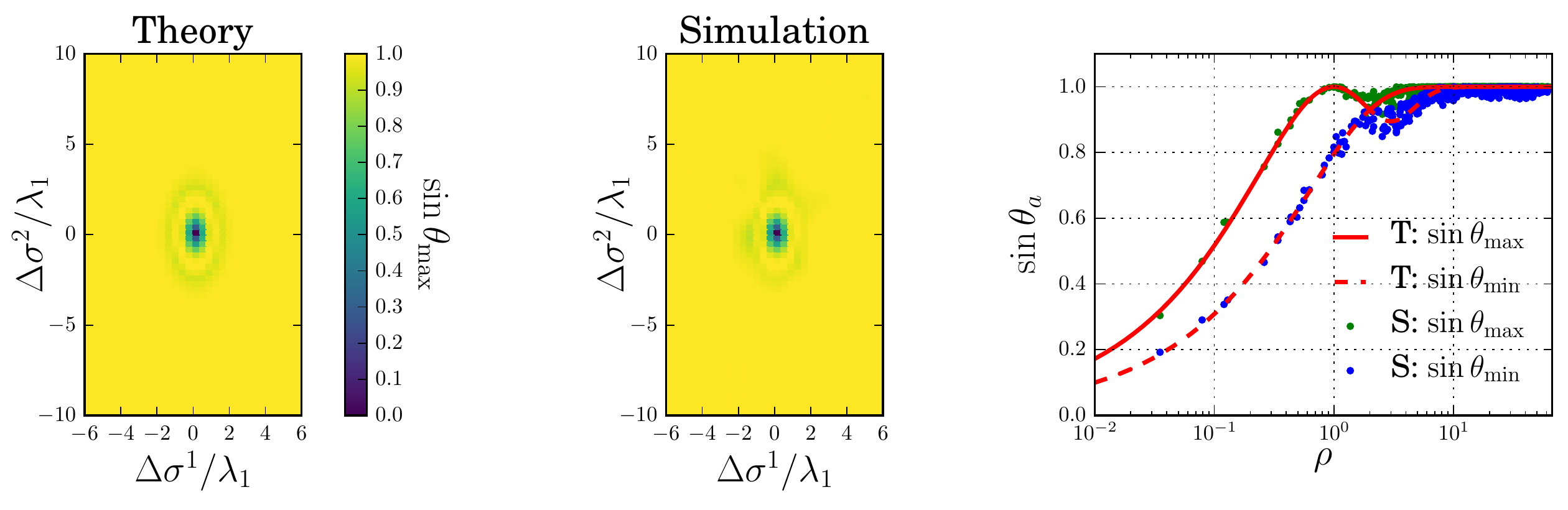}}\label{m:fig:surfang}
  \end{myenuma}
  \caption[Testing formulae for random surfaces]{
  Testing \cref{m:eq:gaussgausschord} and \cref{m:eq:gaussgausspang} with random 2-dimensional surfaces  generated by sampling from the Gaussian processes in \cref{m:eq:randmangauss,m:eq:gaussgausskernel} for $(128, 256)$ evenly spaced values of $\ic^1$ and $\ic^2$ respectively, with $N=200$, $L_\alpha = (12, 20)$ and $\lambda_\alpha = (1, 1.8)$.
  We compute (\ref{m:fig:surfdist}) the Euclidean distance and (\ref{m:fig:surfang}) the principal angles between tangent planes, relative to the central point.
  In the rightmost panels, the red lines indicate our theoretical predictions and green/blue points indicate the results of simulations.
  In (\ref{m:fig:surfang}, right), the solid line and green points are for the larger of the two principal angles and the dashed line and blue points are for the smaller principal angle.
  }\label{m:fig:randsurf}
\end{figure}

For the random manifold ensemble in \cref{m:eq:gaussgausskernel}, the expected cosines of the principal angles $\theta_a$ for $a=1,\dots,K$ between two tangent planes at intrinsic separation $\rho$ is
\begin{equation}\label{m:eq:gaussgausspang}
  \cos\theta_a = \e^{-\rho/2} \quad \text{for} \quad a < K,
  \qquad
  \cos\theta_K = \abs{1 - \rho} \e^{-\rho/2}.
\end{equation}
For $\rho \leq 2$, $\theta_K$ is the largest principal angle.
For $\rho > 2$, the other $\theta_a$ are the largest.
Tests of this formula can be found for $K=2$ in \cref{m:fig:randsurf}\ref{m:fig:surfang}, where we see that the relation is a good approximation, especially at small $\rho$, which is exactly the regime in which we will employ this approximation below.

For the $K=1$ case, \ie 1-dimensional curves, we can also keep track of the orientation of the tangent vectors and distinguish angles $\theta$ and $\pi - \theta$.
This allows us to keep track of the sign of $\cos\theta$.
The cosine of the angle between tangent vectors at two points on the curve is given by
\begin{equation}\label{m:eq:curvecos}
  \cos\theta
    = \frac{\dot{\boldsymbol{\phi}}(\sigma) \cdt \dot{\boldsymbol{\phi}}(\sigma')}{\nrm{\dot{\boldsymbol{\phi}}(\sigma)} \nrm{\dot{\boldsymbol{\phi}}(\sigma')}}
    = (1 - \rho) \, \e^{-\rho/2}.
\end{equation}
where dots indicate derivatives with respect to the intrinsic coordinate.
Tests of this formula can be found in \cref{m:fig:randcurve}\ref{m:fig:curveang}, where we again see that it is a good approximation, especially for small $\rho$.

The largest possible principal angle $\thxs$ between the central plane $\Usp_{\vec{m}}$ and any other plane $\Usp'$ in the cell occurs when $\Usp'$ is at one of the cell corners.
Evaluating \cref{m:eq:gaussgausspang} at the corner yields, for $\gamma \ll 1/\sqrt{K}$,
\begin{equation}\label{m:eq:cellapprox}
  \sin\thxs \approx \frac{\clsz}{2} \, \sqrt{3K}.
\end{equation}

\section{Putting it all together}\label{m:sec:logic}

In \cref{m:sec:boundcell} we saw how to limit the distortion of all chords by limiting the distortion of chords related to the centers of cells.
In \cref{m:sec:randman} we found the typical size, separation and curvature of these cells, which are needed as input for the distortion limits.
In this section we combine these results to find an upper limit on the probability that any chord has distortion greater than $\epsilon$ under a random projection $\proj$.

We do this in two steps.
First, in \cref{m:sec:long}, we will bound this probability for the long chords between different cells.
Then, in \cref{m:sec:short} we will do the same for the short chords within a single cell.
Finally, in \cref{m:sec:all} we will combine these two results with the union bound to find our upper limit on the probability of a random projection causing distortion greater than $\epsilon$ for any chord of the submanifold $\CM$.


\subsection{Long chords}\label{m:sec:long}

Here we combine the results of \cref{m:sec:distinter} and \cref{m:sec:cellsep} to find an upper bound on the probability that any long intercellular chord has distortion greater than $\epsilon$ under a random projection $\proj$.
The detailed calculations behind these results can be found in \cref{s:sec:long}\supp.

The manifold $\CM$ is partitioned into cells $\cell_{\vec{m}}$ with centers $\ic_{\vec{m}}$, as described in \cref{m:eq:celldef}.
By using the union bound, we can write the failure probability for the maximum distortion of all long chords as:
\begin{equation*}
\begin{aligned}
  \delta\lng
     &
     = \Pr\brk{\distpr([\CM-\CM]\lng) > \epsilon} \\
     &= \Pr\brk{\bigcup_{\vec{m}, \vec{n}}\brc{\distpr(\cell_{\vec{m}} - \cell_{\vec{n}}) > \epsilon}}
     \\ &
     \leq \sum_{\vec{m}, \vec{n}} \Pr\brk{\distpr(\cell_{\vec{m}} - \cell_{\vec{n}}) > \epsilon} \\
     &\leq \sum_{\vec{m}, \vec{n}} \Pr\brk{\distpr(\xv_{\vec{m}} - \xv_{\vec{n}}) > \gntc(\epsilon,\thxc)},
\end{aligned}
\end{equation*}%
where $\distpr(\cell - \cell')$ is the maximum distortion over all chords between the sets $\cell,\cell'$.
The second line follows from the definition of the long chords that appear in the first line.
The inequality in the third line comes from the union bound.
The last inequality is a result of the definition of $\gntc(\epsilon,\thxc)$ in \cref{m:eq:gntcdef}, \cref{m:sec:distinter}.
In essence, the contra-positive of \cref{m:eq:gntcdef} states that,
if the distortion of any chord in $\cell_{\vec{m}} - \cell_{\vec{n}}$ is greater than $\epsilon$,
then the distortion of the central chord must be greater than $\gntc(\epsilon,\thxc)$.
This implies that the former event is a subset of the latter event, and hence the probability of the former cannot exceed the probability of the latter.

When $N \gg K^2$, $M\epsilon^2 \gg 1$, $K \gg 1$, and $L_\alpha \gg \lambda_\alpha$ this sum can be approximated with an integral that can be performed using the saddle point method (see \cref{s:eq:alllong,s:eq:longintegral,s:eq:saddledef}, \cref{s:sec:long}\supp).
Combining \cref{m:eq:distcondlong,m:eq:cellchang} with the JL lemma (see \cref{m:eq:JLsingle} and \cite{Johnson1984extensions,indyk1998approximate,Dasgupta2003JLlemma}), leads to
\begin{equation*}
\begin{aligned}
     \delta\lng
      &
     \lesssim \frac{\pi^{\frac{K}{2}} \clsz^{-2K} \V}{\Gamma(\frac{K}{2})} \exp\prn{-\min_\rho \brc{\frac{M}{4}\brk{\epsilon - \clsz\sqrt{\frac{KN}{2M\prn{1 - \e^{-\frac{\rho}{2}}}}}}^2 - \frac{K}{2}\ln\rho}},
\end{aligned}
\end{equation*}%
where $\V = \prod_\alpha \frac{L_\alpha}{\lambda_\alpha}$ (see \cref{s:sec:long}\supp\ for the derivation).

Minimizing with respect to $\clsz$ to obtain the tightest possible bound, we find that
$\clsz^*_\conec \sim \CO(\epsilon\sqrt{M/KN})$, $\sin\thxc^* \sim \CO(\epsilon\sqrt{M/N})$, and $\rho^* \sim \CO(1)$ (see \cref{s:eq:saddle}\supp), and:
\begin{equation}\label{m:eq:distproblong}
  \delta\lng \lesssim \exp \prn{ - \frac{M\epsilon^2}{4}
             + \ln\V  + K \ln \prn{\frac{N M \epsilon^2}{K}}
             + C_0
             - \ln\Gamma\!\prn{\frac{K}{2}}
             },
\end{equation}
where $C_0 = -0.098$.

We note that the optimal value of $\clsz$ and $\thxc$ turn out to be small in the limit $N \gg M$, $K \gg 1 \gg \epsilon$,
and thus satisfy the requirements of the approximations used to derive \cref{m:eq:distcondlong} and \cref{m:eq:cellchang}, as promised.
Moreover, for these small values of $\clsz$ and $\thxc$, the bound demonstrated in \cref{m:eq:distcondlong}, and depicted in \cref{m:fig:cellchord}\ref{m:fig:chorddist}\ref{m:fig:chorddistinv}, is tight.


\subsection{Short chords}\label{m:sec:short}

Here we combine the results of \cref{m:sec:distintra} and \cref{m:sec:cellcurv} to find an upper bound on the probability that any short intracellular chord has distortion greater than $\epsilon$ under a random projection $\proj$.
The detailed calculations behind these results can be found in \cref{s:sec:short}\supp.

By combining the results of \cref{m:sec:distintra}, the union bound, and translational invariance,
we find that:
\begin{equation*}
\begin{aligned}
  \delta\shrt
    &= \Pr\brk{\distpr([\CM-\CM]\shrt) > \epsilon} \\
    &= \Pr\brk{\bigcup_{\vec{m}} \brc{\distpr(\gauss\cell_{\vec{m}}) > \epsilon}} \\
    &\leq \sum_{\vec{m}} \Pr\brk{\distpr(\gauss\,\cell_{\vec{m}}) > \epsilon} \\
    &=  \frac{\V}{\clsz^{K}} \, \Pr\brk{\distpr(\gauss\,\cell_1) > \epsilon} \\
    &\leq \frac{\V}{\clsz^{K}} \, \Pr\brk{\distpr(\Usp_1) > \gnts(\epsilon,\thxs)},\
\end{aligned}
\end{equation*}%
where $\V = \prod_\alpha \frac{L_\alpha}{\lambda_\alpha}$.
The second line follows from the near paralleling of the short chords that appear in the first line and tangent vectors, as discussed in \cref{m:sec:strat,m:sec:distintra} (see \cref{s:eq:shorttangent} in \cref{s:sec:cellcurv}\supp\ for justification).
The inequality in the third line comes from the union bound.
The fourth line follows from translation invariance in the line above, and the fact that the number of cells is $\V/\clsz^{K}$ (see \cref{m:eq:celldef}).
The last inequality is a result of the definition of $\gnts(\epsilon,\thxs)$ in \cref{m:eq:gntsdef}, \cref{m:sec:distintra}.
In essence, the contra-positive of \cref{m:eq:gntsdef} states that,
if the distortion of any tangent plane in $\gauss\cell_{\vec{m}}$ is greater than $\epsilon$,
then the distortion of the central tangent plane must be greater than $\gnts(\epsilon,\thxs)$.
This implies that the former event is a subset of the latter event, and hence the probability of the former cannot exceed the probability of the latter.

This last quantity was bounded by \citet[Lemma 5.1]{baraniuk2008simple} and is reproduced in \cref{m:eq:JLsubspace}.
After combining this with \cref{m:eq:distcondshort,m:eq:cellapprox}
%
and minimizing with respect to $\clsz$ to obtain the tightest possible bound, we find that $\clsz^*_\cones \sim \CO(M\epsilon/KN)$, $\sin\thxs^* \sim \CO(M\epsilon/N)$ (see \cref{s:eq:cellopt}\supp), and:
\begin{equation}\label{m:eq:distprobshort}
  \delta\shrt \lesssim  \exp\prn{-\frac{M\epsilon^2}{16} + \ln\V + K \ln\prn{\frac{9\sqrt{3}\,\e N}{\epsilon\sqrt{K}}} + \frac{K}{2} },
\end{equation}
assuming that $N \gg M \gg K \gg 1 \gg \epsilon$.

We note that the optimal value of $\clsz$ and $\thxs$ turn out to be small in the limit $N \gg M$, $K \gg 1 \gg \epsilon$,
and thus satisfy the requirements of the approximations used to derive \cref{m:eq:distcondshort} and \cref{m:eq:cellapprox}, as promised.
Moreover, for these small values of $\clsz$ and $\thxs$, the bound demonstrated in \cref{m:eq:distcondshort}, and depicted in \cref{m:fig:celltang}\ref{m:fig:tangdist}, is tight.


\subsection{All chords}\label{m:sec:all}

By combining the results for long and short chords, we can compute the probability of failing to achieve distortion less that $\epsilon$ over all chords
\begin{equation*}
  \delta \equiv \Pr\brk{\distpr(\CM-\CM) > \epsilon} \leq \delta\lng + \delta\shrt.
\end{equation*}%
Comparing \cref{m:eq:distproblong} and \cref{m:eq:distprobshort}, we see that $\delta\lng \ll \delta\shrt$.
Therefore, we only need to keep $\delta\shrt$:
\begin{equation}\label{m:eq:distproball}
  \delta \lesssim \exp\prn{-\frac{M\epsilon^2}{16} + \ln\V + K \ln \brk{\frac{9\sqrt{3}\,\e N}{\epsilon\sqrt{K}}} }.
\end{equation}
%

\begin{figure}[tbp]
  \centering
  \begin{myenuma}
    \item\aligntop{\includegraphics[height=0.34\linewidth]{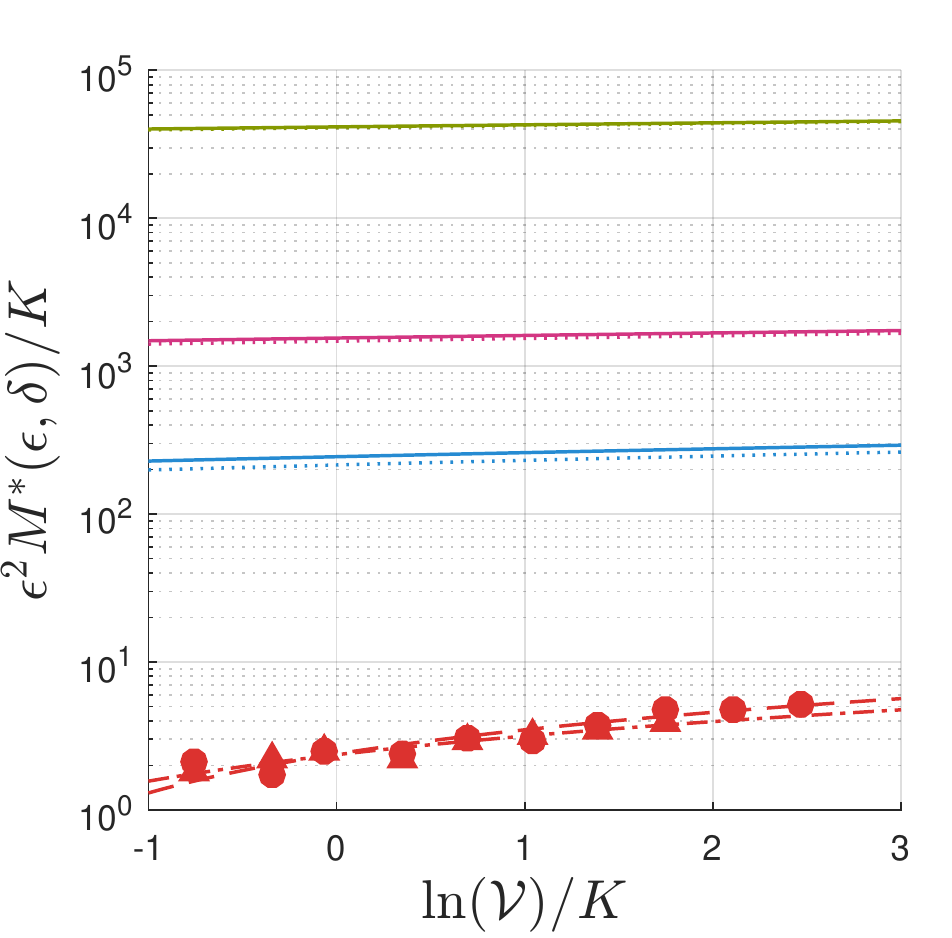}}\label{m:fig:projvol}
    \item\aligntop{\includegraphics[height=0.34\linewidth]{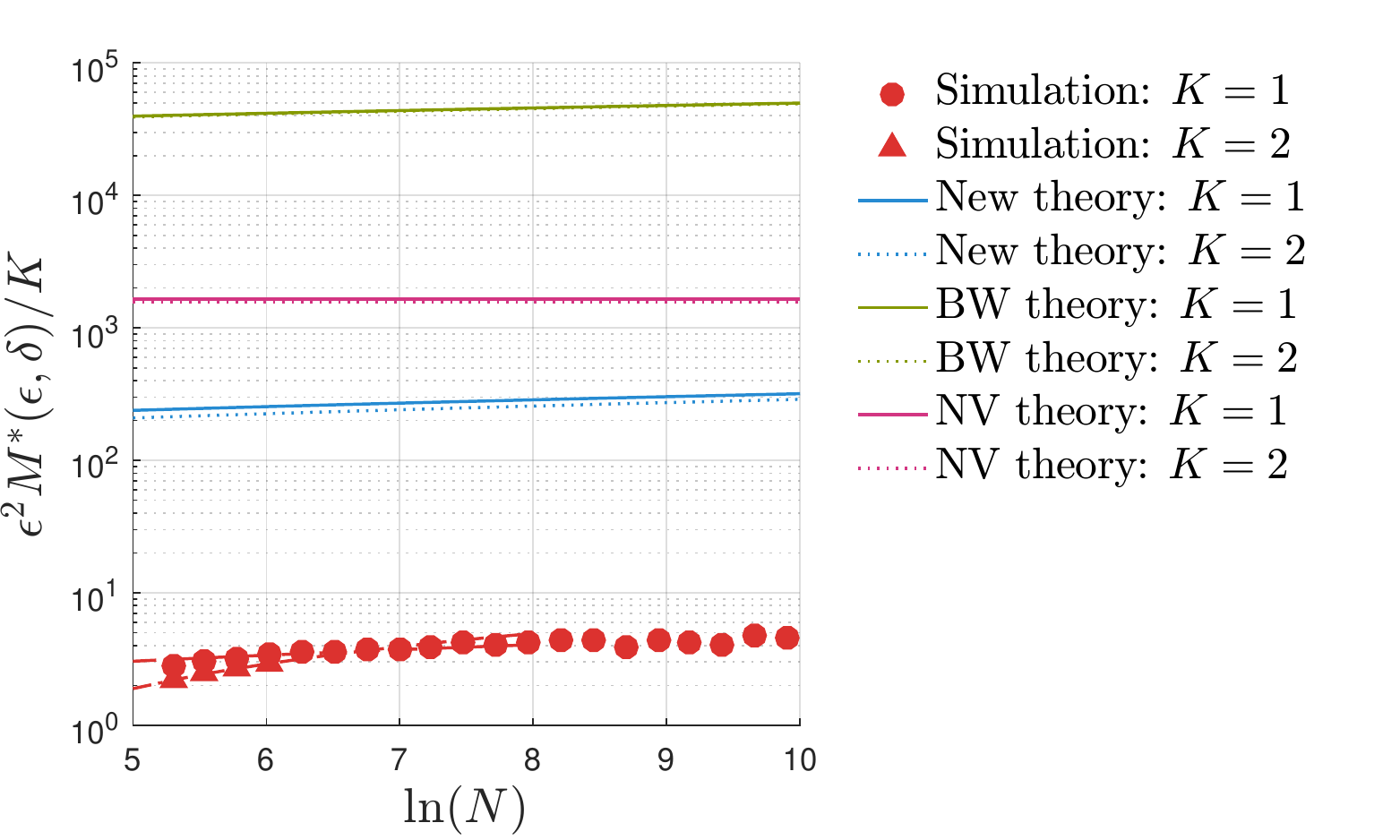}}\label{m:fig:projamb}
  \end{myenuma}
  \caption[Testing formulae for distortion of manifolds]{
  Comparison of the results of \cite{Baraniuk2009JLmfld} (``BW theory''), \cite{verma2011note} (``NV theory'') and our result from \cref{m:eq:numprojprob} (``New theory'') to numerical experiments on random manifolds and random projections.
  (\ref{m:fig:projvol}) For the numerical experiments, we fix the ambient dimension $N=1000$.
  For each $M$ ranging from $4$ to $200$, each $\ln \V$, and each $K$, we generate one random manifold $\CM$.
  We then sample $100$ random projections and for each projection $\proj$ we compute the distortion $\distpr(\CM)$, obtaining an empirical distribution of distortions.
  We then compute the distortion $\epsilon(M, K, \ln \V, N)$ that leads to a failure probability of $\delta=0.05$ under this empirical distribution.
  We then interpolate between between different values of $M$ to find the minimum value of $M$ sufficient to achieve the desired value of $\epsilon = 0.2$ with failure probability $\delta$.
  This numerically extracted minimum value of $M$ can be compared with $M^*(\epsilon=0.2, \delta = 0.05)$ in our theory.
  We compare these numerical results for $M$, $\epsilon$, $\V$, $K$ and $N$ with our new theory predicted in \cref{m:eq:numprojprob}, as well as a previous theories by \cite{Baraniuk2009JLmfld,verma2011note} (see \cref{s:sec:BW}\supp\ for the calculation of relevant geometric quantities in these previous theories).
  (\ref{m:fig:projamb})  Here, for numerical experiments, we follow a similar procedure as in (\ref{m:fig:projvol}) except we fix $\V$ to $(10\sqrt{2}/3)^K$ and we vary $N$ from $200$ to $20,000$.
  Separate plots of the numerical experiments and ``New theory'', with varying values of $\epsilon$, can be found in \cref{s:fig:distmansep}\supp.
 }\label{m:fig:distman}
\end{figure}

So the minimum number of projections necessary to get distortion at most $\epsilon$ with probability at least $1-\delta$ satisfies
\begin{equation}\label{m:eq:numprojprob}
    \mmn(\epsilon, \delta) \lesssim \frac{16\prn{\ln\V  + \ln1/\delta + K \ln \brk{\frac{9\sqrt{3}\,\e N}{\epsilon\sqrt{K}}}}}{\epsilon^2}.
\end{equation}
This is just an upper bound on the minimum required number of projections.
It may be possible to achieve this distortion with fewer projections.
We tested this formula by generating random manifolds, computing the distortion under random projections of different dimensions and seeing how many projection dimensions are needed to get a given maximum distortion 95\% of the time.
The dominant scaling relation between the various quantities can be easily seen by dividing both sides of \cref{m:eq:numprojprob} by $\tfrac{K}{\epsilon^2}$, leading us to plot $\frac{M^*\epsilon^2}{K}$ against $\frac{\ln \V}{K}$ or $\ln N$:
\begin{equation*}
    \frac{\epsilon^2 \mmn(\epsilon, \delta)}{K} \lesssim
       \frac{16\ln\V}{K}
       + \frac{16\ln1/\delta}{K}
       + 16 \ln \brk{\frac{9\sqrt{3}\,\e N}{\epsilon\sqrt{K}}}
\end{equation*}
The results are shown in \cref{m:fig:distman}, along with a comparison with previous results from \citet{Baraniuk2009JLmfld,verma2011note}.

We note that the bounds on random projections of smooth manifolds by \citet{Baraniuk2009JLmfld} contain two geometric properties of the manifold: the geodesic covering regularity, $R$, and the inverse condition number, $\tau$.
In \cref{s:sec:BW}\supp, we find a lower bound on $R$ in \cref{s:eq:geocovreg} and an upper bound on $\tau$ in \cref{s:eq:cond_bnd}.
As these two quantities appear in the result of \citet{Baraniuk2009JLmfld} in the form $\ln(R/\tau)$, we will underestimate their upper bound on the number of projections sufficient to achieve distortion $\epsilon$ with probability $1 - \delta$.
This underestimate is:
\begin{equation*}
  \frac{M^* \epsilon^2}{K} \sim
     {\frac{1352\ln\V}{K}}
     + {\frac{676\ln 1/\delta}{K}}
     + 676\ln \brk{\frac{3100^4 N^3 K}{4\pi\e\,\epsilon^6}}\!.
\end{equation*}%
Despite the fact that we \emph{underestimate} \citeauthor{Baraniuk2009JLmfld}'s upper bound, in \cref{m:fig:distman} we see that
our radically different methods provide a tighter upper bound by more than two orders of magnitude, even relative to this underestimate of previous results.

We also computed a lower bound on the result of \citet{verma2011note}, which is a bound on the distortion of curve lengths which is itself a lower bound on the distortion of chords.
This bound also requires the geodesic covering number as well as a uniform bound on the second fundamental form, which we compute in \cref{s:eq:secformnrm}, \cref{s:sec:BW}\supp.
This underestimate is:
\begin{equation*}
  \frac{M^* \epsilon^2}{K} \sim
     {\frac{64\ln\V}{K}}
     + {\frac{64\ln 1/\delta}{K}}
     + 32\ln \brk{\frac{384^5 \cdt 169 K}{\pi\e\,\epsilon^6}}\!,
\end{equation*}%
In \cref{m:fig:distman}, we see that our bound is roughly an order of magnitude smaller than this underestimate for the parameter values considered here.
As \citeauthor{verma2011note}'s bound is independent of $N$, it will become smaller than \cref{m:eq:numprojprob} for sufficiently large $N$.
For the parameter values considered here, the crossover is at $N \sim \CO(10^{36})$.
Further analytic comparison of these theoretical results can be found in \cref{s:sec:plotsep}\supp.

%


\section{Discussion}

The ways in which the bound on the required projection dimensionality in \cref{m:eq:numprojprob} scales with distortion, volume, curvature, \etc is similar to previous results from \citet{Baraniuk2009JLmfld,clarkson2008tighter,verma2011note}.
However, the coefficients we find are generally smaller, with the exception of the dependence on $\ln N$.
In practical applications, these coefficients are very important.
When one wishes to know how many projections to use for some machine learning algorithm to produce accurate results, knowing that it scales logarithmically with the volume of the manifold is insufficient.
One needs at least an order of magnitude estimate of the actual number.
We have seen in \cref{m:fig:distman} that our methods produce bounds that are significantly tighter than previous results, but there is still room for improvement.
We will discuss these possibilities and other issues in the remaining sections.


\subsection{The approximate nature of our bounds}\label{m:sec:approx}

In contrast to previous work by \citet{Baraniuk2009JLmfld}, \citet{clarkson2008tighter} and \citet{verma2011note}, our bound on the number of projections $M$ sufficient to guarantee preservation of geometry to an accuracy level $\epsilon$ with probability of success $1-\delta$ in \cref{m:eq:numprojprob} should be viewed as an approximate bound.
However, in appropriate limits relevant to practical cases of interest, we expect our bound to be essentially an exact upper bound.
What is the regime of validity of our bound and the rationale for it?
We discuss this in detail in the beginning of the \hyperref[s:appendices]{\suppname}.
However, roughly, we expect our bound to be an exact upper bound in the limit $N \gg M \gg K \gg 1 \gg \epsilon$, along with $N \gg K^2$,  $N \gg \ln \V$, and $L_\alpha \gg \lambda_\alpha$.
Some of these requirements are fundamental, while others can be relaxed, leading to more complex, but potentially tighter approximate upper bounds (see \hyperref[s:appendices]{\suppname}).

\par Most importantly, the constraint that $N \gg \ln \V$ enables us to neglect fluctuations in the geometry of the random manifold $\CM$.
Intuitively, as discussed below, $\V$ measures the number of independent correlation cells in our random manifold ensemble.
A heuristic calculation based on extreme value theory reveals that the constraint $N \gg \ln \V$ ensures that the geometric properties of even the most extreme correlation cell will remain close to the mean across cells, with the probability of $\CO(1)$ fluctuations in geometry exponentially suppressed in $N$.
In contrast, the requirement that $N \gg M$ is less fundamental, as it simply enables us to simplify various formulas associated with chordal and tangential cones; in principle, slightly tighter but more complex bounds could be derived without this constraint.
However, in many situations of interest, especially when random projections are successful, we naturally have $N \gg M$.
Also $\epsilon \ll 1$ and $M \gg K$ are not as fundamental to our approach.
We simply focus on $\epsilon \ll 1$ to ignore cumbersome terms of $\CO(\epsilon^3)$ in the JL lemma; but also $\epsilon \ll 1$ is the interesting limit when random projections do preserve geometry accurately.
Moreover, since the natural scale of distortion $\epsilon$ is $\CO(K/M)$, the focus on the $\epsilon \ll 1$ limit implies the $M \gg K$ limit.
Finally, the constraints $N \gg K^2$, $K \gg 1$ and $L_\alpha \gg \lambda_\alpha$ are, at a technical level, related to the ability to approximate sums over our discretization of $\CM$ into cells from \cref{m:eq:celldef}
with integrals,  while ignoring boundary effects in the integration.
Furthermore, they are required to approximate the resulting integral with a saddle point (see \cref{s:sec:long}\supp).
These constraints are technical limitations of our theoretical approach, but they do not exclude practical cases of interest.


\subsection{The gap between our theoretical bounds and numerical experiments}\label{m:sec:gap}

As seen in \cref{m:fig:distman}, our upper bound in \cref{m:eq:numprojprob}, while still being $2$ orders of magnitude tighter than an underestimate of an upper bound derived by \cite{Baraniuk2009JLmfld}, nevertheless exhibits a gap of about $2$ orders of magnitude relative to actual numerical simulations.
What is the origin of this gap?
First, our numerical simulations obey the approximate scaling law (fits not shown):
\begin{equation}\label{m:eq:numscale}
  \mmn(\epsilon, \delta) \approx \frac{ 1.2 \ln \V + 2.5 K}{\epsilon^2}.
\end{equation}
Comparing the numerical scaling in \cref{m:eq:numscale} to our theoretical upper bound in \cref{m:eq:numprojprob}, we see two dominant sources of looseness in our bound: (a) the pre-factor of $16$, and (b) the term involving $\ln \tfrac{N}{\epsilon}$.
We discuss each of these in turn.

First, the pre-factor of $16$ in \cref{m:eq:numprojprob} originates from our reliance on the subspace JL lemma in \cref{m:eq:JLsubspace}.
In essence, we required this lemma to bound the failure probability of preserving the geometry all tangent planes within a tangential cone, by bounding the failure probability of single tangent plane at the center of a cell.
However, one can see through random matrix theory, that the subspace JL lemma is loose, relative to what one would typically see in numerical experiments, precisely by this factor of $16$.
Indeed, when viewed as an upper bound on the distortion $\epsilon$ incurred by projecting a $K$ dimensional subspace in $\R^N$, down to $M$ dimensions, \cref{m:eq:JLsubspace} predicts approximately, $\epsilon  \lesssim 4 \sqrt{K/M}$.
However, for a $K$ dimensional subspace, this distortion is precisely related to the maximum and minimum singular values of an appropriately scaled random matrix, whose singular value distribution, for large $M$ and $K$, is governed by the Wishart distribution \citep[see][]{wishart1928generalised}.
A simple calculation based on the Wishart distribution then yields a typical value of distortion $\epsilon = \sqrt{{K}/{M}}$ (see \eg \citet[Sec. 5.3]{advani2013statistical}).   Thus the subspace JL lemma bound on $\epsilon^2$ is loose by a factor of 16 relative to typical distortions that actually occur and are accurately predicted by random matrix theory.

Second, the term involving $\ln \tfrac{N}{\epsilon}$ in \cref{m:eq:numprojprob} originates from our strategy of surrounding cells by chordal cones or tangential cones that explore the entire $N$ dimensional ambient space, rather than being restricted to the neighborhood of the $K$ dimensional manifold $\CM$.
For example, to bound the failure probability of preserving the geometry of all short chords within a cell, we want to bound the failure probability of all tangent planes within a cell.
Since this set is difficult to describe, we instead bound the failure probability of a strictly larger set: the tangential cone of \emph{all} subspaces in $\R^N$ within a maximal principal angle of the tangent plane at the center (see \cref{m:sec:distintra}).
This tangential cone contains many subspaces that twist in the ambient $\R^N$ in ways that the tangent planes of $\CM$ restricted to the cell do not.
As a result, preserving the geometry of the tangential cone to $\CO(\epsilon)$ requires the angle of the tangent cone $\thxs$ to be $\CO(M\epsilon/N)$.
Then summing the failure probability over all these small tangential cones via the union bound yields the $\ln \tfrac{N}{\epsilon}$ term in \cref{m:eq:numprojprob}.
Roughly, for $N=1000$ and $\epsilon=\CO(0.1)$ as in \cref{m:fig:distman},  we obtain $\ln \tfrac{N}{\epsilon} \approx 10$.
When combined with the multiplicative factor of $16$, we roughly explain the two order magnitude looseness of our theoretical upper bound, relative to numerical experiments.

Thus any method to derive a tighter JL lemma for subspaces, or a proof strategy that does not involve needlessly controlling the distortion of all possible ways tangent planes could twist in the ambient $\R^N$, would lead to even tighter bounds.
Indeed, the fact that the bounds proved by \citet{clarkson2008tighter,verma2011note} are independent of $N$ (though at the expense of extremely large constants that dominate for values of $N$ used in practice), in addition to the numerical scaling in \cref{m:eq:numscale}, suggests that the term involving $\ln{N}$ in \cref{m:eq:numprojprob} could potentially be removed.
Overall, the simple numerical scaling we observe in \cref{m:eq:numscale} illustrates the remarkable power of random projections to preserve the geometry of curved manifolds. To our knowledge, the precise constants and scaling relation we observe in \cref{m:eq:numscale} have never before been concretely ascertained for any ensemble of curved manifolds. Indeed this scaling relation provides a precise benchmark for testing previous theories and presents a concrete target for future theory.


\subsection{A measure of manifold complexity through random projections}\label{m:sec:complexity}

Intriguingly, random projections provide a potential answer to a fundamental question: what governs the size, or geometric complexity of a manifold?
The intrinsic dimension does not suffice as an answer to this question.
For example, intuitively, a $K$ dimensional linear subspace seems less complex than a $K$ dimensional curved manifold, yet they both have the same intrinsic dimension.
Despite their same intrinsic dimensionality, from a machine learning perspective, it can be harder to both learn functions on curved manifolds and compress them using dimensionality reduction, compared to linear subspaces.
In contrast to the measure of intrinsic dimensionality, the number random projections $M$ required to preserve the geometry of a manifold to accuracy $\epsilon$ with success probability $1-\delta$ is much more sensitive to the structure of the manifold.
Indeed, this number can be naturally interpreted as a potential description of the size or complexity of the manifold itself.

The naturalness of this interpretation can be seen directly in the analogous results for a cloud of $P$ points in \cref{m:eq:JLfail}, a $K$ dimensional subspace in \cref{m:eq:JLsubspace}, and our smooth manifold ensemble in \cref{m:eq:numprojprob}.
To leading order, the number of projections $M$ grows as $\ln P$ for a cloud of $P$ points, as the intrinsic dimension $K$ for a linear subspace, and as the intrinsic dimension $K$ \emph{plus} and additional term $\ln \V$ for a smooth manifold.
Here $\V$ measures the volume of the entire manifold $\scriptstyle{\prod}_{\alpha=1}^K L_\alpha$ in intrinsic coordinates, in units of the volume of an autocorrelation cell $\scriptstyle{\prod}_{\alpha=1}^K \lambda_\alpha$.
Thus $\V$ is a natural measure of the number of independently moveable degrees of freedom, or correlation cells, in our manifold ensemble.
The entropy-like logarithm of $\V$ is the additional measure of geometric complexity, as measured through the lens of random projections, manifested in a $K$ dimensional nonlinear curved manifold $\CM$, compared to a $K$ dimensional flat linear subspace.

To place this perspective in context, we review other measures of the size or complexity of a subset of Euclidean space that are related to random projection theory.
For example, the statistical dimension of a subset $\cell \subset \R^N$ measures how maximally correlated a normalized vector restricted to $\cell$ can be with a random Gaussian vector $\xv \in \R^N$ \citep[see][]{amelunxen2014living}.
Generally, larger sets have larger statistical dimension.
This measure governs a universal phase transition in the probability that the subset $\cell$ lies in the null-space of a random projection $\proj$ \citep[see][]{oymak2015universality}.
Essentially, if the dimensionality $M$ of the projection is more (less) than the statistical dimension of $\cell$, then the probability that $\cell$ lies in the null-space of $\proj$ is exceedingly close to $0$ ($1$).
Thus sets with smaller statistical dimension require fewer projections to escape the null-space.
Also related to statistical dimension, which governs when points in $\cell$ shrink to $0$ under a random projection, a certain excess width functional of $\cell$ governs, more quantitatively, the largest amount a point in $\cell$ can shrink under a random projection.
When $\cell$ is a linear $K$ dimensional subspace, the shrinkage factor is simply the minimum singular value of an $M$ by $K$ submatrix of $\proj$, but for more general $\cell$ the shrinkage factor is a restricted singular value \citep[see][]{oymak2015universality}.
Finally, another interesting measure of the size of a subset  $\cell \subset \R^N$ is its Gelfand width \citep[see][]{pinkus2012n}. The $M$-Gelfand width of $\cell$ is the minimal diameter of the intersection of $\cell$ with all possible $N-M$ dimensional null-spaces of projections down to $M$ dimensions.
Again, larger sets have larger Gelfand widths.
In particular, sets of small $M$-Gelfand width have their geometry well preserved under a random projection down to $M$ dimensions \citep[see][]{baraniuk2008simple}.

While the number of random projections required to preserve geometry, the statistical dimension, the excess width functional, and the Gelfand width all measure the size or geometric complexity of a subset, it can be difficult to compute the latter three measures, especially for smooth manifolds.
Here we have studied an ensemble of random manifolds from the perspective of random projection theory, but it may be interesting to study such an ensemble from the perspective of these other measures as well.




\subsection*{Acknowledgements}



We thank the Burroughs-Wellcome, Genentech, Simons, Sloan, McDonnell, and McKnight Foundations, and the Office of Naval Research for support.



\appendix\phantomsection\label{s:appendices}
\section*{\suppname}

In these appendices, we provide the derivations of the results presented in \cref{m:sec:boundcell,m:sec:randman,m:sec:logic}\main.

First, in \cref{s:sec:approx} we explicitly list our approximations and assumptions in advance of their use in the subsequent \lcnamecrefs{s:sec:approx}.
In \cref{s:sec:prelim}, we will describe the formalism and the mathematical concepts we use, such as principal angles and definitions of distortion for different geometrical objects.

In \cref{s:sec:boundcell}, we find constraints that need to be satisfied by chords involving the cell centers so that we are guaranteed that all chords have distortion less than $\epsilon$.
In \cref{s:sec:distinter}, we present the derivation of \cref{m:eq:distcondlong} from \cref{m:sec:distinter}, regarding the distortion of long, intercellular chords.
In \cref{s:sec:distintra}, we present the derivation of \cref{m:eq:distcondshort} from \cref{m:sec:distintra}, regarding the distortion of short, intracellular chords.

These results will depend on the size, separation and curvature of these cells, which we will calculate in \cref{s:sec:randman}.
In \cref{s:sec:cellsep}, we derive \cref{m:eq:cellchang} in \cref{m:sec:cellsep} by finding the radius of a cell and the distance between two cell centers.
In \cref{s:sec:cellcurv}, we derive \cref{m:eq:cellapprox} in \cref{m:sec:cellcurv} by bounding the principal angles between tangent planes at the center and edges of cells.
In \cref{s:sec:BW}, we calculate bounds for geometric quantities, such as the geodesic regularity, condition number and the norm of the second fundamental form, for the ensemble of random manifolds we consider here.
These quantities appear in the formulae derived by \citet{Baraniuk2009JLmfld} and \citet{verma2011note}.
This allows us to find the lower bounds on their formulae, which are plotted in \cref{m:fig:distman} and compared with our result, \cref{m:eq:numprojprob}, and simulations.

In \cref{s:sec:logic}, we combine these results to bound the failure probability for all chords of the submanifold.
We present the detailed derivation of \cref{m:eq:distproblong} in \cref{m:sec:long} by combining the results of \cref{s:sec:distinter,s:sec:cellsep}.
We present the detailed derivation of \cref{m:eq:distprobshort} in \cref{m:sec:short} by combining the results of \cref{s:sec:distintra,s:sec:cellcurv}.
In \cref{s:sec:plotsep}, we provide separate plots of the simulations and our result from \cref{m:eq:numprojprob}.
We also compare the \cref{m:eq:numprojprob} with the results of \citet{Baraniuk2009JLmfld} and \citet{verma2011note} analytically.


\section{List of approximations used}\label{s:sec:approx}

A central aspect of our approach is that we derive approximate upper bounds on the failure probability of the preservation of geometry through random projections.   In order to be explicit about the nature of our approximations, why we need them, and their regime of validity, we first discuss the particular approximations here, before delving into our approach.  In particular, our approximations require $N,M,K,\epsilon,L_\alpha\lambda_\alpha$ to lie in a certain regime.  Fortunately this regime does not exclude many cases of practical interest.

We require $N \gg M$ to derive approximate forms of the functions $\gntc(\epsilon,\thxc)$ and $\gnts(\epsilon,\thxs)$ that appear in \cref{s:sec:boundcell}, \cref{s:eq:distcondlong,s:eq:distcondshort}.

In \cref{s:sec:randman}, we require $N \gg K\log(L/\lambda)$ to neglect fluctuations about the self-averaging results in \cref{s:eq:selfavgvalidpair,s:eq:selfavgvalidsingle}.
To derive approximate forms of $\thxc$ and $\thxs$ in \cref{s:eq:cellchang,s:eq:cellapprox} we need $\clsz \ll 1/\sqrt{K}$, which becomes $\epsilon \sqrt{M/N} \ll 1$ and $M \epsilon/N \ll 1$ for the values of $\clsz$ used there.
As we already assumed that $N \gg M$, the only additional requirement is $\epsilon \not\gg 1$.

In order to neglect the $\epsilon^3$ terms in the JL lemma, \cref{s:eq:JLlemma}, and its subspace analogue, \cref{s:eq:distprobcent}, both in \cref{s:sec:logic}, we need $\epsilon \ll 1$.
As typically $\epsilon \sim \CO(\sqrt{K/M})$ (see \cite{Baraniuk2009JLmfld,baraniuk2008simple}), this implies that $M \gg K$.

In \cref{s:sec:long}, we need $\clsz \ll 1/(KNM\epsilon^2)^{1/4}$ to approximate sums with integrals for \cref{s:eq:longintegral}, which amounts to $N \gg (M\epsilon^2)^3/K \sim K^2$ for the value of $\clsz$ used there.
To perform the integrals with the saddle point method for \cref{s:eq:saddledef} we need $M\epsilon^2, K \gg 1$, and ignoring boundary corrections requires $L_\alpha \gg \lambda_\alpha$.
Neglecting $\delta\lng$ relative to $\delta\shrt$ in \cref{m:eq:distproball}\main\ also requires $M\epsilon^2,K \gg 1$, but this is already assumed.

In summary, we will work in the regime $N \gg M \gg K \gg 1 \gg \epsilon$, along with $N \gg K^2$, $N \gg K\log(L/\lambda)$ and $L_\alpha \gg \lambda_\alpha$.


\section{Preliminary definitions}\label{s:sec:prelim}

Before embarking on the derivation of our bound, we will define some useful concepts and conventions for subspaces, projections \etc as well as the definition of the various types of distortion we will come across.


\subsection{Subspaces, projections and principal angles}\label{s:sec:angles}

In this section, we describe the notation and conventions we will use for subspaces, projections and the angles between them, and some of their useful properties.

Let $\Usp$ and $\Usp'$ be two $K$ dimensional subspaces of $\R^N$, \ie members of the Grassmannian $\grass_{N,K}$.
They can be represented by orthonormal bases that can be put into $N \times K$ column-orthogonal matrices, $\U,\U'$ with $\U\trans\U = \U'\transp\U' = \I_K$.
However, any orthonormal linear combinations of these would provide equally good bases:
\begin{equation*}
  \U \to \U \mathbf{V},
  \qquad
  \mathbf{V} \in O(K).
\end{equation*}%
Any meaningful descriptor of these subspaces is invariant under these $O(K)$ transformations.

Now we define the principal angles between these two spaces.
First we can find the vectors in each space that have the smallest angles with each other, \ie the unit vectors, $\uv \in \Usp$ and $\uv' \in \Usp'$ that maximize $\uv\cdt\uv' = \cos\theta$.
Then we do the same in the subspaces of $\Usp$ and $\Usp'$ that are perpendicular to $\uv$ and $\uv'$ respectively.
We can repeat this process until there are no dimensions left.
The $K$ resulting angles can be put in the vector $\pang$.
Unit vectors in $\Usp$ can be written as $\uv = \U\sv$, with $\sv \in \Sp^{K-1}_1$ (the unit sphere in $\R^K$).
The maximization we need is then
\begin{equation*}
  \max_{\sv,\sv' \in \Sp^{K-1}_1} \sv\trans \U\trans \U' \sv'.
\end{equation*}%
This results in the Singular Value Decomposition (SVD) of $\U\trans\U'$:
\begin{equation}\label{s:eq:anglesvd}
  \U\trans \U' = \mathbf{W} \operatorname{diag}(\cos\pang) \mathbf{V}\trans,
  \qquad
  \mathbf{W},\mathbf{V} \in O(K).
\end{equation}
This means that the eigenvalues of $\U\trans \U' \U'\transp \U$ and $\U'\transp \U \U\trans \U'$ are $\cos^2\pang$.

An $M$ dimensional projector is an $M \times N$ matrix $\proj$.
The subspace of $\R^N$ spanned by its rows is denoted by $\projsp$.
We will assume that $M>K$ and that the projection is row orthogonal, $\proj\proj\trans = \I_M$, except where explicitly stated.

We can define principal angles between a subspace $\Usp$ and the projection $projsp$ via the SVD of $\proj\U$:
\begin{equation*}
  \proj\U = \mathbf{W} \operatorname{diag}(\cos\prang) \mathbf{V}\trans,
  \qquad
  \mathbf{W} \in \R^{M,K}, \mathbf{W}\trans\mathbf{W} = \I_K, \;
  \mathbf{V} \in O(K).
\end{equation*}%
The eigenvalues of $\U\trans\proj\trans\proj\U$ are $\cos^2\prang$, but the eigenvalues of $\proj\U\U\trans\proj\trans$ are $\cos^2\prang$ with an additional $M-K$ zeros.
If the projector is not orthogonal, the singular values will not be interpretable as cosines of angles.

Another way of characterizing a subspace is via the projection matrix $\U\U\trans$.
One could also describe the difference between two subspaces via $\U\U\trans-\U'\U'\transp$.
What are the singular values of this matrix?
One can look at the eigenvalues of its square
\begin{equation}\label{s:eq:projev}
\begin{aligned}
  \brk{\U\U\trans + \U'\U'\transp - \U\U\trans\U'\U'\transp - \U'\U'\transp\U\U\trans} \vv &= \lambda\vv \\
  \implies \quad
  \brk{\I - \U\trans\U'\U'\transp\U} \U\trans\vv &= \lambda \U\trans\vv,\\
  \text{and} \quad
  \brk{\I - \U'\transp\U\U\trans\U'} \U'\transp\vv &= \lambda \U'\transp\vv,
\end{aligned}
\end{equation}
where the second line comes from multiplying the first from the left by $\U\trans$ and the third by multiplying the first by $\U'\transp$.
If both $\U\trans\vv$ and $\U'\transp\vv$ are zero then $\lambda$ must be zero.
Otherwise, either $\U\trans\vv$ is an eigenvector of $\U\trans\U'\U'\transp\U$, which has eigenvalues $\cos^2\theta_a$, or $\U'\transp\vv$ is an eigenvector of $\U'\transp\U\U\trans\U'$, which also has eigenvalues $\cos^2\theta_a$.
So the eigenvalues are either 0 or $\sin^2\theta_a$, and the singular values are either 0 or $\sin\theta_a$.
Taking the trace of the equation above:
\begin{equation*}
  \nrm{\U\U\trans-\U'\U'\transp}^2_\text{F} = 2K - 2 \nrm{\U\trans\U'}^2_\text{F} = 2 \sum_a\sin^2\theta_a.
\end{equation*}%
This implies that the $\sin\theta_a$ singular values have multiplicity 2 and 0 has multiplicity $N-2K$ (unless some of the angles are zero).

We can apply the same analysis to $\proj\trans\proj - \U\U\trans$.
If $\proj\vv$ is nonzero it must be an eigenvector of $\proj\U\U\trans\proj\trans$, which has eigenvalues 0 and $\cos^2\phi_a$.
If $\U\trans\vv$ is nonzero it must be an eigenvector of $\U\trans\proj\trans\proj\U$, which has eigenvalues $\cos^2\phi_a$.
If both are zero, the eigenvalue must be zero.
Therefore, the eigenvalues are either 0, 1 or $\sin^2\phi_a$, and the singular values are either 0, 1 or $\sin\phi_a$.
This only works if we are using orthogonal projections.
Taking the trace:
\begin{equation*}
  \nrm{\proj\proj\trans - \U\U\trans}^2_\text{F} = M + K - 2 \nrm{\proj\U}^2_\text{F} = M-K + 2 \sum_a\sin^2\varphi_a.
\end{equation*}%
This implies that the $\sin\phi_a$ singular values have multiplicity 2, 0 has multiplicity $N-M-K$ and 1 has multiplicity $M-K$ (unless some of the angles are zero or $\pi/2$).

If $\svd_i$ are the singular values of $\mathbf{B}$, $\svd'_i$ are the singular values of $\mathbf{B}'$ and $\rho_i$ are the singular values of $\mathbf{B}-\mathbf{B}'$, Weyl's inequality states that
\begin{equation*}
  \rho\lmin \leq \abs{\svd_i - \svd'_i} \leq \rho\lmax.
\end{equation*}%
If we apply this to $\proj\trans\proj - \U\U\trans$ and $\proj\trans\proj - \U'\U'\transp$, we find
\begin{equation}\label{s:eq:weylang}
  \abs{\sin\varphi_a - \sin\varphi'_a} \leq \sin\thx.
\end{equation}
Again, this only works if we are using orthogonal projections.


\subsection{Distortions under projections}\label{s:sec:distort}

In this section we list the definitions of distortion under a projection $\proj$ for various structures, such as vectors, vector spaces, Grassmannians and manifolds.

The distortion of an $N$ dimensional vector $\uv$ under an M-dimensional projection $\proj$ is defined as
\begin{equation*}
  \distpr(\uv) = \abs{\sqrt{\frac{N}{M}} \frac{\nrm{\proj\uv}_2}{\nrm{\uv}_2} - 1}.
\end{equation*}%
This depends only on the direction of $\uv$ and not its magnitude.

If $\cell_1$ and $\cell_2$ are subsets of $\R^N$, their relative distortion is defined as
\begin{equation*}
  \distpr(\cell_1 - \cell_2) = \max_{\xv_1 \in \cell_1, \xv_2 \in \cell_2} \distpr(\xv_1 - \xv_2).
\end{equation*}%
The main subset we will be interested in is a submanifold, $\CM$,  of $\R^N$:
\begin{equation}\label{s:eq:distman}
  \distpr(\CM - \CM) = \max_{\xv_1, \xv_2 \in \CM} \distpr (\xv_1 - \xv_2).
\end{equation}
where $\xv_1, \xv_2$ are points of $\R^N$ that lie on the submanifold, so $\xv_1 - \xv_2$ is a chord.

The distortion of a vector subspace, $\Usp$, of $\R^N$ is defined as
\begin{equation*}
  \distpr(\Usp) = \max_{\uv \in \Usp} \brc{\distpr (\uv)}.
\end{equation*}%
This is related to singular value decomposition.
Without loss of generality, we can assume $\uv$ is a unit vector.
Unit vectors in $\Usp$ can be written as $\U\sv$, where $\sv$ is a unit vector in $\R^K$.
Then
\begin{equation}\label{s:eq:distspcsvd}
\begin{aligned}
  \distpr(\Usp) &= \max_{\sv \in \Sp^{K-1}} \abs{\sqrt{\frac{N}{M}} \nrm{\proj\U\sv}_2 - 1} \\
    &= \max \brc{ \sqrt{\frac{N}{M}} \, \svd\lmax - 1, 1 - \sqrt{\frac{N}{M}} \, \svd\lmin } \\
    &= \max \brc{ \sqrt{\frac{N}{M}} \cos\varphi\lmin - 1, 1 - \sqrt{\frac{N}{M}} \cos\varphi\lmax },
\end{aligned}
\end{equation}
where $\svd_a$ are the singular values of $\proj\U$.
The last line requires an orthogonal projection.

The Grassmannian, $ \grass_{K,N}$ is the set of all $K$ dimensional vector subspaces of $\R^N$.
The distortion of a subset of the Grassmannian, $\cell$, is defined as
\begin{equation*}
  \distpr(\cell) = \max_{\Usp \in \cell} \distpr (\Usp).
\end{equation*}%
One particularly interesting subset is the image of a submanifold $\CM$ under the Gauss map.
The Gauss map, $\gauss$, takes points of $\CM$ to the points in the Grassmannian corresponding to the tangent plane of $\CM$ at that point, $T_m\CM$, regarded as a subspace of $\R^N$, so that $\gauss\CM$ is the set of all tangent planes of $\CM$:
\begin{equation*}
  \distpr(\gauss\CM) = \max_{m\in\CM} \distpr(\mathcal{T}_m\CM).
\end{equation*}%
This is the maximum distortion of any tangent vector of $\CM$.
As every tangent vector can be approximated arbitrarily well by a chord, $\distpr(\CM - \CM) \geq \distpr(\gauss\CM)$.
If there exists a tangent vector parallel to every chord, $\distpr(\CM - \CM) = \distpr(\gauss\CM)$.
There are manifolds that do not have this property, for example a helix, as shown in \cref{m:fig:mfld}\ref{m:fig:helix}.

The distortion of the submanifold itself, however, can be defined in two different ways.
In addition to the chord distortion in \cref{s:eq:distman}, there is another type of distortion involving geodesic distances.
Suppose $\CM$ is embedded in $\R^N$ as $\ec^i = \phi^i(\ic^\alpha)$, where $i,j = 1,\ldots,N$, $\alpha,\beta = 1,\ldots,K$. Consider a curve on $\CM$ given by $\ic^\alpha = \gamma^\alpha(t)$, so that $\ec^i = \phi^i(\gamma^\alpha(t)) \equiv \Gamma^i(t)$.
The length of this curve is given by
\begin{equation*}
  \mathcal{L}[\gamma] = \int_0^T \!\!\! \sqrt{h_{\alpha\beta} \dot{\gamma}^\alpha \dot{\gamma}^\beta} \, \dt = \int_0^T \nrm{\dot{\boldsymbol{\Gamma}}}_2 \dt,
\end{equation*}%
where the sums over the repeated $\alpha,\beta$ indices are implicit (Einstein summation convention).
The geodesic distance between two points $\ic$ and $\ic'$ is defined as
\begin{equation}\label{s:eq:geodist}
  d(\ic,\ic') = \min_{\gamma(t) \,:\, \gamma(0)=\ic, \gamma(T)=\ic'} \mathcal{L}[\gamma],
\end{equation}

The distortion of a curve is given by
\begin{equation}\label{s:eq:distcurv}
  \distpr(\gamma) = \abs{\sqrt{\frac{N}{M}} \frac{\intd[_0^T]{t} \nrm{\proj\dot{\boldsymbol{\Gamma}}}_2}{\intd[_0^T]{t} \nrm{\dot{\boldsymbol{\Gamma}}}_2} - 1}.
\end{equation}
and distortion of geodesic distance is the distortion of the curve that achieves the minimum in \cref{s:eq:geodist}.
The geodesic distortion of the manifold is the maximum distortion over all geodesics.
As any tangent vector can be thought of as an infinitesimal geodesic, this must be an upper bound for $\distpr(\gauss\CM)$.
However, we can see from \cref{s:eq:distcurv} that the distortion of any curve is bounded by the maximum distortion of $\dot{\boldsymbol{\Gamma}}$.
As $\dot{\boldsymbol{\Gamma}}$ is a tangent vector of $\CM$, its distortion is upper bounded by $\distpr(\gauss\CM)$.
Therefore, the geodesic distortion of the submanifold is equal to $\distpr(\gauss\CM)$ \citep[see also][]{verma2011note}.


\section{Bounding distortion for cells}\label{s:sec:boundcell}

In this section we are going to present the details behind the formulae presented in \cref{m:sec:boundcell}\main.
These were criteria that representative chords involving cell centers have to satisfy so that all other chords are guaranteed to have distortion less than $\epsilon$.
The probability of the representative chords failing to meet these criteria is then an upper bound on the probability of any other chord failing to have distortion less than $\epsilon$.

Here, we derive these criteria, for the long intercellular chords in \cref{s:sec:distinter} and the short intracellular chords in \cref{s:sec:distintra}.


\subsection{Bounding distortion between different cells}\label{s:sec:distinter}

In this section we present the derivation of the function $\gntc(\epsilon,\thx)$ from \cref{m:eq:distcondlong} of \cref{m:sec:distinter}\main.
This function is an upper limit on the distortion of the chord between the centers of two different cells such that \emph{all} chords between these two cells are guaranteed to have distortion less than $\epsilon$.

We have two points in $\R^N$, $\xv_1$ and $\xv_2$, the centers of two cells, and balls of diameter $\cldm$ centered at each point, with $\cldm$ chosen so that these balls completely enclose the corresponding cells.
The set of chords between the two balls form a cone that we will refer to as the chordal cone:
\begin{equation*}
  \conec = \set{\yv_1 - \yv_2}{\nrm{\yv_1 - \xv_1}, \nrm{\yv_2 - \xv_2} \leq \frac{\cldm}{2}}
\end{equation*}%
Consider a vector in this cone, $\yv \in \conec$.
Let $\Delta\xv = \yv - \xv$
These chords are described schematically in \cref{m:fig:cellchord}\ref{m:fig:conceptchord}\main, with $\Delta\xv = \Delta\xv_1 - \Delta\xv_2$.
The outermost vectors in this cone will be tangent to the spheres that bound the two balls, and therefore $\Delta\xv_{1,2}$ will be perpendicular to $\yv$ (see \cref{m:fig:cellchord}\ref{m:fig:conceptchord}\main).
This means that the angle between the central vector, $\xv$, and the edge of the cone is given by:
\begin{equation*}
  \sin\thxc = \frac{\cldm}{x},
\end{equation*}%
where $x = \nrm{\xv}$.

How small do we have to make the $\distpr(\xv)$ so that we are \emph{guaranteed} that $\distpr(\yv) \leq \epsilon$ for \emph{all} vectors in this cone, and thus all chords between the cells?
Call this quantity $\gntc(\epsilon,\thxc)$:
\begin{equation}\label{s:eq:gntcdef}
  \distpr(\xv) \leq \gntc(\epsilon,\thxc)
  \qquad \implies \qquad
  \distpr(\yv) \leq \epsilon \quad \forall \yv \, \in \conec.
\end{equation}
We will derive the form of this function here.

Without loss of generality, we can choose the subspace of the projection to be spanned by the first $M$ Cartesian axes.
It will be helpful to split all vectors into the first $M$ and last $N-M$ components, $\xv = (\xv',\xv'')$, etc., \ie $\xv' = \proj\trans\proj\xv$ and $\xv'' = (\I - \proj\trans\proj)\xv$.
We will indicate the norms of these vectors by $\ec$, $\ec'$ and $\ec''$, etc.

The distortion of $\xv$ can be written as
\begin{equation*}
  \distpr(\xv) = \abs{\sqrt{\frac{N}{M}} \frac{\ec'}{\ec} - 1} = \abs{\sqrt{\frac{N}{M}\frac{\ec\psq}{\ec\psq + \ec\ppsq}} - 1}
    = \abs{\sqrt{\frac{N}{M}\frac{1}{1 + \frac{\ec\ppsq}{\ec\psq}}} - 1}.
\end{equation*}%

\begin{figure}[tbp]
  \centering
  \begin{myenuma}
    \item\aligntop{\includegraphics[width=0.44\linewidth]{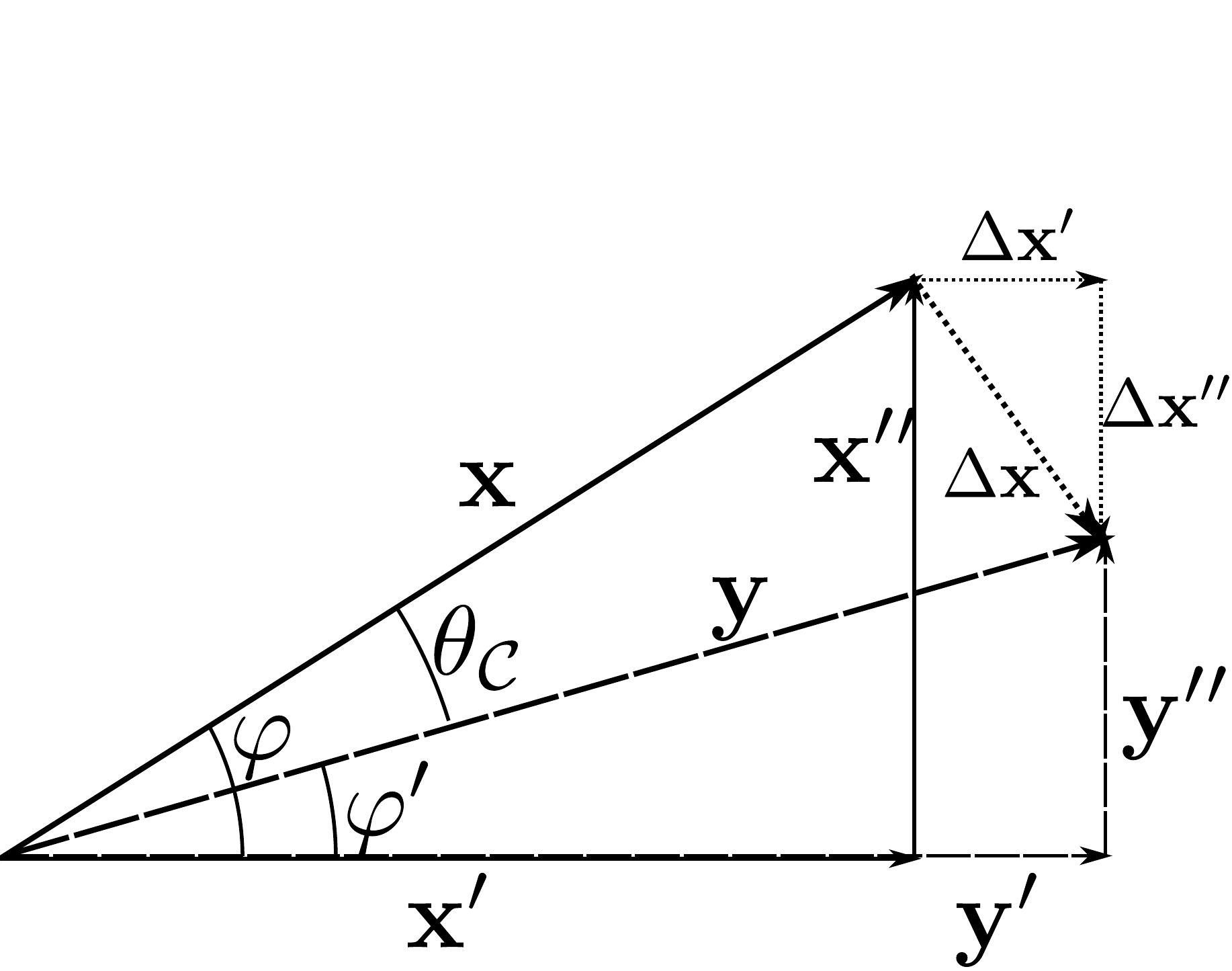}}\label{s:fig:chord_angle_in}
    \item\aligntop{\includegraphics[width=0.44\linewidth]{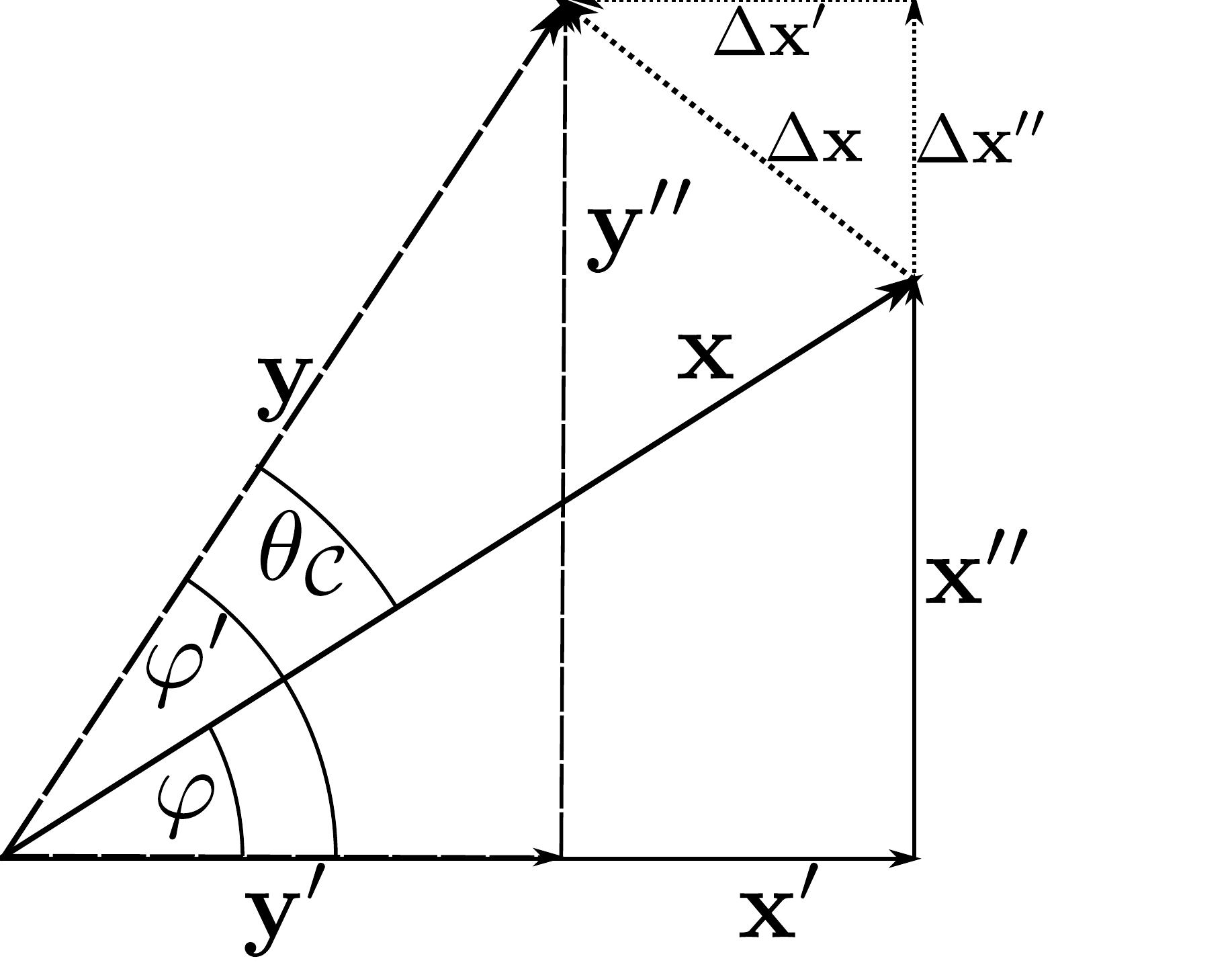}}\label{s:fig:chord_angle_out}
  \end{myenuma}
  \caption{Description of central chord $\xv$, with components $\xv'$ parallel to, and $\xv''$ perpendicular to the projector, as well as corresponding components of $\yv$, a vector on the edge of the chordal cone, $\conec$.
  The angles between the projector and $\xv(\yv)$ is $\varphi(\varphi')$, and $\thxc$ is the angle between $\xv$ and $\yv$.
  (\ref{s:fig:chord_angle_in}) When $\varphi > \varphi'$.
  (\ref{s:fig:chord_angle_out}) When $\varphi < \varphi'$.
  }\label{s:fig:chord_angle}
\end{figure}

Now, consider the distortion of $\yv = \xv + \Delta\xv$.
We define $\Delta\xv'$ and $\Delta\xv''$ in the same way as $\xv'$ and $\xv''$, with $\Delta\ec'$ and $\Delta\ec''$ defined as their norms.
This means that the component of $\yv$ parallel to the projection subspace $\projsp$ is given by $\yv' = \xv' + \Delta\xv'$ and the orthogonal component is given by $\yv'' = \xv'' + \Delta\xv''$.
We use $\theta'$ ($\theta''$) to denote the angles between $\xv'$ ($\xv''$) and $\Delta\xv'$ ($\Delta\xv''$).
The condition we need to guarantee is
\begin{equation}\label{s:eq:distdelta}
  \distpr(\yv) =
   \abs{\sqrt{\frac{N}{M}\frac{1}{1 + \frac{\nrm{\yv''}^2}{\nrm{\yv'}^2 }}} - 1} =
    \abs{\sqrt{\frac{N}{M}\frac{1}{1 + \frac{\ec\ppsq + 2\ec''\Delta \ec''\cos\theta'' + \Delta \ec\ppsq}{\ec\psq + 2\ec'\Delta \ec'\cos\theta' + \Delta \ec\psq}}} - 1} \leq \epsilon.
\end{equation}
Note that
\begin{equation}\label{s:eq:deltanorm}
  \sqrt{\Delta \ec\psq + \Delta \ec\ppsq} = \nrm{\Delta\xv_1 - \Delta\xv_2} \leq \nrm{\Delta\xv_1} + \nrm{\Delta\xv_2} \leq \cldm.
\end{equation}
Using this, in the case where the quantity inside the absolute value in \cref{s:eq:distdelta} is positive, we see that the worst case scenario is when $\theta' = 0$, $\theta'' = \pi$ and $\Delta \ec'' = \sqrt{\cldm^2 - \Delta \ec\psq}$.
This means that the component of $\Delta\xv$ parallel to $\projsp$ is parallel to the corresponding component of $\xv$, which makes the length of $\proj\yv$ as large as possible.
The component of $\Delta\xv$ perpendicular to $\projsp$ is antiparallel to the corresponding component of $\xv$, which makes the length of $\yv$ as small as possible.
This situation is described in \cref{s:fig:chord_angle}\ref{s:fig:chord_angle_in}.
Furthermore, saturating \cref{s:eq:deltanorm} requires $\Delta\xv_1$ and $\Delta\xv_2$ to be antiparallel, with lengths as large as possible, so they go from the center of their respective cells to opposite point on the boundaries of their respective balls.

In the case where the quantity inside the absolute value in \cref{s:eq:distdelta} is negative, we see that the worst case scenario is when $\theta' = \pi$, $\theta'' = 0$ and $\Delta \ec'' = \sqrt{\cldm^2 - \Delta \ec\psq}$.
This means that the component of $\Delta\xv$ parallel to $\projsp$ is antiparallel to the corresponding component of $\xv$, which makes the length of $\proj\yv$ as small as possible.
The component of $\Delta\xv$ perpendicular to $\projsp$ is parallel to the corresponding component of $\xv$, which makes the length of $\yv$ as large as possible.
This situation is described in \cref{s:fig:chord_angle}\ref{s:fig:chord_angle_out}.
Again, saturating \cref{s:eq:deltanorm} requires $\Delta\xv_1$ and $\Delta\xv_2$ to go from the center of their respective cells to opposite point on the boundaries of their respective balls.

In both cases, $\Delta\xv'$ and  $\Delta\xv''$ lie in the subspace spanned by $\xv'$ and $\xv''$, and therefore so do $\yv'$ and $\yv''$.
In this situation, all of the relevant vectors can be drawn in a plane, as shown in \cref{s:fig:chord_angle}.
If $\varphi$ is the angle between $\xv$ and the projector and $\varphi'$ is the corresponding angle for $\yv$, the conditions in \cref{s:eq:gntcdef} can be rewritten as
\begin{alignat}{5}
  \distpr(\xv) &\leq \gntc(\epsilon, \thxc)
  &&\quad\means&\quad
  \sqrt{\frac{M}{N}}(1 - \gntc(\epsilon,\thxc)) &\leq &&\cos\varphi& &\leq \sqrt{\frac{M}{N}}(1 + \gntc(\epsilon,\thxc)),
  \label{s:eq:cosxrange}\\
  \distpr(\yv) &\leq \epsilon
  &&\quad\means&\quad
  \sqrt{\frac{M}{N}}(1 - \epsilon) &\leq &&\cos\varphi'& &\leq \sqrt{\frac{M}{N}}(1 + \epsilon).
  \label{s:eq:cosyrange}
\end{alignat}
Because $\varphi'$ lies in the range $[\varphi - \thxc, \varphi + \thxc]$, we have:
\begin{equation*}
  \cos\varphi \cos\thxc - \sin\varphi \sin\thxc \leq \cos\varphi' \leq \cos\varphi \cos\thxc + \sin\varphi \sin\thxc.
\end{equation*}%
Combining this with \cref{s:eq:cosxrange} leads to
\begin{multline}
  \sqrt{\frac{M}{N}}(1 - \gntc) \cos\thxc - \sqrt{1 - \frac{M}{N}(1 - \gntc)^2} \sin\thxc
  \\ \leq \cos\varphi' \leq \\
  \sqrt{\frac{M}{N}}(1 + \gntc) \cos\thxc + \sqrt{1 - \frac{M}{N}(1 + \gntc)^2} \sin\thxc.
\end{multline}
Therefore, \cref{s:eq:cosyrange} is guaranteed to hold if
\begin{equation*}
\begin{aligned}
  (1 - \gntc) \cos\thxc - \sqrt{\frac{N}{M} - (1 - \gntc)^2} \sin\thxc &\geq 1 - \epsilon, \\
  (1 + \gntc) \cos\thxc + \sqrt{\frac{N}{M} - (1 + \gntc)^2} \sin\thxc &\leq 1 + \epsilon.
\end{aligned}
\end{equation*}%
These inequalities reveal that, in order to guarantee that the distortion of every vector within a chordal cone should have a distortion, $\epsilon$, that is the same order of magnitude as the distortion $\gntc$ at the central chord, the cone must be small;
its size $\thxc$ must be $\CO(\epsilon\sqrt{M/N})$, which will be justified by our choice of $\clsz$ in \cref{s:eq:saddle}).
Furthermore, if we make the assumption $M \ll N$,
both of these conditions reduce to the same inequality.
We find that the distortion of $\xv$ needs to satisfy the bound:
\begin{equation}\label{s:eq:distcondlong}
  \distpr(\xv) \leq \epsilon - \sqrt{\frac{N}{M}}\sin\thxc \equiv \gntc(\epsilon,\thxc).
\end{equation}
This is \cref{m:eq:distcondlong} of \cref{m:sec:distinter}\main.


\subsection{Bounding distortion within a single cell}\label{s:sec:distintra}

In this section we present the derivation of the function \cref{m:eq:distcondshort} from \cref{m:sec:distintra}\main.
This is an upper limit on the distortion of the tangent plane at the center of the cell such that \emph{all} tangent planes in this cell are guaranteed to have distortion less than $\epsilon$.

Suppose we have a $K$ dimensional subspace, $\Usp$, the tangent plane at a cell center,
and the largest principal angle with the tangent planes of points in the cell is bounded by $\thxs$.
The region around $\Usp$ in the Grassmannian where all subspaces $\Usp'$ have $\theta_a \leq \thxs$ forms a ``cone'' of subspaces that we will refer to as the tangential cone:
\begin{equation*}
  \cones = \set{\Usp' \in \grass_{K,N}}{\angle(\Usp, \Usp') \leq \thxs},
\end{equation*}%
where $\angle(\Usp, \Usp')$ denotes the set of principal angles between the subspaces $\Usp$ and $\Usp'$.
These subspaces are described schematically in \cref{m:fig:celltang}\ref{m:fig:concepttang}\main.

How small do we have to make the distortion at $\Usp$ so that we are \emph{guaranteed} that \emph{all} subspaces in this region (and thus all chords within the cell) have distortion at most $\epsilon$?
Call this quantity $\gnts(\epsilon,\thxs)$:
\begin{equation}\label{s:eq:gntsdef}
  \distpr(\Usp) \leq \gnts(\epsilon,\thxs)
  \qquad \implies \qquad
  \distpr(\Usp') \leq \epsilon \quad \forall \, \Usp' \in \cones.
\end{equation}
We will find such a function here.

The condition we need to guarantee is $\distpr(\Usp') \leq \epsilon$, which, according to \cref{s:eq:distspcsvd}, is equivalent to
\begin{equation*}
  \cos\varphi'\lmin \leq \sqrt{\frac{M}{N}}(1+\epsilon)
  \quad\text{and}\quad
  \cos\varphi'\lmax \geq \sqrt{\frac{M}{N}}(1-\epsilon),
\end{equation*}%
where $\varphi'\lmin$ and $\varphi'\lmax$ are the minimum and maximum principal angles between the projection subspace $\projsp$ and the subspace $\Usp'$.

This can be rewritten as
\begin{equation}\label{s:eq:distsin}
  \sin\varphi'\lmin \geq \sqrt{1-\frac{M}{N}(1+\epsilon)^2}
  \qquad \text{and} \qquad
  \sin\varphi'\lmax \leq \sqrt{1-\frac{M}{N}(1-\epsilon)^2}
\end{equation}
From Weyl's inequality in \cref{s:eq:weylang}, we see that the difference between the sines of principal angles of $\Usp$ and $\Usp'$ (relative to $\projsp$) is bounded by $\sin\thxs$, the largest possible principal angle \emph{between} $\Usp$ and $\Usp'$.
The worst case scenario is when the smallest principal angle between $\projsp$ and $\Usp'$ is made as small as possible, and the largest principal angle is made as large as possible, i.e.:
\begin{equation}\label{s:eq:projangworst}
  \sin\varphi'\lmin = \sin\varphi\lmin - \sin\thxs,
  \qquad
  \sin\varphi'\lmax = \sin\varphi\lmax + \sin\thxs.
\end{equation}
where $\varphi\lmin$ and $\varphi\lmax$ are the minimum and maximum principal angles between the projection subspace $\projsp$ and the  central subspace $\Usp'$.
The first of these could occur if $\projsp$ spanned the first $M$ Cartesian axes, $\Usp$ and $\Usp'$ spanned one vector in the $M$--$M+1$ plane and the $M+2$ to $M+K$ axes, with the vector in the $M$--$M+1$ plane for $\Usp$ making a larger angle with the $M$-axis than the vector for $\Usp'$.
In the limit as these two angles tend to zero, Weyl's inequality would be saturated and the first equation in \cref{s:eq:projangworst} would hold.
The second of these could occur if $\projsp$ spanned the first $M$ Cartesian axes, $\Usp$ and $\Usp'$ spanned one vector in the $M$-$M+1$ plane and the first $K-1$ axes, with the vector in the $M$-$M+1$ plane for $\Usp$ making a smaller angle with the $M$-axis than the vector for $\Usp'$.
In the limit as these two angles tend to zero, Weyl's inequality would be saturated and the second equation would hold.

This means that \cref{s:eq:distsin} is guaranteed to hold if
\begin{equation*}
  \sin\varphi\lmin \geq \sqrt{1-\frac{M}{N}(1+\epsilon)^2} \,+\,  \sin\thxs
  \quad \text{and} \quad
  \sin\varphi\lmax \leq \sqrt{1-\frac{M}{N}(1-\epsilon)^2} \,-\,  \sin\thxs.
\end{equation*}%
This means that we need $\Usp$ to satisfy
\begin{equation*}
\begin{aligned}
  \cos\varphi\lmin &\leq \sqrt{\frac{M}{N}(1+\epsilon)^2 - 2\sin\thxs\sqrt{1-\frac{M}{N}(1+\epsilon)^2} - \sin^2\thxs}, \\
  \cos\varphi\lmax &\geq \sqrt{\frac{M}{N}(1-\epsilon)^2 + 2\sin\thxs\sqrt{1-\frac{M}{N}(1-\epsilon)^2} - \sin^2\thxs}.
\end{aligned}
\end{equation*}%
Therefore, the distortion at $\Usp$ needs to satisfy the bounds:
\begin{equation*}
\begin{aligned}
  \distpr(\Usp) &\leq \brk{(1+\epsilon)^2 - 2\sin\thxs\sqrt{\frac{N}{M}\brk{\frac{N}{M}-(1+\epsilon)^2}} - \frac{N}{M}\sin^2\thxs}^{\frac{1}{2}} - 1
  \!\!&\!\!\!\!&\equiv \gnts^+(\epsilon,\thxs), \\
  \distpr(\Usp) &\leq 1 - \brk{(1-\epsilon)^2 + 2\sin\thxs\sqrt{\frac{N}{M}\brk{\frac{N}{M}-(1-\epsilon)^2}} - \frac{N}{M}\sin^2\thxs}^{\frac{1}{2}}
  \!\!&\!\!\!\!&\equiv \gnts^-(\epsilon,\thxs).
\end{aligned}
\end{equation*}%
These inequalities reveal that, in order to guarantee that the distortion of every subspace within a tangential cone should have a distortion, $\epsilon$, that is the same order of magnitude as the distortion $\gnts$ at the central tangent plane, the cone must be small;
its size $\thxs$ must be $\CO(M\epsilon/N)$, which will be justified by our choice of $\clsz$ in \cref{s:eq:cellopt}.
Furthermore, if we make the assumption $M \ll N$,
\begin{equation}\label{s:eq:needapprox}
 \gnts^\pm(\epsilon, \thxs) \approx \epsilon - \frac{N}{M} \, \sin\thxs.
\end{equation}
So we have
\begin{equation}\label{s:eq:distcondshort}
  \gnts(\epsilon,\thxs) = \min\brc{\gnts^+(\epsilon,\thxs),\gnts^-(\epsilon,\thxs)} \approx \epsilon - \frac{N}{M} \, \sin\thxs.
\end{equation}
This is \cref{m:eq:distcondshort} from \cref{m:sec:distintra}\main.
This expression for $\gnts(\epsilon,\thxs)$, when inserted into \cref{s:eq:gntsdef}, yields a potentially overly strict upper bound on distortion at the central plane $\distpr(\Usp)$, sufficient to guarantee that the distortion at all planes $\Usp'$ in the tangential cone $\cones$ is less than $\epsilon$.
However, this condition, while sufficient, may not be necessary; other potentially tighter bounds on perturbations of singular values could lead to less strict requirements on the distortion at the central tangent plane required to preserve the entire tangential cone.


\section{Geometry of Gaussian random manifolds}\label{s:sec:randman}

In this section, we will compute the diameters of the cells, the distance between their centers and the maximum principal angle between the tangent planes at cell centers and all tangent planes in the same cell for a class of random manifolds that we will now describe.

We will consider $K$ dimensional Gaussian random submanifolds, $\CM$, of $\R^N$, described by embeddings: $\ec^i = \phi^i(\ic^\alpha)$, as in \cref{m:sec:randmanmod}.
The Cartesian coordinates for the ambient space $\R^N$ are $\ec^i$ ($i,j = 1,\ldots,N$), and the intrinsic coordinates for the manifold are $\ic^\alpha$ ($\alpha,\beta = 1,\ldots,K$).
The embedding maps $\phi^i(\ic)$ are (multidimensional) Gaussian processes with
\begin{equation}\label{s:eq:randmangauss}
  \av{\phi^i(\ic)} = 0,
  \qquad
  \av{\phi^i(\ic_1) \phi^j(\ic_2)} = Q^{ij}(\ic_1 - \ic_2).
\end{equation}
Here we will consider the case when the kernel is given by
\begin{equation}\label{s:eq:gaussgausskernel}
  Q^{ij}(\Delta\ic) = \frac{\ell^2}{N} \, \delta^{ij} \e^{-\frac{\rho}{2}},
  \qquad
  \rho = \sum_\alpha \prn{\frac{\Delta\ic^\alpha}{\lambda_\alpha}}^2.
\end{equation}
The $\lambda_\alpha$ are the correlation lengths with respect to the coordinates $\ic^\alpha$.
We will partition the manifold into cells with sides of coordinate length $\clsz\lambda_\alpha$, as as follows:
the cell $\cell_{\vec{m}}$ has center $\ic_{\vec{m}}$:, where
\begin{equation}\label{s:eq:celldef}
  m^\alpha \clsz\lambda_\alpha \leq \ic^\alpha < (m^\alpha+1) \clsz\lambda_\alpha,
  \qquad
  \ic^\alpha_{\vec{m}} = \prn{m^\alpha + \frac{1}{2}}\clsz\lambda_\alpha.
\end{equation}
We will choose values of $\clsz$ that are small but nonzero (see \cref{s:eq:saddle} and \cref{s:eq:cellopt}).


\subsection{Separation and size of cells}\label{s:sec:cellsep}

In this section, we present the derivation of \cref{m:eq:gaussgausschord,m:eq:cellchang} in \cref{m:sec:cellsep}\main.
These were formulae for the Euclidean distance between two points on the manifold which lead to the diameter of a cell and the distance between cell centers, \ie the quantities $\cldm$ and $x$ that determine $\thx$ in \cref{s:eq:distcondlong}.

The squared distance between two points on the manifold is given by
\begin{equation*}
  \nrm{\xv_1 - \xv_2}^2 = \sum_i\brk{\phi^i(\ic_1) - \phi^i(\ic_2)}^2
    = \phi^i(\ic_1) \phi^i(\ic_1) + \phi^i(\ic_2) \phi^i(\ic_2) - 2 \phi^i(\ic_1) \phi^i(\ic_2),
\end{equation*}%
where the sums over the repeated $i$ index are implicit (Einstein summation convention).

Contractions of the $i,j$ indices are sums over $N$ quantities that are each $\sim \CO(1/N)$ (see \cref{s:eq:gaussgausskernel}).
In the limit of large $N$, they are expected to be self-averaging, \ie they can be replaced by their expectations, with the probability of $\CO(1)$ fractional deviations being suppressed exponentially in $N$ due to their standard deviation being $\sim \CO(1/\sqrt{N})$.
However, there is a large number of cells to consider, so the largest deviation will typically be significantly larger.
If the we have $n$ iid Gaussian random variables with standard deviation $\sigma$, the expected value of the maximum deviation from the mean for any of these random variables is $\sigma \sqrt{2\ln n}$.
In the situations considered in this section, the number of \emph{pairs} of cells is $\frac{\V}{2\clsz^{K}}\prn{\frac{\V}{\clsz^{K}}-1}$, where $\V = \prod_\alpha L_\alpha / \lambda_\alpha$.
However, cells that lie within a correlation length, $\lambda_\alpha$, of each other will be correlated due to the smoothness of the manifold.
We can group the cells into clusters whose sides are of length $\lambda_\alpha$, so that the clusters are approximately independent of each other (referred to as ``correlation cells'' in \cref{m:sec:complexity}\main).
Thus, the effective number of independent samples will be $\CO(\V^2)$.
This means that the maximum fractional deviation will be small when:
\begin{equation}\label{s:eq:selfavgvalidpair}
  \sqrt{\frac{4\ln\V}{N}} \ll 1
  \qquad \means \qquad
  \V \ll \e^{-\frac{N}{4}}.
\end{equation}
As $\log\V \sim \CO(K\log(L/\lambda))$, this requires $N \gg K\log(L/\lambda)$.

This means that
\begin{equation*}
  \nrm{\xv_1 - \xv_2}^2 = 2 \prn{Q^{ii}(0) - Q^{ii}(\ic_1 - \ic_2)}.
\end{equation*}%

We are going to assume that the kernel $Q(\Delta\ic)$ takes the form given in \cref{s:eq:gaussgausskernel}.
This means that
\begin{equation}\label{s:eq:gaussgausschord}
  \nrm{\xv_1 - \xv_2}^2 = 2\ell^2 \prn{1 - \e^{-\frac{\rho}{2}}},
\end{equation}
with the probability of $\CO(1)$ fractional deviations suppressed exponentially in $N$.

As we partition the manifold into the cells described in \cref{s:eq:celldef}, this means that the distance between the centers of two cells is given by
\begin{equation*}
  x_{\vec{m}\vec{n}}^2 = \nrm{\xv_{\vec{m}} - \xv_{\vec{n}}}^2 = 2 \ell ^2 \prn{1 - \e^{-\frac{\clsz^2 \nrm{\vec{m} - \vec{n}}^2}{2}}}.
\end{equation*}%
The largest distance from the center of the cell to any other point in the cell will be to one of the corners:
\begin{equation}\label{s:eq:cellsize}
  \max_{\xv \in \cell_{\vec{m}}} \nrm{\xv_{\vec{m}} - \xv}^2 = 2 \ell^2 \prn{1 - \e^{-\frac{\clsz^2K}{8}}} \approx \frac{\clsz^2\,\ell^2{K}}{4},
\end{equation}
where we assume that $\clsz \ll 1/\sqrt{K}$, this will be justified by our choice of $\clsz$ in \cref{s:eq:saddle}.
Thus, cells can be enclosed in balls of radius $\frac{\cldm}{2} \equiv \sqrt{2} \ell \prn{1 - \e^{-\frac{\clsz^2K}{8}}}^{\frac{1}{2}} \approx \frac{\clsz\,\ell\sqrt{K}}{2}$.
This gives the maximum angle between the central chord and any other chord between the two cells:
\begin{equation}\label{s:eq:cellchang}
  \sin\thxc = \frac{\cldm}{x} \approx \clsz \sqrt{\frac{K/2}{1 - \e^{-\frac{\clsz^2 \nrm{\vec{m} - \vec{n}}^2}{2}}}}.
\end{equation}
%


\subsection{Curvature of cells}\label{s:sec:cellcurv}

In this section, we present the derivation of \cref{m:eq:gaussgausspang,m:eq:cellapprox} in \cref{m:sec:cellcurv}\main.
These were formulae for quantities related to the principal angles between tangent planes.
This lead to an upper bound on $\sin\thxs$, the largest principal angle between the tangent plane at the center of a cell, $\Usp_{\vec{m}}$, and any other tangent plane in the same cell, the quantity that appears in \cref{s:eq:distcondshort}.

In \cref{s:eq:anglesvd}, we saw that the cosines of the principal angles are given by the singular values of the matrix $\U\trans\U'$, where the columns of $\U$ and $\U'$ are orthonormal bases for the tangent planes.
To do this, we can first construct an orthonormal basis, $e_a^\alpha$, for the intrinsic tangent space of $\CM$ (a vielbein or tetrad) under the induced metric, $h_{\alpha\beta}$, and pushing it forward to the ambient space:
\begin{equation}\label{s:eq:vielbein}
  h_{\alpha\beta}(\ic) = \phi^i_\alpha(\ic) \phi^i_\beta(\ic), \qquad
  h_{\alpha\beta} \, e_a^\alpha \, e_b^\beta = \delta_{ab}, \qquad
  U_{ia} = \phi^i_\alpha \, e_a^\alpha.
\end{equation}
where sums over repeated $i,j$ and $\alpha,\beta$ indices are implicit.
The vielbein $e_a^\alpha$ can be thought of as an inverse square root of the metric $h_{\alpha\beta}$.
Therefore, the cosines of the principal angles are the singular values of
\begin{equation}\label{s:eq:embedcosnorm}
\begin{aligned}
  \brk{\U\trans\U'}_{ab}
    &= e_a^\alpha(\ic) \phi^i_\alpha(\ic) \; \phi^i_\beta(\ic') e_b^\beta(\ic') \\
    &= \brk{\phi^i_a \phi^i_\alpha(\ic)}^{-\frac{1}{2}} \phi^j_\alpha(\ic) \;
       \phi^j_\beta(\ic') \brk{\phi^k_\beta \phi^k_b(\ic')}^{-\frac{1}{2}}.
\end{aligned}
\end{equation}

Like in the \hyperref[s:sec:cellsep]{previous section}, contractions of the $i,j$ indices are sums over $N$ quantities that are $\CO(1/\sqrt{N})$.
However, in the situations considered in this section we consider single cells rather than pairs of cells.
As in the \hyperref[s:sec:cellsep]{previous section}, the number of approximately independent clusters of cells is $\V$.
Therefore, the condition analogous to \cref{s:eq:selfavgvalidpair} is
\begin{equation}\label{s:eq:selfavgvalidsingle}
    \sqrt{\frac{2\ln\V}{N}} \ll 1
  \qquad \means \qquad
  \V \ll \e^{-\frac{N}{2}}.
\end{equation}
As before, in the limit of $N \gg K\log(L/\lambda)$, these quantities are self-averaging, \ie they can be replaced by their expectations, with the probability of $\CO(1)$ fractional deviations being suppressed exponentially in $N$.

This means that
\begin{equation*}
   \brk{\U\trans\U'}_{ab} = - [-Q^{ii}_{a\alpha}(0)]^{-\frac{1}{2}} {Q^{jj}_{\alpha\beta}(\ic-\ic')} [-Q^{kk}_{\beta b}(0)]^{-\frac{1}{2}},
\end{equation*}%
where the following relation was used:
\begin{equation*}
  \av{\phi^i_\alpha(\ic) \phi^j_\beta(\ic')} = - Q^{ij}_{\alpha\beta}(\ic-\ic').
\end{equation*}%
%

We are going to assume that the kernel, $Q(\Delta\ic)$ in \cref{s:eq:randmangauss}, takes the form of \cref{s:eq:gaussgausskernel}.
This means that
\begin{equation}\label{s:eq:gaussgaussmet}
  Q^{ii}_{\alpha\beta}(\Delta\ic) = \ell^2 \prn{\frac{\ic^\alpha \ic^\beta}{(\lambda_\alpha \lambda_\beta)^2} - \frac{\delta_{\alpha\beta}}{\lambda_\alpha^2}} \e^{-\frac{\rho}{2}},
  \quad
  -Q^{ii}_{\alpha\beta}(0) = \frac{\ell^2}{\lambda_\alpha^2} \delta_{\alpha\beta},
  \quad
  [-Q^{ii}_{a\alpha}(0)]^{-\frac{1}{2}} = \frac{\lambda_\alpha}{\ell} \delta_{a\alpha}.
\end{equation}
So we have
\begin{equation*}
  [\U\trans\U']_{ab} = \prn{\delta_{ab} - \frac{\Delta\ic^a \Delta\ic^b}{\lambda_a \lambda_b}} \e^{-\frac{\rho}{2}},
\end{equation*}
and therefore
\begin{equation}\label{s:eq:gaussgaussfrob}
  \cos\theta_a = \e^{-\rho/2} \quad \text{for} \quad a < K,
  \qquad
  \cos\theta_K = \abs{1 - \rho} \e^{-\rho/2}.
\end{equation}
with the probability of $\CO(1)$ fractional deviations suppressed exponentially in $N$,

Looking at \cref{s:eq:gaussgaussfrob} and \cref{s:eq:celldef}, we see that the largest principal angle with any tangent plane in the cell and $\Usp_{\vec{m}}$ will occur with high probability in one of the corners, where $\rho = \clsz^2 K / 4$.
Then
\begin{equation*}
  \max_{\Usp' \in \gauss\cell_{\vec{m}}} \sin \angle(\Usp_{\vec{m}},\Usp')
    = \max\brc{\sqrt{1 - \e^{-\frac{\clsz^2 K}{4}}},
      \sqrt{1 - \prn{1 - \tfrac{\clsz^2 K}{4}}^2 \e^{-\frac{\clsz^2 K}{4}}}}
    \equiv \sin\thxs,
\end{equation*}%
where $\sin \angle(\Usp_{\vec{m}},\Usp')$ denotes the set of sines of principal angles between the subspaces $\Usp$ and $\Usp'$.
For $\clsz \ll 1/\sqrt{K}$, which will be justified by our choice of $\clsz$ in \cref{s:eq:cellopt}, this can be approximated as
\begin{equation}\label{s:eq:cellapprox}
  \sin\thxs \approx \frac{\clsz}{2} \, \sqrt{3K}.
\end{equation}
%

%

In \cref{m:sec:strat,m:sec:distintra,m:sec:short}\main, we claimed that any short chord connecting two points in the same cell would be approximately parallel to some tangent vector for small enough $\clsz$.
We will now justify that claim.
Consider two points in the same cell, $\ec^i_1 = \phi^i(\ic_1)$ and $\ec^i_2 = \phi^i(\ic_2)$.
First, let us try to find a unit tangent vector $\uv$ at a point $\ic_3$ that has the largest possible overlap with $\xv_1 - \xv_2$.

Any unit tangent vector can be written as $u^i = \phi^i_\alpha s^\alpha$, where $h_{\alpha\beta} s^\alpha s^\beta = 1$.
Using the same self-averaging arguments as above, we have:
\begin{equation*}
\begin{aligned}
  (\xv_1 - \xv_2) \cdot \uv
       &= \brk{\phi^i(\ic_1) - \phi^i(\ic_2)} \phi^i_\alpha(\ic_3) s^\alpha \\
       &= \brk{ - Q^{ii}_\alpha(\ic_1 - \ic_3) + Q^{ii}_\alpha(\ic_2 - \ic_3)} s^\alpha \\
       &= \sum _\alpha \frac{\ell^2 \brk{ (\ic_1 - \ic_3)^\alpha \e^{-\rho_{13}/2} - (\ic_2 - \ic_3)^\alpha \e^{-\rho_{23}/2}} s^\alpha}{\lambda_\alpha^2}.
\end{aligned}
\end{equation*}
This is maximized when $\ic_3 = \frac{\ic_1 + \ic_2}{2}$ and $s^\alpha = \frac{\ic_1^\alpha - \ic_2^\alpha}{\ell\sqrt{\rho}}$.
For this choice of $\ic_3$ and $\uv$, making use of \cref{s:eq:gaussgausschord}, the angle between $\xv_1 - \xv_2$ and $\uv$ is
\begin{equation*}
  \cos\psi = \frac{(\xv_1 - \xv_2) \cdot \uv}{\nrm{\xv_1 - \xv_2}}
    = \frac{\sqrt{\rho} \,\e^{-\rho/8}}{\sqrt{2\brk{1 - \e^{-\rho/2}}}}
    = \sqrt{\frac{\rho/4}{\sinh(\rho/4)}}.
\end{equation*}
This is a decreasing function of $\rho$.
From \cref{s:eq:cellsize}, the largest $\rho$ can be for two points in the same cell is $\clsz^2 K$.
This means that, for $\clsz \ll 1/\sqrt{K}$ (justified by \cref{s:eq:cellopt}), we have
\begin{equation}\label{s:eq:shorttangent}
  \psi\lmax \approx  \frac{\clsz^2 K}{4\sqrt{6}} \ll \thxs.
\end{equation}
In this regime, we are justified in claiming that any short chord connecting two points in the same cell would be approximately parallel to some tangent vector.


\subsection{Condition number and geodesic covering regularity}\label{s:sec:BW}

The bounds on random projections of smooth manifolds found by \cite{Baraniuk2009JLmfld} contain two geometric properties of the manifold: the geodesic covering regularity, $R$, and the inverse condition number, $\tau$.
We will find a lower bound on $R$ and an upper bound on $\tau$.
As they appear in the result of \cite{Baraniuk2009JLmfld} in the form $\ln(R/\tau)$, we will underestimate their bound on the number of projections required to achieve distortion $\epsilon$ with probability $1 - \delta$.
As we see in \cref{m:fig:distman}, this underestimate is 2 orders of magnitude greater than our result and 4 orders of magnitude greater than the simulations.

The geodesic covering number, $G(r)$, is the minimum number of points on the manifold needed such that everywhere on the manifold is within geodesic distance $r$ of at least one of the points.
Equivalently, it is the minimum number of geodesic balls of radius $r$ needed to cover $\CM$.
The geodesic covering regularity is then defined as the smallest $R$ such that $G(r) \leq K^{K/2} \vol(\CM) (R/r)^K$.

The total volume of all the covering balls must be at least as large as the volume of $\CM$.
This means that $G(r)$ must be at least as large as the ratio of the volumes of $\CM$ and one of these balls.
From \cref{s:eq:gaussgaussmet} we see that, when self averaging is valid, the induced metric is given by $h_{\alpha\beta} = \delta_{\alpha\beta} (\ell/\lambda_\alpha)^2$.
This means that geodesics are straight lines in the $\ic$-coordinates, the length of a geodesic is $\ell\sqrt{\rho}$, the volume of the manifold is $\ell^K \V$ and the volume of a geodesic ball is the usual Euclidean expression in terms of $r$.
Combined with the inequality $\Gamma(n+1) > (n/\e)^n$, this leads to:
\begin{equation}\label{s:eq:geocover}
  \frac{K^{\frac{K}{2}} \vol(\CM)}{r^K} \, R^K \geq G(r)
     \geq \frac{\vol(\CM)}{\vol(\B^K_r)}
     = \frac{\Gamma\prn{\frac{K}{2} + 1} \vol(\CM)}{\pi^{\frac{K}{2}} r^K}
     \geq \frac{K^{\frac{K}{2}} \vol(\CM)}{r^K} \frac{1}{\prn{2\pi\e}^{\frac{K}{2}}}.
\end{equation}
Therefore the geodesic covering regularity satisfies the inequality:
\begin{equation}\label{s:eq:geocovreg}
  R \geq \frac{1}{\sqrt{2\pi\e}}.
\end{equation}

\begin{figure}[tbp]
  \centering
  \begin{myenuma}
    \item\aligntop{\includegraphics[height=0.35\linewidth]{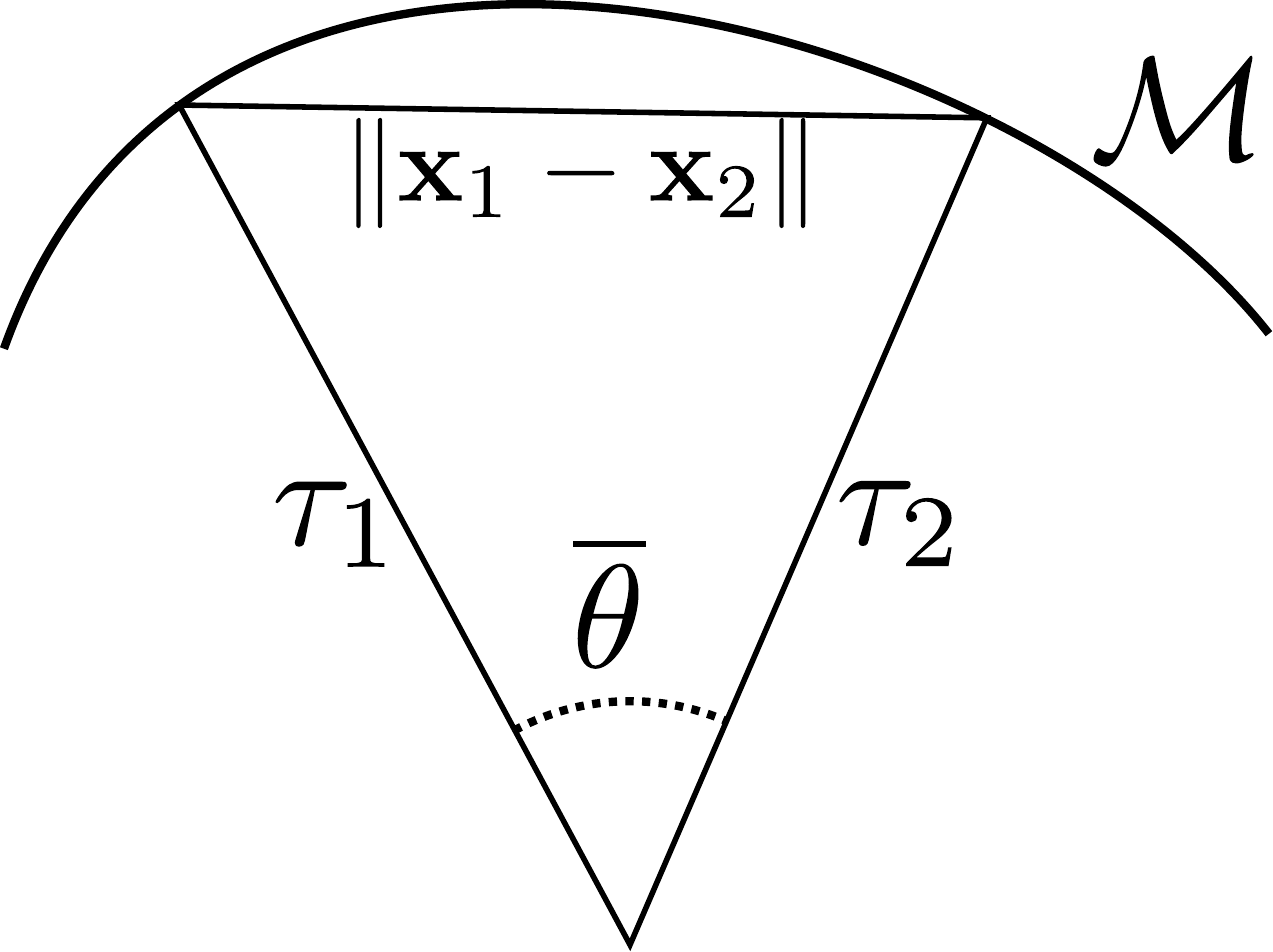}}\label{s:fig:cond_gen}
    \hspace{0.01\linewidth}
    \item\hspace{0.01\linewidth}\aligntop{\includegraphics[height=0.35\linewidth]{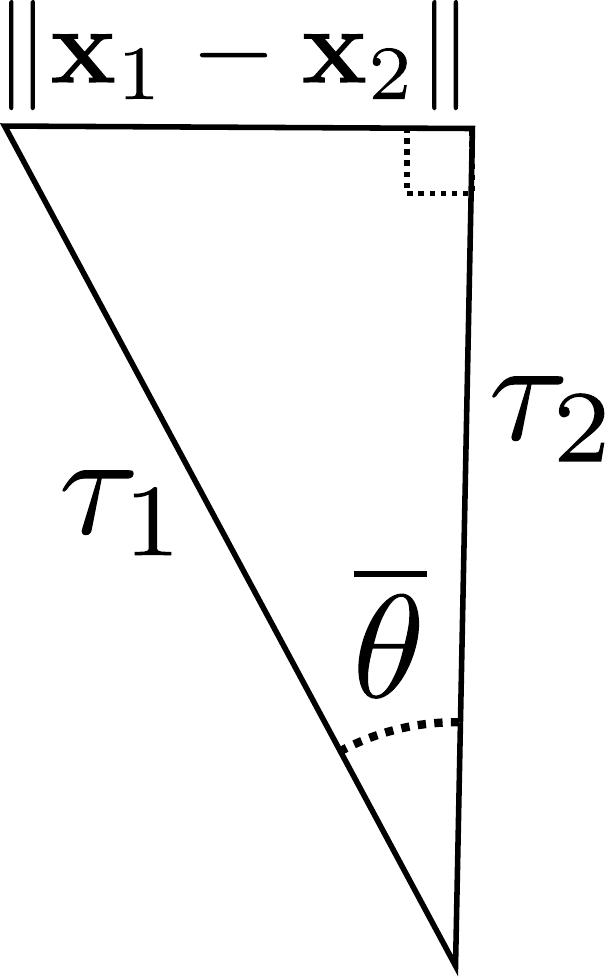}}\label{s:fig:cond_max}
  \end{myenuma}
  \caption{A point where two normal balls touch.
  (\ref{s:fig:cond_gen}) Two points, $\xv_1$ and $\xv_2$, on the manifold $\CM$ have normal spaces that touch at distances $\tau_1$ and $\tau_2$ from the respective points on $\CM$, where the lines joining $\xv_1$ and $\xv_2$ to the touching point cross at angle $\overline{\theta}$.
  (\ref{s:fig:cond_max}) The largest that $\tau_1$ and $\tau_2$ can possibly be, for fixed $\nrm{\xv_1 - \xv_2}$ and $\overline{\theta}$, is when the chord ${\xv_1 - \xv_2}$ meets one of the two lines joining $\xv_1$ and $\xv_2$ to the touching point at a right angle.
  }\label{s:fig:cond_num}
\end{figure}

The inverse condition number is the largest distance in the ambient space that we can expand all points of the manifold in all perpendicular directions before two of these normal balls touch.
Such a situation is depicted in \cref{s:fig:cond_num}\ref{s:fig:cond_gen}.
At the first instance where the normal balls touch, $\tau$ would be the larger of the two perpendicular distances from the manifold ($\tau_1$ and $\tau_2$), because extending the balls by only the smaller of the two distances would not result in them touching.
Let the two points on the manifolds be $\xv_1$ and $\xv_2$ (the centers of the two balls), and let $\overline{\theta}$ be the angle between the two vectors joining the point of first intersection to $\xv_1$ and $\xv_2$.
For fixed $\nrm{\xv_1 - \xv_2}$ and $\overline{\theta}$, the furthest one could extend $\tau_1$ is depicted in \cref{s:fig:cond_num}\ref{s:fig:cond_max} (regardless of whether or not this is the first intersection of two normal balls), as lengthening $\tau_1$ any further would make it impossible for a chord of length $\nrm{\xv_1 - \xv_2}$ to reach the other line.
In that case, the larger of the two lengths is $\nrm{\xv_1 - \xv_2}/\sin\overline{\theta}$.

Therefore
\begin{equation}\label{s:eq:cond_min}
  \tau \leq \min_{\xv_1,\xv_2 \in \CM} \brc{\frac{\nrm{\xv_1 - \xv_2}}{\sin\overline{\theta}}}
       \leq \min_{\xv_1,\xv_2 \in \CM} \brc{\frac{\nrm{\xv_1 - \xv_2}}{\sin\overline{\theta}\lmin}}.
\end{equation}
Where $\overline{\theta}\lmin$ is the smallest \emph{nonzero} principal angle between the normal planes at $\xv_1$ and $\xv_2$.
As the normal spaces have dimension $N - K$, they must share at least $N - 2K$ directions (assuming $N > 2K$).
However, the vectors joining the point of closest contact to $\xv_1$ and $\xv_2$ will have no components in these shared directions: if they did, we could find a closer point of contact that is also in the normal space at both points simply by eliminating these components.

Following \cref{s:eq:projev}, with $\U_\perp$ denoting an orthonormal basis for the normal space and noting that $\U \U\trans + \U_\perp \U_\perp\trans = \I$, we can write:
\begin{equation*}
\begin{aligned}
  \U_\perp \U\trans_\perp - \U'_\perp \U'\transp_\perp
     =  (\I - \U \U\trans) -  (\I - \U' \U'\transp)
     = \U' \U'\transp - \U \U\trans.
\end{aligned}
\end{equation*}%
From the discussion below \cref{s:eq:projev}, we know that the singular values of these differences of projection operators are the sines of principal angles.
Therefore, the nonzero principal angles satisfy $\theta_a = \overline{\theta}_a$.
Comparing \cref{s:eq:gaussgausschord,s:eq:gaussgaussfrob}, we see that
\begin{equation*}
  \nrm{\xv_1 - \xv_2} = \sqrt{2}\ell \left\{
\begin{aligned}
     &\sin \overline{\theta}\lmin, &
     \rho &\leq 2,\\
     &\sin \overline{\theta}\lmax, &
     \rho &> 2. \\
\end{aligned}\right.
\end{equation*}
Therefore, from \cref{s:eq:cond_min}, we have:
\begin{equation}\label{s:eq:cond_bnd}
  \tau \leq \sqrt{2}\ell.
\end{equation}

We can also compute an underestimate of the bound derived by \cite{verma2011note} (note that this was a bound for distortion of curve length, which is a lower bound on the distortion of chord lengths, as discussed in \cref{s:sec:distort}).
This bound also involves the geodesic covering number, $G(r)$, for which we found an upper bound in \cref{s:eq:geocover}.
In addition, the value of $r$ used by \cite{verma2011note} depends on a uniform upper bound on the norm of the second fundamental form:
\begin{equation}\label{s:eq:secformbnd}
  \frac{1}{\tilde{\tau}} \geq  S^i_{\alpha\alpha}(\ic) \, \eta_i \, u^\alpha u^\beta
    \quad \forall \; \ic, \boldsymbol{\eta}, \mathbf{u} \quad \text{s.t.} \quad
  \eta_i \eta_i = h_{\alpha\beta} \, u^\alpha u^\beta = 1,
\end{equation}
where the induced metric, $h_{\alpha\beta}$, was defined in \cref{s:eq:vielbein}.
This quantity can be computed for our ensemble of manifolds in \cref{s:eq:randmangauss,s:eq:gaussgausskernel}, in the regime of \eqref{s:eq:selfavgvalidsingle} where self averaging is valid.
\begin{equation}\label{s:eq:secformnrm}
\begin{aligned}
  \frac{1}{\tilde{\tau}^2} &= \max_{\mathbf{u}} \brc{S^i_{\alpha\beta} S^i_{\gamma\delta}
                                                   u^\alpha u^\beta u^\gamma u^\delta} \\
   &= \max_{\mathbf{u}} \brc{\bar{h}^{ij} \phi^i_{\alpha\beta} \phi^j_{\gamma\delta}
                             u^\alpha u^\beta u^\gamma u^\delta} \\
   &= \max_{\mathbf{u}} \brc{Q^{ii}_{\alpha\beta\gamma\delta}(0) u^\alpha u^\beta u^\gamma u^\delta} \\
   &= \max_{\mathbf{u}} \brc{ \prn{\frac{h_{\alpha\beta} h_{\gamma\delta} + h_{\alpha\gamma} h_{\beta\delta} + h_{\alpha\delta} h_{\beta\gamma}}{\ell^2}} u^\alpha u^\beta u^\gamma u^\delta}\\
   &= \frac{3}{\ell^2},
\end{aligned}
\end{equation}
where $\bar{h}^{ij}$ is the projection operator normal to the manifold, defined as $\bar{h}^{ij} = \delta^{ij} - h^{ij}$, where $h^{ij} = \phi^i_\alpha h^{\alpha\beta} \phi^j_\beta$ is the projection operator tangent to the manifold, and $h^{\alpha\beta}$ is the inverse of the induced metric, $h_{\alpha\beta}$.


\section{Overall logic}\label{s:sec:logic}

In this section we combine the results of \cref{s:sec:boundcell} and \cref{s:sec:randman} to find an upper limit on the probability that any chord has distortion greater than $\epsilon$ under a random projection $\proj$.

We will do this in two steps.
First, in \cref{s:sec:long}, we will bound this probability for the long chords between different cells.
Then, in \cref{s:sec:short} we will do the same for the short chords within single cells.

%

\subsection{Long chords}\label{s:sec:long}

Here we combine the results of \cref{s:sec:distinter} and \cref{s:sec:cellsep} to find an upper bound on the probability that any long intercellular chord has distortion greater than $\epsilon$ under a random projection $\proj$.

We consider Gaussian random manifolds $\CM$ of the type described in \cref{s:sec:randman}, partitioned into cells $\cell_{\vec{m}}$ with centers $\ic_{\vec{m}}$, where each side has coordinate length $\clsz\lambda_\alpha$, as described in \cref{s:eq:celldef}.
We have
\begin{equation*}
\begin{aligned}
  \delta\lng = \Pr\brk{\distpr([\CM-\CM]\lng) > \epsilon}
    &= \Pr\brk{\bigcup_{\vec{m} \neq \vec{n}}\brc{\distpr(\cell_{\vec{m}} - \cell_{\vec{n}}) > \epsilon}} \\
    &\leq \sum_{\vec{m} \neq \vec{n}} \Pr\brk{\distpr(\cell_{\vec{m}} - \cell_{\vec{n}}) > \epsilon},
\end{aligned}
\end{equation*}%
where we used the union bound.

Now we will use the results of \cref{s:sec:cellsep}.
From \cref{s:eq:cellchang}, we saw that when $\clsz \ll 1/\sqrt{K}$, we have
\begin{equation*}
 \sin\thxc \approx \clsz \sqrt{\frac{K/2}{1 - \e^{-\frac{\clsz^2 \nrm{\vec{m} - \vec{n}}^2}{2}}}}.
\end{equation*}%
This quantity can be used to bound the distortion of all vectors between these cells, using the results of \cref{s:sec:distinter}.

As the function $\gntc(\epsilon,\thxc)$ defined in \cref{s:eq:distcondshort} has the property that $\distpr(\xv_{\vec{m}} - \xv_{\vec{n}}) < \gntc(\epsilon,\thxc)$ guarantees that $\distpr(\cell_{\vec{m}} - \cell_{\vec{n}}) < \epsilon$, it follows that
\begin{equation*}
  \Pr\brk{\distpr(\cell_{\vec{m}} - \cell_{\vec{n}}) > \epsilon} \leq \Pr\brk{\distpr(\xv_{\vec{m}} - \xv_{\vec{n}}) > \gntc(\epsilon,\thxc)}.
\end{equation*}%
A bound on this last quantity was computed in \cite{Johnson1984extensions,Dasgupta2003JLlemma}:
\begin{equation}\label{s:eq:JLlemma}
  \Pr\brk{\distpr(\xv) > \epsilon} \leq 2 \, \e^{-\frac{M}{2}\prn{\frac{\epsilon^2}{2} - \frac{\epsilon^3}{3}}}.
\end{equation}
We can ignore the $\epsilon^3$ term when $\epsilon \ll 1$.

Putting it all together, we find that:
\begin{equation}\label{s:eq:alllong}
  \delta\lng \lesssim \sum_{\vec{m}, \vec{n}}
     \exp\prn{ - \frac{M}{4} \brk{
        \epsilon - \clsz\sqrt{\frac{KN}{2M\prn{1 - \e^{-\frac{\clsz^2\nrm{\vec{m} - \vec{n}}^2}{2}}}}}
        }^2} .
\end{equation}
If we define $\vec{y}_{1,2} = \clsz\vec{m},\clsz\vec{n}$, when $\clsz \ll 1/(KNM\epsilon^2)^{1/4}$ the change in the summand across a cell, \eg when $m_1$ changes by 1 with all other $m_\alpha, n_\alpha$ fixed, is much smaller than the summand itself, so we can approximate the sum with an integral:
\begin{equation}\label{s:eq:longintegral}
    \delta\lng \lesssim \clsz^{-2K} \intd{\vec{y}_1} \intd{\vec{y}_2}
    \exp\prn{ - \frac{M}{4} \brk{
      \epsilon - \clsz\sqrt{\frac{KN}{2M\prn{1 - \e^{-\frac{\nrm{\vec{y}_1 - \vec{y}_2}^2}{2}}}}}
      }^2}.
\end{equation}
We can the make a change of variables to $\vec{y} = {\vec{y}_1 - \vec{y}_2}$, and use spherical coordinates with $y = \nrm{\vec{y}}$:
\begin{equation*}
\begin{aligned}
    \delta\lng &\lesssim \clsz^{-2K} \intd{\vec{y}_1} \intd{y}
       \frac{2\pi^{\frac{K}{2}}}{\Gamma(\frac{K}{2})} y^{K-1}
       \exp\prn{ - \frac{M}{4} \brk{
         \epsilon - \clsz\sqrt{\frac{KN}{2M\prn{1 - \e^{-\frac{y^2}{2}}}}}
         }^2} \\
    &= \frac{\pi^{\frac{K}{2}} \clsz^{-2K}}{\Gamma(\frac{K}{2})} \intd{\vec{y}_1} \int\!\! \frac{\dr\rho}{\rho}  \, \rho^{\frac{K}{2}}
    \exp\prn{-\frac{M}{4}\brk{\epsilon - \clsz\sqrt{\frac{KN}{2M\prn{1 - \e^{-\frac{\rho}{2}}}}}}^2}
\end{aligned}
\end{equation*}%
where we made the change of variables $\rho = y^2$ and $\frac{2\pi^{\frac{K}{2}}}{\Gamma(\frac{K}{2})}$ is the volume of the sphere $\Sp^{K-1}$.
When $M\epsilon^2 \gg 1$ and $K \gg 1$, we can perform the integral using the saddle point method (method of steepest descent) as the result is $\CO(M\epsilon^2,K)$, whereas first correction to this is $\CO(\log(M\epsilon^2,K))$ and all higher order corrections scale as negative powers of $(M\epsilon^2,K)$.
As the value of $\rho$ at the saddle point given below is $\CO(1)$, the difference in $\ic^\alpha$ for the relevant pairs of cells is $\CO(\lambda_\alpha)$.
Thus, so long as $L_\alpha \gg \lambda_\alpha$, boundary effects can be neglected and the integral over $\vec{y}_1$ produces a factor of $\V = \prod_\alpha \frac{L_\alpha}{\lambda_\alpha}$:
\begin{equation}\label{s:eq:saddledef}
    \delta\lng \lesssim \frac{\pi^{\frac{K}{2}} \clsz^{-2K} \V}{\Gamma(\frac{K}{2})} \exp\prn{-\min_\rho \brc{\frac{M}{4}\brk{\epsilon - \clsz\sqrt{\frac{KN}{2M\prn{1 - \e^{-\frac{\rho}{2}}}}}}^2 - \frac{K}{2}\ln\rho}}.
\end{equation}

We can minimize this expression over $\clsz$ to obtain the tightest bound, which leads to
\begin{equation}\label{s:eq:saddle}
\begin{aligned}
  \rho^* &
          = -1 - 2W_{-1}\prn{-\frac{1}{2\sqrt{\e}}} = 2.513,
  &
  \clsz^*_\conec &= \sqrt{\frac{M\rho^*\,\e^{-\frac{\rho^*}{2}}}{2KN}}\prn{\epsilon - \sqrt{\epsilon^2 - \frac{16K}{M}}}, \\
  &&
   \sin\thxc^* &\approx \frac{1}{2} \sqrt{\frac{M}{N}}\prn{\epsilon - \sqrt{\epsilon^2 - \frac{16K}{M}}},
\end{aligned}
\end{equation}
where $W_{-1}(x)$ is the $-1$ branch of the Lambert W-function: $W(x)\e^{W(x)} = x$.

Note that $\sin\thxc^* \sim \CO(\epsilon\sqrt{M/N})$, justifying the assumption made in deriving the approximate form of $\gntc(\epsilon,\thxc)$ in \cref{s:eq:distcondlong}, and $\clsz^*_\conec \sim \CO(\epsilon\sqrt{M/KN})$, justifying the small $\clsz$ assumption used to approximate $\thxc$ in \cref{s:eq:cellchang}.
With this value of $\clsz$, the condition for replacing the sum with an integral in \cref{s:eq:longintegral},
$\clsz \ll 1/(KNM\epsilon^2)^{1/4}$, becomes $KN \gg (M\epsilon^2)^3$.
When $\epsilon \sim \CO(\sqrt{K/M})$, this reduces to $N \gg K^2$.

Introducing $\mu = \frac{M\epsilon^2}{K}$, we have:
\begin{equation*}
\begin{aligned}
  \delta\lng &\lesssim& \!\!\!\!\!
            \exp \Bigg( - \frac{K}{8} \Bigg[&
             \mu + \sqrt{\mu(\mu - 16)}
             + 16\ln\prn{\frac{\mu - \sqrt{\mu(\mu - 16)}}{\sqrt{\mu}}}
             - \frac{8}{K}\ln\V \\&&& - 8\ln N
             + 4(\ln\rho^* - \rho^*) - 8 - 4\ln4\pi
             + \frac{8}{K}\ln\Gamma\!\prn{\frac{K}{2}}
             \Bigg]\Bigg) \\
   &<& \!\!\!\!
            \exp \Bigg( - \frac{M\epsilon^2}{4}
             &+ \ln\V  + K \ln \prn{\frac{N M \epsilon^2}{K}}
             + C_0
             - \ln\Gamma\!\prn{\frac{K}{2}}
             \Bigg),
\end{aligned}
\end{equation*}%
where $C_0 = \frac{\rho^*}{2} + \frac{1}{2} \ln ({\pi} / {\rho^*}) + 2 - 5\ln2 = -0.097651$.%
\footnote{We also used the fact that $\sqrt{\mu(\mu - c)} + c\ln\prn{\mu - \sqrt{\mu(\mu - c)}} > \mu + c\ln\frac{c}{2\sqrt{e}}$ for all $\mu > c > 0$.
This can be proved by noting that the derivative the left hand side \wrt $\mu$ is less than 1, and that the inequality is saturated in the limit $\mu \ra \infty$.
As this quantity is multiplied by a negative quantity in the exponent, replacing it with a smaller quantity will increase the exponential.
\label{s:ft:muineq}}


\subsection{Short chords}\label{s:sec:short}

Here we combine the results of \cref{s:sec:distintra} and \cref{s:sec:cellcurv} to find an upper bound on the probability that any short intracellular chord has distortion greater than $\epsilon$ under a random projection $\proj$.

As discussed in \cref{m:sec:strat}\main, short chords will be approximately parallel to some tangent vector when the size of a cell is much smaller than the length scale of curvature.
In \cref{s:eq:shorttangent}, we saw that this holds when $\clsz \ll 1/\sqrt{K}$, as also required for \cref{s:eq:cellapprox}.
The set of tangent planes at points in the cell $\cell_{\vec{m}}$ are denoted by $\gauss\,\cell_{\vec{m}}$, and the central tangent plane is denoted by $\Usp_{\vec{m}}$.
We have
\begin{equation}\label{s:eq:distprobmanshort}
\begin{aligned}
  \delta\shrt &= \Pr\brk{\distpr([\CM-\CM]\shrt) > \epsilon} \\
    &= \Pr\brk{\distpr(\gauss\CM) > \epsilon} \\
    &= \Pr\brk{\bigcup_{\vec{m}} \brc{\distpr(\gauss\cell_{\vec{m}}) > \epsilon}} \\
    &\leq \sum_{\vec{m}} \Pr\brk{\distpr(\gauss\cell_{\vec{m}}) > \epsilon} \\
    & = \clsz^{-K} \V\, \Pr\brk{\distpr(\gauss\cell_1) > \epsilon},
\end{aligned}
\end{equation}
where we used relation between short chords and tangent vectors to go from the first to the second line.
We used the union bound to go from the third to the fourth line.
To go from the fourth to fifth line, we used translational invariance and wrote the number of cells as $\clsz^{-K} \V$, where $\V = \prod_\alpha \frac{L_\alpha}{\lambda_\alpha}$.

In \cref{s:eq:cellapprox}, we saw that $\sin\thxs \approx \frac{\clsz}{2} \, \sqrt{3K}$.
As the function $\gnts(\epsilon,\thxs)$ defined in \cref{s:sec:distintra} has the property that $\distpr(\Usp_{\vec{m}}) < \gnts(\epsilon,\thxs)$ guarantees that $\distpr(\Usp') < \epsilon$ in the cell $\gauss\,\cell_{\vec{m}}$, it follows that
\begin{equation}\label{s:eq:distprobcelltang}
  \Pr\brk{\distpr(\gauss\,\cell_{\vec{m}}) > \epsilon} \leq \Pr\brk{\distpr(\Usp_{\vec{m}}) > \gnts(\epsilon,\thxs)}.
\end{equation}
A bound on this last quantity was computed by \citet[Lemma 5.1]{baraniuk2008simple}:
\begin{equation}\label{s:eq:distprobcent}
  \Pr\brk{\distpr(\Usp) > \epsilon} \leq
       2 \, \e^{- \frac{M}{16} \brk{\epsilon^2 - \epsilon^3/3}
         + 2 K \log\frac{12}{\epsilon}},
\end{equation}
where the $\epsilon^3$ term is negligible when $\epsilon \ll 1$.

Combining \cref{s:eq:distprobmanshort,,s:eq:distprobcelltang,s:eq:distprobcent}, we find that:
\begin{equation}\label{s:eq:allshort}
  \delta\shrt \lesssim \exp\prn{\ln\V - K\ln\clsz - \frac{M}{16} \brk{\epsilon - \frac{\clsz N \sqrt{3K}}{2M}}^2
      + 2 K \log\frac{12}{\epsilon}}.
\end{equation}
This is true for all $\clsz$, but we can minimize this over $\clsz$ to obtain the tightest bound:
\begin{equation}\label{s:eq:cellopt}
  \clsz^*_\cones = \frac{M\epsilon - \sqrt{M(M\epsilon^2 - 32K)}}{N\sqrt{3K}},
  \qquad \implies \qquad
  \sin\thxs^* \approx \frac{M\epsilon - \sqrt{M(M\epsilon^2 - 32K)}}{2N}.
\end{equation}
Note that $\sin\thxs^* \sim \CO(M\epsilon/N)$, justifying the assumptions made in deriving \cref{s:eq:needapprox}, and $\clsz^*_\cones \sim \CO(M\epsilon/\sqrt{K}N)$, justifying the small $\clsz$ approximation in \cref{s:eq:cellapprox} and above \cref{s:eq:distprobmanshort}.

Introducing $\mu = \frac{M\epsilon^2}{K}$, we have:
\begin{equation*}
\begin{aligned}
  \delta\shrt &\lesssim \exp \prn{ \ln\V - \frac{K}{32} \brk{\mu + \sqrt{\mu(\mu - 32)}}
          + K \brk{\frac{1}{2} + \ln\prn{\frac{ 144 \sqrt{3} N }{\epsilon \sqrt{K}
              \prn{\mu - \sqrt{\mu(\mu - 32)}} }} } } \\
          &< \exp\prn{-\frac{M\epsilon^2}{16} + \ln\V + K \ln\prn{\frac{9\sqrt{3}\,\e N}{\epsilon\sqrt{K}}} },
\end{aligned}
\end{equation*}%
where we used the inequality discussed in \cref{s:ft:muineq}.

Note that we used different values for $\clsz$ in \cref{s:eq:saddle} and \cref{s:eq:cellopt}, but as the most important differences between \cref{s:eq:alllong} and \cref{s:eq:allshort} are the coefficients of the terms proportional to $M$, this will not change the fact that $\delta\lng \ll \delta\shrt$.


\subsection{Comparison of theories and simulations}\label{s:sec:plotsep}

\begin{figure}[tbp]
  \centering
  \begin{myenuma}
    \item\aligntop{\includegraphics[height=0.32\linewidth]{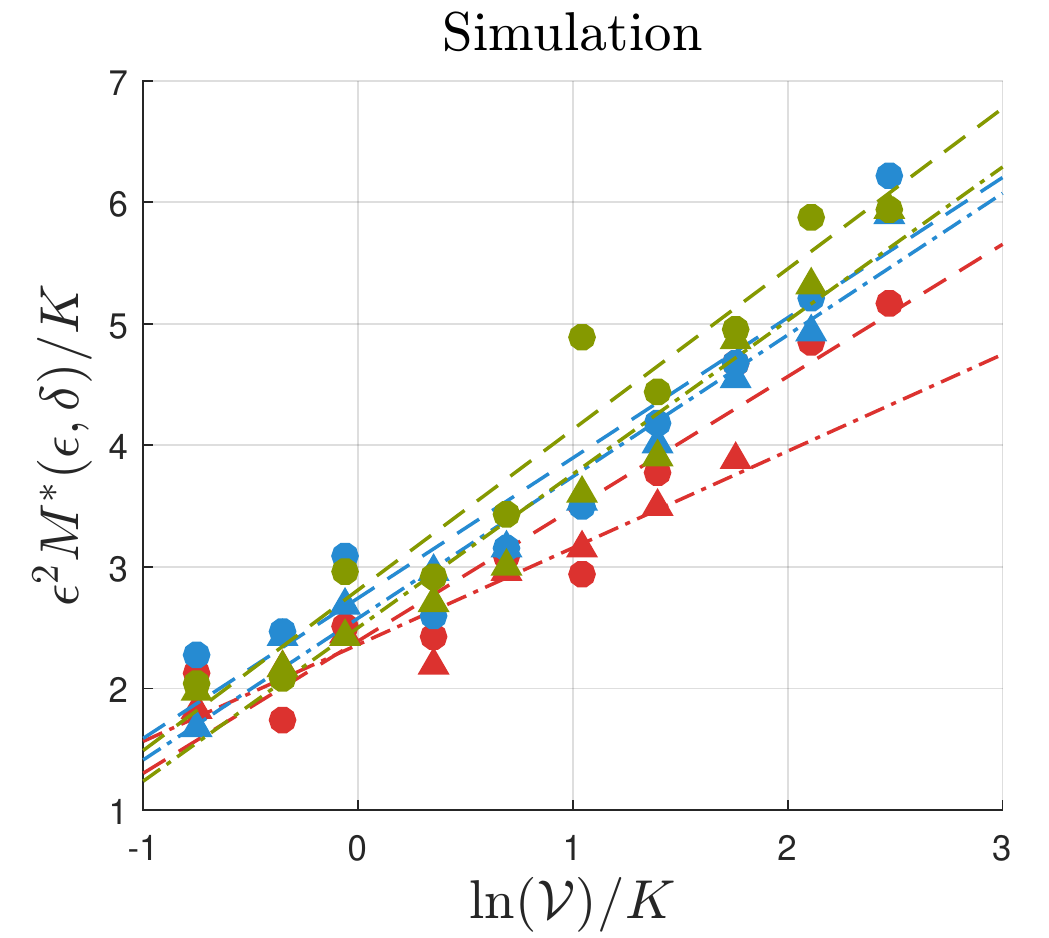}}\label{s:fig:distmanVsim}
    \item\aligntop{\includegraphics[height=0.32\linewidth]{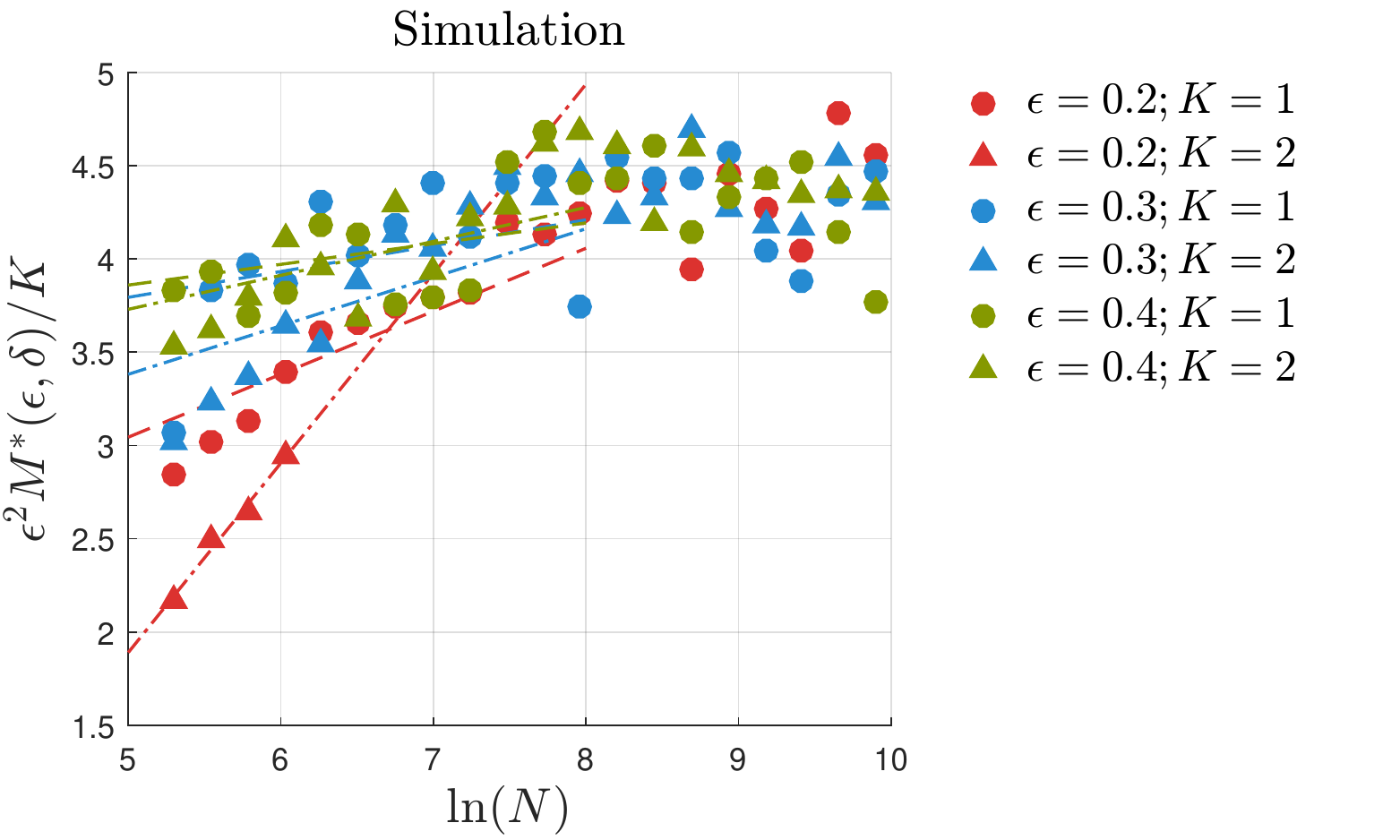}}\label{s:fig:distmanNsim}

    \vspace{0.02\linewidth}
    \item\aligntop{\includegraphics[height=0.31\linewidth]{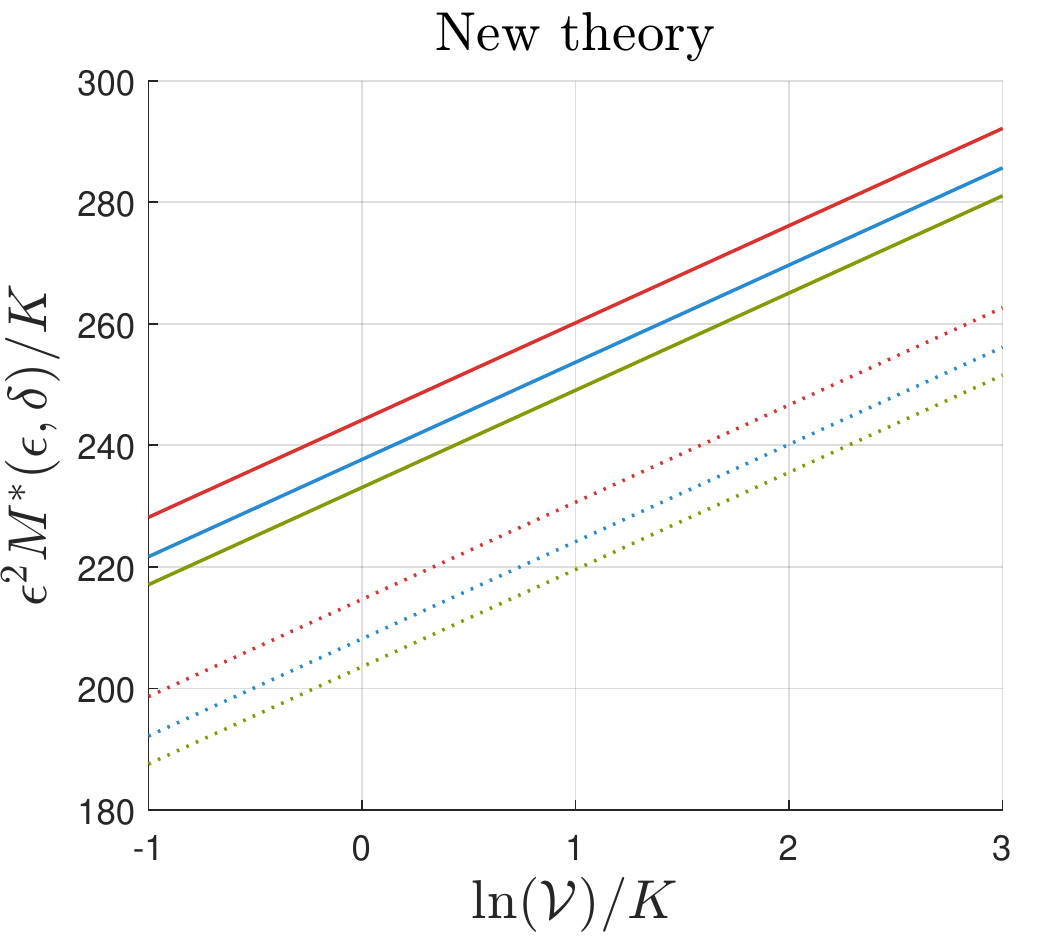}}\label{s:fig:distmanVlgg}
    \item\aligntop{\includegraphics[height=0.31\linewidth]{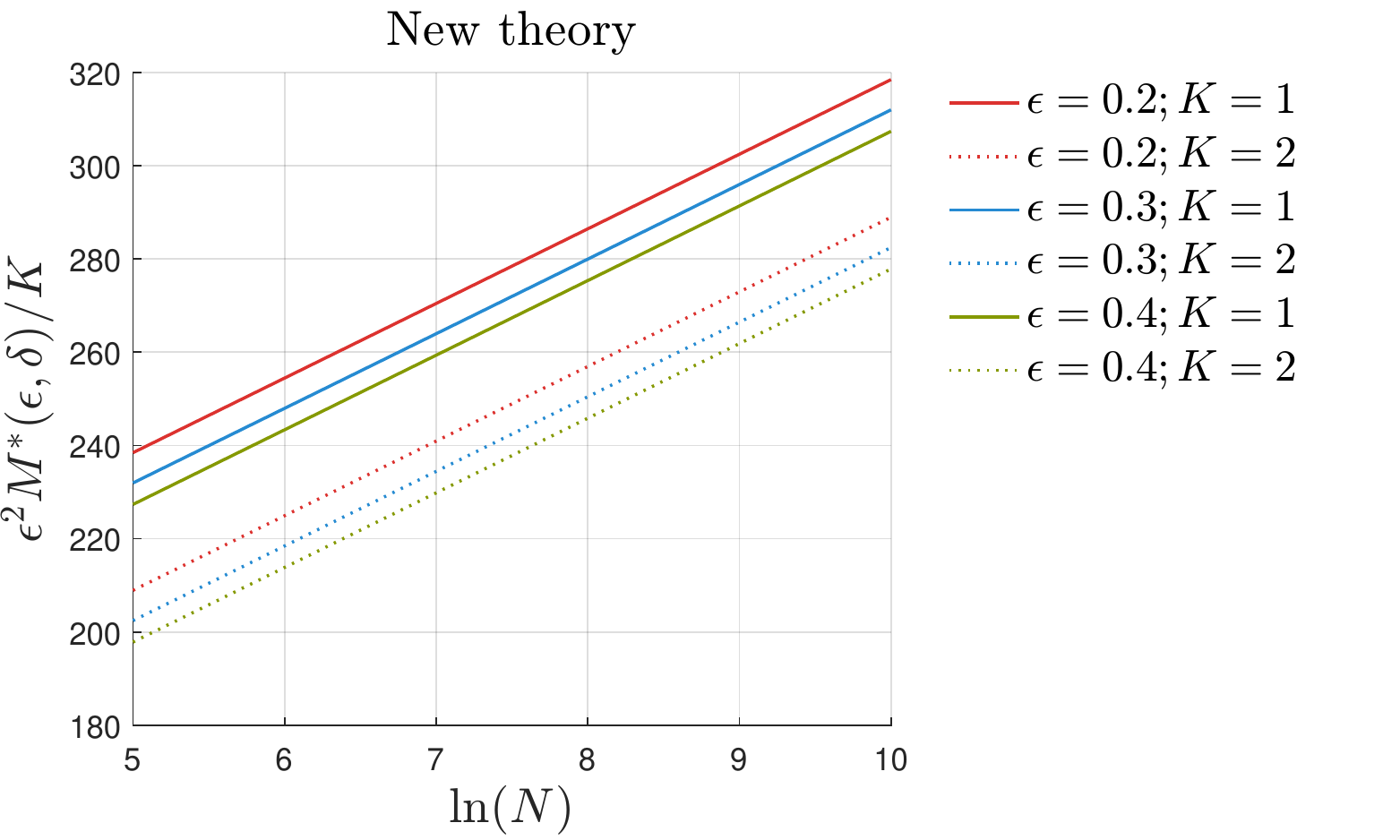}}\label{s:fig:distmanNlgg}


  \end{myenuma}
  \caption[Comparison of ``New theory'' to numerical experiments.]{Comparison of ``New theory'' from \cref{m:eq:numprojprob}\main\ to numerical experiments on random manifolds and random projections, from \cref{m:fig:distman}\main\, plotted separately.
  (\ref{s:fig:distmanVsim}) For the numerical experiments, we fix the ambient dimension $N=1000$.
   For each $M$ ranging from $4$ to $200$, each $\ln \V$, and each $K$, we generate one random manifold $\CM$.
   We then sample $100$ random projections and for each projection $\proj$ we compute the distortion $\distpr(\CM)$, obtaining an empirical distribution of distortions.
   We then compute the distortion $\epsilon(M, K, \ln \V, N)$ that leads to a failure probability of $\delta=0.05$ under this empirical distribution.
   We then interpolate between the result for different values of $M$ to find the minimum value of $M$ sufficient to achieve the desired value of $\epsilon$ with failure probability $\delta$.
   We compare these numerical results for $M$, $\epsilon$, $\V$, $K$ and $N$ with our new theory predicted in \cref{m:eq:numprojprob}\main.
  (\ref{s:fig:distmanNsim})  Here, for numerical experiments, we follow a similar procedure as in (\ref{s:fig:distmanVsim}) except we fix $\V$ to $(10\sqrt{2}/3)^K$ and we vary $N$ from $200$ to $20,000$.
  (\ref{s:fig:distmanVlgg},\ref{s:fig:distmanNlgg}) Here we plot the formula in \cref{m:eq:numprojprob}\main\ for the same values of $M$, $\epsilon$, $\delta$, $\V$, $K$ and $N$ as in (\ref{s:fig:distmanVsim},\ref{s:fig:distmanNsim}).
  }\label{s:fig:distmansep}
\end{figure}

In \cref{s:fig:distmansep} we present the comparison between simulations and theory shown in \cref{m:fig:distman}\main, but plotted separately so that the details can be seen.
Below, we compare the analytic formulae with separated terms so that the coefficients can be compared.

Our bound, from \cref{m:eq:numprojprob}\main, is
\begin{equation}\label{s:eq:ourbound}
  \frac{M^* \epsilon^2}{K} \leq
     16\brk{\frac{\ln\V}{K}}
     + 16\brk{\frac{\ln 1/\delta}{K}}
     + 16\ln\brk{\frac{1}{\epsilon}}
     + 16\ln N
     - 8 \ln {K}
     + 16\ln 9\sqrt{3}\,\e.
\end{equation}
Our underestimate of the bound derived by \citet{Baraniuk2009JLmfld} is
\begin{equation}\label{s:eq:BWbound}
  \frac{M^* \epsilon^2}{K} \leq
     1352\brk{\frac{\ln\V}{K}}
     + 676\brk{\frac{\ln 1/\delta}{K}}
     + 4056\ln\brk{\frac{1}{\epsilon}}
     + 2028\ln N
     + 676 \ln K
     + 676\ln \brk{\frac{3100^4}{4\pi\e}}\!.
\end{equation}
The bound derived by \citet{verma2011note} is for the distortion of tangent vectors.
This is less strict than a bound on the distortion of chords (see \cref{s:sec:distort}), but it is equivalent to the distortion of short chords (see \cref{s:sec:short}), which were the dominant contribution in \cref{m:eq:distproball}\main.
Our underestimate of this bound is
\begin{equation}\label{s:eq:Vermabound}
  \frac{M^* \epsilon^2}{K} \leq
     64\brk{\frac{\ln\V}{K}}
     + 64\brk{\frac{\ln 1/\delta}{K}}
     + 192\ln\brk{\frac{1}{\epsilon}}
     + 0 \cdt\ln N
     + 32 \ln K
     + 32\ln \brk{\frac{384^5 \cdt 169}{\pi\e}}\!,
\end{equation}
where we have dropped subleading terms for large $M,K,\V$ and small $\delta,\epsilon$.

In all but one case, the coefficients in \cref{s:eq:ourbound} are the smallest.
The exception is the $\ln N$ term, which is absent from \cref{s:eq:Vermabound}.
This means that, for sufficiently large $N$, \cref{s:eq:Vermabound} will be the smallest.
Otherwise, the smallest will be \cref{s:eq:ourbound}.
The two bounds cross over at
\begin{equation}\label{s:eq:crossover}
  N = \prn{3.5 \times 10^{27 }} \frac{K^{\frac{3}{2}}}{\epsilon^{11}} \prn{\frac{\V}{\delta}}^{\frac{3}{K}}.
\end{equation}
For the values of $K, \V, \epsilon, \delta$ used here, the crossover is at $N \sim \CO(10^{36})$.



\bibliographystyle{utcaps_plainnat}
\bibliography{randproj-extra}

\providecommand{\natexlab}[1]{#1}
\providecommand{\href}[2]{#2}
\begingroup\raggedright
\begin{thebibliography}{23}

\bibitem[Advani et~al.(2013)Advani, Lahiri, and Ganguli]{advani2013statistical}
M.~Advani, S.~Lahiri, and S.~Ganguli, ``{Statistical mechanics of complex
  neural systems and high dimensional data},''
  \href{http://dx.doi.org/10.1088/1742-5468/2013/03/P03014}{{\em J. Stat.
  Mech.} {\bf 2013} (2013) no.~03, P03014},
  \href{http://arxiv.org/abs/1301.7115}{{\tt arXiv:1301.7115 [q-bio.NC]}}.

\bibitem[Amelunxen et~al.(2014)Amelunxen, Lotz, McCoy, and
  Tropp]{amelunxen2014living}
D.~Amelunxen, M.~Lotz, M.~B. McCoy, and J.~A. Tropp, ``{Living on the edge:
  Phase transitions in convex programs with random data},''
  \href{http://dx.doi.org/10.1093/imaiai/iau005}{{\em Information and
  Inference} {\bf 1} (Apr., 2014) 1--52},
  \href{http://arxiv.org/abs/1303.6672}{{\tt arXiv:1303.6672 [cs.IT]}}.

\bibitem[Baraniuk et~al.(2008)Baraniuk, Davenport, DeVore, and
  Wakin]{baraniuk2008simple}
R.~Baraniuk, M.~Davenport, R.~DeVore, and M.~Wakin, ``{A simple proof of the
  restricted isometry property for random matrices},''
  \href{http://dx.doi.org/10.1007/s00365-007-9003-x}{{\em Constructive
  Approximation} {\bf 28} (2008) no.~3, 253--263}.

\bibitem[Baraniuk and Wakin(2009)]{Baraniuk2009JLmfld}
R.~G. Baraniuk and M.~B. Wakin, ``{Random projections of smooth manifolds},''
  \href{http://dx.doi.org/10.1007/s10208-007-9011-z}{{\em Foundations of
  Computational Mathematics} {\bf 9} (Dec., 2009) 51--77}.

\bibitem[Blum(2006)]{blum2006random}
A.~Blum, ``{Random projection, margins, kernels, and feature-selection},''
  \href{http://dx.doi.org/10.1007/11752790\_3}{{\em Subspace, Latent Structure
  and Feature Selection} {\bf 3940} (2006) 52--68}.

\bibitem[Candes and Tao(2005)]{candes2005decoding}
E.~J. Candes and T.~Tao, ``{Decoding by Linear Programming},''
  \href{http://dx.doi.org/10.1109/TIT.2005.858979}{{\em IEEE Trans. Inf.
  Theory} {\bf 51} (Dec., 2005) 4203--4215},
  \href{http://arxiv.org/abs/0502327}{{\tt arXiv:0502327 [math.MG]}}.

\bibitem[Clarkson(2008)]{clarkson2008tighter}
K.~L. Clarkson, ``Tighter Bounds for Random Projections of
  Manifolds,''\href{http://dx.doi.org/10.1145/1377676.1377685}{ in {\em Proc.
  24th Annual Symp. on Computational Geometry}, SCG '08,  pp.~39--48.
\newblock ACM, New York, NY, USA, 2008}.

\bibitem[Dasgupta and Gupta(2003)]{Dasgupta2003JLlemma}
S.~Dasgupta and A.~Gupta, ``{An Elementary Proof of a Theorem of Johnson and
  Lindenstrauss},'' \href{http://dx.doi.org/10.1002/rsa.10073}{{\em Random
  Structures and Algorithms} {\bf 22} (Jan., 2003) 60--65}.

\bibitem[Davenport et~al.(2007)Davenport, Duarte, Wakin, Laska, Takhar, Kelly,
  and Baraniuk]{davenport2007smashed}
M.~A. Davenport, M.~F. Duarte, M.~B. Wakin, J.~N. Laska, D.~Takhar, K.~F.
  Kelly, and R.~Baraniuk, ``{The smashed filter for compressive classification
  and target recognition},'' \href{http://dx.doi.org/10.1117/12.714460}{{\em
  Proc. SPIE 6498, Computational Imaging V} {\bf 6498} (Jan., 2007)
  64980H--1--12}.

\bibitem[Duarte et~al.(2007)Duarte, Davenport, Wakin, Laska, Takhar, Kelly, and
  Baraniuk]{duarte2007multiscale}
M.~F. Duarte, M.~A. Davenport, M.~B. Wakin, J.~N. Laska, D.~Takhar, K.~F.
  Kelly, and R.~G. Baraniuk, ``{Multiscale Random Projections for Compressive
  Classification},''\href{http://dx.doi.org/10.1109/ICIP.2007.4379546}{ in {\em
  IEEE International Conference on Image Processing, 2007. ICIP 2007.}, vol.~6,
   pp.~161--164, IEEE.
\newblock 2007}.

\bibitem[Duarte et~al.(2006)Duarte, Davenport, Wakin, and
  Baraniuk]{duarte2006sparse}
M.~F. Duarte, M.~A. Davenport, M.~B. Wakin, and R.~G. Baraniuk, ``{Sparse
  Signal Detection From Incoherent
  Projections},''\href{http://dx.doi.org/10.1109/ICASSP.2006.1660651}{ in {\em
  Acoustics, Speech and Signal Processing, ICASSP Proceedings.}, vol.~3,
  pp.~305--308, IEEE.
\newblock 2006}.

\bibitem[Ganguli and Sompolinsky(2012)]{ganguli2012annrevs}
S.~Ganguli and H.~Sompolinsky, ``{Compressed Sensing, Sparsity, and
  Dimensionality in Neuronal Information Processing and Data Analysis},''
  \href{http://dx.doi.org/10.1146/annurev-neuro-062111-150410}{{\em Annual
  Review of Neuroscience} {\bf 35} (2012) no.~1, 485--508}.

\bibitem[Gao and Ganguli(2015)]{simplicitycomp15}
P.~Gao and S.~Ganguli, ``{On simplicity and complexity in the brave new world
  of large-scale neuroscience},''
  \href{http://dx.doi.org/10.1016/j.conb.2015.04.003}{{\em Current Opinion in
  Neurobiology} {\bf 32} (2015) 148--155},
  \href{http://arxiv.org/abs/1503.08779}{{\tt arXiv:1503.08779 [q-bio.NC]}}.

\bibitem[Haupt et~al.(2006)Haupt, Castro, Nowak, Fudge, and
  Yeh]{haupt2006compressive}
J.~Haupt, R.~Castro, R.~Nowak, G.~Fudge, and A.~Yeh, ``{Compressive Sampling
  for Signal
  Classification},''\href{http://dx.doi.org/10.1109/ACSSC.2006.354994}{ in {\em
  2006 Fortieth Asilomar Conference on Signals, Systems and Computers},
  pp.~1430--1434, IEEE.
\newblock 2006}.

\bibitem[Hegde et~al.(2007)Hegde, Wakin, and Baraniuk]{hegde2007random}
C.~Hegde, M.~Wakin, and R.~Baraniuk, ``Random Projections for Manifold
  Learning,''\href{http://papers.nips.cc/paper/3191-random-projections-for-manifold-learning}{
  in {\em Adv.\ Neural Inf.\ Process.\ Syst.\ 20}, J.~C. Platt, D.~Koller,
  Y.~Singer, and S.~T. Roweis, eds.,  pp.~641--648.
\newblock Curran, 2007}.

\bibitem[Indyk and Motwani(1998)]{indyk1998approximate}
P.~Indyk and R.~Motwani, ``{Approximate nearest neighbors: towards removing the
  curse of
  dimensionality},''\href{http://dx.doi.org/10.4086/toc.2012.v008a014}{ in {\em
  Proceedings of the thirtieth annual ACM symposium on Theory of computing},
  pp.~604--613, ACM.
\newblock 1998}.

\bibitem[Johnson and Lindenstrauss(1984)]{Johnson1984extensions}
W.~B. Johnson and J.~Lindenstrauss, ``{Extension of Lipschitz maps into a
  Hilbert space},'' \href{http://dx.doi.org/10.1090/conm/026/737400}{{\em
  Contemp. Math.} {\bf 26} (1984) no.~189-206, 189--206}.

\bibitem[Lawrence and Hyv{\"{a}}rinen(2005)]{Lawrence2005gplvm}
N.~Lawrence and A.~Hyv{\"{a}}rinen, ``{Probabilistic non-linear principal
  component analysis with Gaussian process latent variable models},''
  \href{http://www.jmlr.org/papers/v6/lawrence05a.html}{{\em Journal of Machine
  Learning Research} {\bf 6} (2005) no.~Nov, 1783--1816}.

\bibitem[Oymak and Tropp(2015)]{oymak2015universality}
S.~Oymak and J.~A. Tropp, ``{Universality laws for randomized dimension
  reduction, with applications},'' \href{http://arxiv.org/abs/1511.09433}{{\tt
  arXiv:1511.09433 [math.PR]}}.

\bibitem[Pinkus(1985)]{pinkus2012n}
A.~Pinkus, \href{http://dx.doi.org/10.1007/978-3-642-69894-1}{{\em {$n$}-widths
  in approximation theory}}, vol.~7.
\newblock Springer, 1985.

\bibitem[Verma(2011)]{verma2011note}
N.~Verma, ``{A note on random projections for preserving paths on a
  manifold},''\href{http://citeseerx.ist.psu.edu/viewdoc/summary?doi=10.1.1.295.9077}{
  Tech. Rep. CS2011-0971, UC San Diego, 2011}.

\bibitem[Wishart(1928)]{wishart1928generalised}
J.~Wishart, ``{The generalised product moment distribution in samples from a
  normal multivariate population},''
  \href{http://dx.doi.org/10.1017/S0305004100011063}{{\em Biometrika} {\bf 20}
  (1928) no.~1, 32--52}.

\bibitem[Zhou et~al.(2009)Zhou, Lafferty, and Wasserman]{zhou2009compressed}
S.~Zhou, J.~Lafferty, and L.~Wasserman, ``{Compressed and privacy-sensitive
  sparse regression},'' \href{http://dx.doi.org/10.1109/TIT.2008.2009605}{{\em
  IEEE Trans. Inf. Theory} {\bf 55} (2009) no.~2, 846--866}.

\end{thebibliography}
\endgroup

\end{document}